\PassOptionsToPackage{table}{xcolor}
\documentclass[11pt]{article}
\usepackage{arena-layout}
\usepackage{latexsym}
\usepackage{xspace}
\usepackage{enumitem}
\usepackage{placeins}

\title{On Evaluating User Simulators for Interactive Agents}

\author{%
  Hengguang~Zhou\textsuperscript{1,2} \quad
  I-Hung~Hsu\textsuperscript{2} \quad
  Li~Chen\textsuperscript{2} \quad
  Priyanshi~Shrivastava\textsuperscript{2}\\[3pt]
  Anastasios~Angelopoulos\textsuperscript{2} \quad
  Wei-Lin~Chiang\textsuperscript{2} \quad
  Ion~Stoica\textsuperscript{2} \quad
  Cho-Jui~Hsieh\textsuperscript{1,2}%
}
\renewcommand{\affiliations}{\textsuperscript{1}\,UCLA \quad \textsuperscript{2}\,Arena}

\newcommand{\chg}[1]{\textcolor{blue}{#1}}   
\newcommand{\orig}[1]{\begingroup\color{red}#1\endgroup}  
\newif\iffullpaper
\fullpapertrue
\begin{document}
\suppressfloats[t]
\maketitle
\iffullpaper
\begin{tcolorbox}[enhanced,colback=ArenaPaper,colframe=ArenaBorderFaint,
boxrule=0.75pt,arc=6pt,boxsep=0pt,left=12pt,right=12pt,top=12pt,bottom=12pt,
before skip=16pt,after skip=12pt]
\begin{abstract}

User simulators offer a scalable way to evaluate interactive agents, but how to assess their quality remains underexplored. In particular, it is unclear whether existing user simulator evaluation frameworks reliably detect meaningful failures in simulated user behavior. This uncertainty makes it difficult to interpret simulator evaluation. To understand what a high benchmark score tells us about simulator quality, we \textbf{(meta-)evaluate} the metrics themselves. To instantiate this meta-evaluation, we deliberately perturb simulated replays to induce identified failures, then test whether the metrics penalize these failures. Evaluation methods from four prior studies detect at most 6 of the 10 perturbation categories; some of their aggregated scores even favor populations instructed to accept broken work. Motivated by these gaps, we propose a new evaluation suite that combines utility of user simulators for evaluating agents and two measures of fidelity to real users: population-level fidelity assesses collective interaction patterns, while trace-level fidelity assesses the preservation of individual users’ goals, constraints, and persona. Under the same detection criterion, our evaluation suite detects all 10 perturbations. We evaluated six user simulators on 351 real multi-turn sessions in which 348 distinct users across 23 languages and agents work together to create and refine artifacts such as full-stack web apps, games, design assets, and automated workflows. In this process, we find that single-level measures can miss when a simulated user pursues the wrong goal, ignores a constraint, or behaves out of character, highlighting the need to assess fidelity at comprehensive granularities. Project page: \url{https://hengguangzhou.github.io/on-evaluating-user-simulators/}
\end{abstract}

\end{tcolorbox}

\section{Introduction}
With the capability advances, LLM agents help users write code, create documents, and revise other artifacts~\citep{kimiteam2026k25,openai2026gpt54,anthropic2026opus46,yang2024sweagent,xie2024osworld} through multi-turn interactions~\citep{dou2025simulatorarena,wu2026swetogether,baumann2026swechat}. During these interactions, users clarify what they want, inspect the work produced, and provide feedback. Evaluating this process requires a user who can respond to the agent's progress, but recruiting real users for every evaluation is costly and makes comparisons difficult to repeat~\citep{dou2025simulatorarena,phan2026mimesis,jurenka2024responsible}. User simulators offer a practical way to run these interactions at scale and are used in both offline agent evaluation and training~\citep{barres2025tau2,qian2025userrl, suh2026utility}. However, using simulated
users in place of real users can change measured agent success rates~\citep{seshadri2026lost}.


To assess whether simulators can reliably stand in for real users in multi-turn interactions, recent studies have developed several ways to evaluate user simulators, including similarity to human language and behavior, consistency with user specifications~\citep{zheng2026argus,jain2026userproxybench}, and agreement with human assessments of agent performance~\citep{dou2025simulatorarena,lu2026vista,zhou2026sim2real,wu2026swetogether}. What remains less clear is whether these measures reliably identify behavior that makes a simulator unsuitable for agent evaluation where artifacts deeply involved in the interaction. For example, a simulated user may sound natural while approving a broken artifact or overlooking an important requirement. Evaluating fidelity is also difficult because a replay should preserve the original user's needs while responding to an agent whose actions may differ from the original session. We therefore need to examine how evaluation metrics respond when these aspects of simulation go wrong.

To understand what a high benchmark score tells us about simulator quality, we \textbf{(meta-)evaluate} the metrics themselves. Specifically, we introduce failures through controlled perturbations~\citep{sai-etal-2021-perturbation} to the simulator's setup, then replay each session with the assigned agent. We test whether the metrics assign lower scores to these perturbed replays than to controls generated from the same source sessions and agent assignments. We design 10 perturbations based on three sources of motivation. First, we draw on reported failure modes, including overly helpful
simulated users~\citep{naous2026flipping}, premature information
disclosure~\citep{zhou2026sim2real,jain2026userproxybench,liu2026disclosure},
and self-preference in LLM evaluation~\citep{zheng2023judging,panickssery2024llm,spiliopoulou2025playfavorites}.
Second, we draw on failures observed in our analysis of 442,516 real sessions and 20,100 replays.
Third, we include basic defects that evaluation metrics should readily detect. These include changing the user's goal, removing constraints, forcing early termination, encouraging acceptance of broken work, and favoring a selected agent. We ask whether the existing user simulator evaluation metrics penalize these changes and what replays are retained when using their score for filtering.


The results reveal several gaps in the evaluation suites from SimulatorArena, VISTA, SWE-Together, and Sim2Real~\citep{dou2025simulatorarena,lu2026vista,wu2026swetogether,zhou2026sim2real}. We count a perturbation as detected if at least one metric in an evaluation suite shows a reliable score decrease relative to the control (Section~\ref{sec:sensitivity-overview}). Under this criterion, the evaluation suites adapted from SimulatorArena, VISTA, and SWE-Together detect six, five, and six of the ten perturbations, respectively, and Sim2Real detects none. All four evaluation suites fail to detect the agent favoritism perturbation (more favorable feedback from the simulated user toward a selected agent) despite the shifts it causes in agents’ scores and ranks. The composite scores for some metrics even favor perturbed replays that accept broken work. These findings suggest that high scores on existing criteria do not necessarily establish that a simulator provides useful feedback for help evaluating interactive agents.

Motivated by these findings, we introduce a new evaluation suite that measures user simulators at three levels illustrated in Figure~\ref{fig:evaluation-framework-implementation}. \textbf{Downstream ranking agreement} checks whether simulated interactions recover a supplied reference ranking of agents. This assesses the simulator's utility for agent evaluation, but gives limited information about why it succeeds or fails. We therefore also measure \textbf{population-level fidelity}, which compares collective interaction patterns with those of real users, and \textbf{trace-level fidelity}, which checks whether each replay preserves the user's goals, constraints, and persona. The two fidelity levels help distinguish broad behavior shifts from failures within individual sessions. Together, these measures provide a stronger and more holistic assessment of how well a simulator supports agent evaluation at a broader range of failures than any of the existing benchmarks we evaluated.

We show that our evaluation suite detects all ten tested perturbations under the same criterion in Sec~\ref{sec:metaeval-results}. We also test whether our evaluation scores can help filtering out simulator failures. We report the fraction of control traces retained at a threshold that admits 5\% of perturbed samples. Our metric retains 98\% of control
populations in the agent favoritism filtering experiment, compared with 10\%
for the best external benchmark. Under the same conditions,
action distribution retains 100\% of control populations in the
approval-of-broken-work experiment, compared with 26\% for the best
external benchmark. At the trace level, the persona score retains
68\% of real-user traces, compared with 28\% for the best external
benchmark. Finally, we apply the evaluation suite to six simulators on 351 real sessions. None of the six simulators we evaluated leads across all three levels. The simulator from SimulatorArena achieves the highest downstream ranking agreement, while our proposed profile-based simulator lead on population-level and trace-level fidelity. These results show the value of examining all three levels when assessing a simulator.


Our contribution are four-fold:
\begin{itemize}[topsep=0pt, itemsep=-1.5pt, leftmargin=13pt]
\item A perturbation-based meta-evaluation protocol for examining what user-simulator evaluation metrics detect and retain in Section \ref{sec:metaeval}.

\item A new evaluation suite grounded in real user sessions that combines downstream ranking agreement with population- and trace-level fidelity in Section \ref{sec:suite}.


\item A 351 real-user sessions used in our evaluation to support future research on user simulation for agent evaluation.

\item An empirical study of existing evaluation criteria, their use in data selection, and the strengths and weaknesses of six simulators in Section \ref{sec:application}.

\end{itemize}

\section{Related Work}
\paragraph{User simulators for agent evaluation.}
Task-oriented benchmarks such as $\tau^2$-bench use an environment-grounded user simulator in interactive agent evaluation \citep{barres2025tau2}. SimulatorArena studies open-ended assistant interactions using simulated users and compares simulator-derived assessments with human judgments \citep{dou2025simulatorarena}. SWE-Together reconstructs developer sessions and uses a state-conditional user simulator to preserve task intent while adapting feedback to the new agent \citep{wu2026swetogether}. SWE-DIVE converts coding benchmarks into multi-turn evaluations in which a simulated user responds to the agent's replies and codebase changes~\citep{brehme2026swedive}, and Cowork-Gym synthesizes persistent work environments with a stateful simulated user that observes the workspace~\citep{zou2026coworkgym}. Lost in Simulation documents discrepancies between agent success rates measured with simulated and real users~\citep{seshadri2026lost}. Disclosure-Gated User Simulation examines how disclosure mechanisms affect companion-agent evaluation~\citep{liu2026disclosure}. These settings motivate assessing both simulated behavior and its use in agent evaluation.

\paragraph{Evaluating the simulator.}
Surveys distinguish behavioral consistency from downstream utility~\citep{pietquin2013metrics} and discuss simulation of general and individual users~\citep{ni2026survey}. Prior work considers behavioral realism, consistency with user specifications, and similarity to real-user populations \citep{dou2025simulatorarena,lu2026vista,wu2026swetogether,zhou2026sim2real}. ARGUS evaluates behavioral consistency with scenario requirements~\citep{zheng2026argus}, while UserProxyBench measures adherence to task-specific user instructions and separately checks user tool execution in Telecom~\citep{jain2026userproxybench}. UserArena evaluates shopping user simulators at five levels, from output validity to sequential behavior, personalization, and outcomes~\citep{zhao2026userarena}. CUE assesses whether simulated users reproduce the success and failure patterns of real interactions~\citep{kantharuban2026cue}, and AutoSim compares simulator-derived rankings of conversational search systems with rankings based on relevance judgments~\citep{zerhoudi2026autosim}. Section~\ref{sec:metaeval-setup} describes the evaluation criteria used in our comparisons. These studies provide complementary criteria; we examine how such criteria respond to controlled replay perturbations and what data they retain when used as filters.

\paragraph{Meta-evaluation of LLM judges.}
Meta-evaluation is the systematic assessment of an evaluation's quality and utility~\citep{stufflebeam2001metaevaluation}. Our meta-evaluation treats the scoring criteria themselves as objects of study, following prior work on metric validity and perturbation testing~\citep{mathur-etal-2020-tangled,sai-etal-2021-perturbation,muller2025grouse}. It compares scores for control and perturbed interactions generated from the same source sessions, then examines selection under score thresholds. Perturbations act on the simulator's setup, allowing the resulting interaction to evolve; the test therefore concerns an interactive generation condition rather than a static textual edit.

\fi

\section{User-Simulator Evaluation}
\label{sec:user-simulator-evaluation}

\subsection{Problem Formulation}
\label{sec:problem-formulation}

\begin{figure}[!htbp]
\centering
\includegraphics[width=\textwidth]{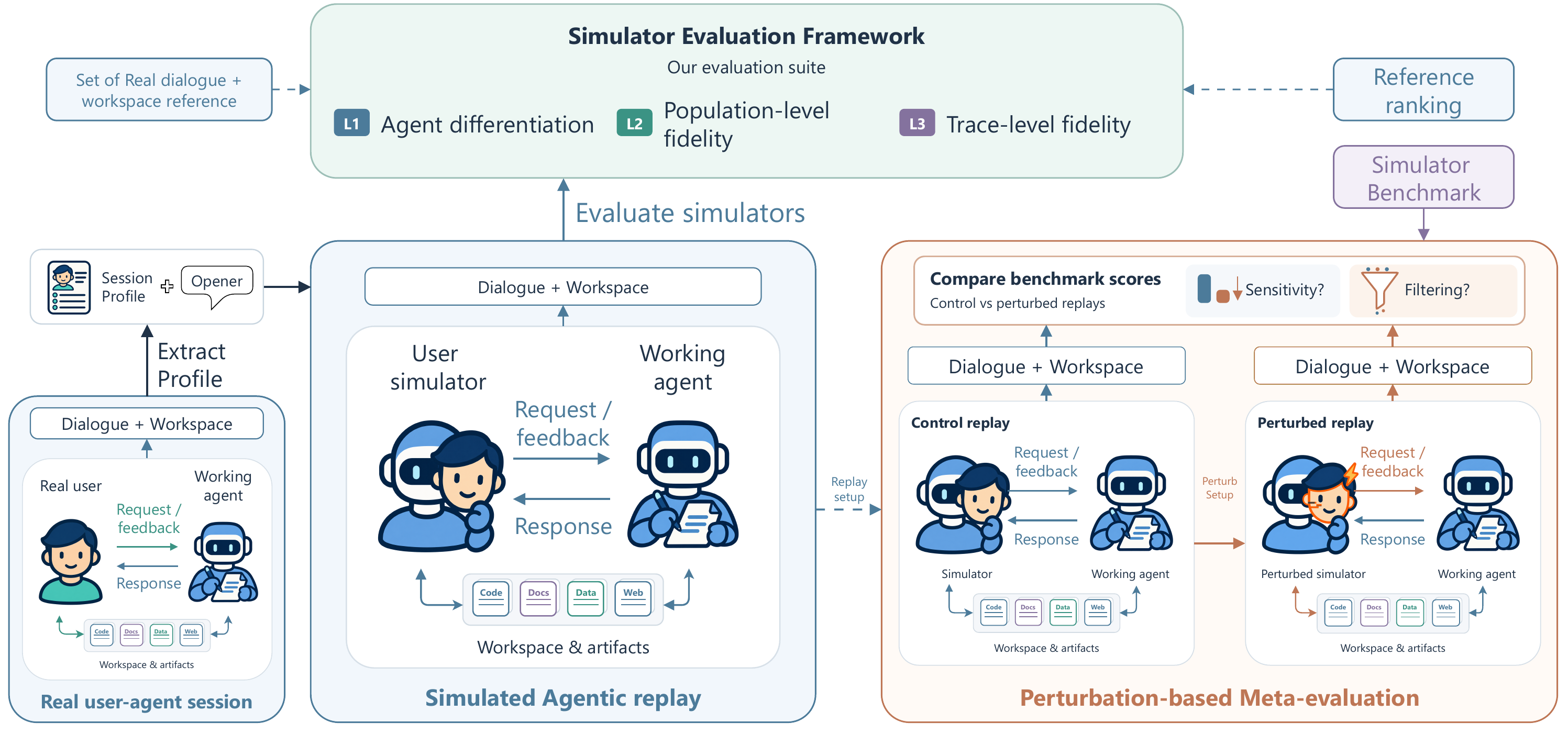}
\caption{Simulator evaluation and perturbation-based meta-evaluation. Replays are compared through agent ranking agreement (L1), population-level fidelity (L2) and trace-level fidelity (L3). Paired control and perturbed replays test benchmark sensitivity and filtering.}
\label{fig:overview}
\end{figure}

We study user simulation through replay of real user sessions in
artifact-producing workspaces. We assume access to user--agent traces
containing dialogue and, where available, uploaded artifacts and
workspace evidence. As shown in Figure~\ref{fig:overview}, a source
session provides the initial request and a representation of the user
for replay. The simulator interacts with an agent whose actions can
modify the workspace and produce artifacts, using the developing
dialogue and available workspace evidence to generate feedback or stop.

The replay agent may produce different responses and artifacts from
the original agent. The source trace therefore does not specify the
correct user response at every turn. Fidelity requires preserving the
source user's intent and constraints while adapting feedback to the
new interaction and the work actually produced.

\subsection{Evaluation Principles}
\label{sec:evaluation-principles}

Evaluating user simulators for interactive agents requires examining
the downstream utility and fidelity to the users they represent.
We organize these requirements into three complementary levels.

\paragraph{Downstream evaluation utility (L1).}
Simulated interactions should support valid assessments of agent
performance in the intended setting~\citep{pietquin2013metrics}.
Prior work shows that simulator-based evaluations can misestimate
agent success rates relative to evaluations with real
users~\citep{seshadri2026lost}.
Reliable agent comparisons therefore require simulated users to
respond appropriately to differences in agent performance.
Their feedback and stopping decisions should reflect the quality
of the agent's work under comparable standards across agents.
In artifact-producing tasks, evaluation must account for how well
the resulting work meets user needs: plausible dialogue alone
provides insufficient evidence of successful performance~\citep{bodhwani2026gauge,jain2026userproxybench,zheng2026argus}.
These requirements motivate evaluating downstream utility directly:
whether simulated interactions support reliable comparisons of agent
performance. This complements behavioral fidelity by checking whether
the simulator serves its intended purpose as an evaluation tool.

\paragraph{Population-level fidelity (L2).}
Simulated populations should reproduce relevant patterns of real-user
behavior. These include how users introduce requirements, provide
information, respond to progress and errors, and decide that work is
complete. Reported failures include excessive cooperation~\citep{zhou2026sim2real,chopra2026personapolicies}
and premature or unconditional information
disclosure~\citep{jain2026userproxybench,liu2026disclosure}.
In workspace interactions, fidelity also concerns how feedback
relates to the state and quality of artifacts. Systematic differences
in these patterns change the evaluation conditions agents encounter.

\paragraph{Trace-level fidelity (L3).}
Individual interactions should preserve the represented user's intent,
constraints, and persona~\citep{ni2026survey} while responding to the agent's actions.
Prior work highlights fidelity failures such as goal and constraint
violations, premature termination, ungrounded confirmations,
fabricated observations, and claims of actions that were never
performed~\citep{zheng2026argus,jain2026userproxybench}.
Population-level similarity can conceal these failures within
individual replays, motivating evaluation against the particular
user each replay represents.
In our real-user traces, users inspect produced artifacts and base
their acceptance or requests for changes on whether the work meets
their requirements.
For artifact-producing tasks, fidelity therefore concerns both
preserving the source user's requirements and applying their
judgment criteria when assessing the resulting work.

Existing benchmarks typically cover only parts of these principles. Population-level comparisons, such as those used within Sim2Real, characterize collective behavior, but matching aggregate patterns can leave changes to individual users' intent, constraints, or persona unobserved \citep{zhou2026sim2real}. Individual fidelity therefore requires examining whether each replay preserves the particular user it represents. VISTA evaluates consistency with user specifications, while SWE-Together checks whether simulated guidance preserves the original task scope. However, consistency and scope preservation alone do not establish whether the simulator reproduces the represented user's degree and manner of participation \citep{lu2026vista,wu2026swetogether}. SimulatorArena and SWE-Together assess agent outcomes alongside message similarity or preservation of source-user intents, but assessing these separately does not directly establish whether the user's acceptance or criticism is justified by the work produced \citep{dou2025simulatorarena,wu2026swetogether}.

Motivated by the above-mentioned limitations, we develop a meta-evaluation framework to quantitatively evaluate existing user simulator evaluation suites in 
Section~\ref{sec:metaeval}, and then propose a new evaluation suite to overcome those issues in Section~\ref{sec:suite}.

\section{Perturbation-Based Meta-Evaluation}
\label{sec:meta-evaluation}
\label{sec:metaeval}

Assessing the reliability of an evaluation method is itself a
challenging problem. Agreement with human judgments can depend strongly
on the outputs used for assessment and carry substantial statistical
uncertainty
\citep{mathur-etal-2020-tangled,deutsch-etal-2021-statistical}.
Moreover, agreement on overall quality does not establish that a
metric captures each relevant evaluation dimension
\citep{sai-etal-2021-perturbation}. For user simulation, this challenge is compounded by the replay
setting: source traces do not prescribe responses to new agent
actions, and appropriate feedback depends on evolving dialogue,
workspace state, and artifacts.

We therefore propose a perturbation-based meta-evaluation.
Following criteria-targeted perturbation testing
\citep{sai-etal-2021-perturbation}, we modify simulator setups to
challenge the principles in
Section~\ref{sec:evaluation-principles}.
Figure~\ref{fig:overview} shows the procedure: we generate paired control and perturbed replays to test whether a failure lowers a replay's score relative to its control (\emph{sensitivity}), and whether high scores still separate valid replays from flawed ones across different sessions (\emph{filtering}).
\subsection{Session Data}
\label{sec:meta-evaluation-data}

Our meta-evaluation uses real user--agent sessions involving
goal-directed work on deliverables as source sessions. We use two sets that differ
primarily in interaction length. The \emph{easy} set contains
shorter interactions (median: 7 user turns; 10th--90th percentile
range: 5--11). The \emph{hard} set  involves longer-running
projects with more complex tasks and deliverables (median:
26 user turns; 10th--90th percentile range: 8--65).

\subsection{Replay Setup}
\label{sec:control-replay}
\label{sec:replay-protocol}

For each source-session--agent pair and each perturbation,
we compare a \emph{control replay} generated using the baseline
setup below with a \emph{perturbed replay} generated using the
corresponding intervention from Section~\ref{sec:perturbation-design}.

\paragraph{Control replay.}
We extract a session profile from the source dialogue, final workspace,
and uploaded materials, describing the intended deliverable,
constraints, preferences, persona, and stopping conditions. 
The simulator uses this profile to represent the source user in a
new interaction with the assigned agent. See Appendix~\ref{sec:profile-example} for a profiling example.

The interaction begins with the original opener and associated files
in a persistent sandbox. Later uploads are available for attachment
during interaction. At subsequent turns, the simulator receives the
profile, developing dialogue, and current workspace evidence, including
text-file contents, metadata, and tool-error summaries. It generates
feedback or stops, with instructions to ground acceptance in evidence
supporting the agent's claims. Each replay ends when the simulator stops or reaches the cap of 30 user turns.

\paragraph{Perturbed replay.}
We modify a specified part of this baseline setup, such as the profile,
response instructions, or turn budget for perturbation. The source session, opener,
initial files, and assigned agent remain the same.

Both replay use \texttt{gpt-5.6-terra} as the simulator model. Each condition runs
with a fresh agent instance and a separate workspace, allowing its
dialogue and artifacts to develop independently. The following
subsection defines the interventions and the evaluation requirements
they challenge.

\subsection{Perturbation Design}
\label{sec:perturbation-design}
\label{sec:perturbations}
\label{par:scope}

We design 10 perturbations drawing on three sources:
failure modes reported in prior work, failures observed in
our replay analyses, and basic defects that an evaluation
should readily detect. We organize these perturbations according
to the evaluation principles introduced in
Section~\ref{sec:evaluation-principles}.
We describe each intervention below, with implementation details in
Appendix~\ref{sec:appendix-perturbations}.

\paragraph{Downstream evaluation utility (L1).}
\emph{Agent favoritism} instructs the simulated user to favor a particular agent by excusing errors, accepting imperfect deliverables, and giving warmer feedback.
\emph{Profile extraction artifact} turns dissatisfaction with the original agent into a persistent user trait~\citep{tennenholtz2026controllable}, making the simulator remain critical even when the replay agent meets requirements.

\paragraph{Population-level fidelity (L2).}
\emph{Approve broken work} instructs the simulated user to approve deliverables despite errors or missing requirements.
\emph{Premature stop} truncates replays early, reducing opportunities for users to inspect deliverables and request corrections.
\emph{Front-loading} moves requirements and acceptance criteria that would otherwise emerge during interaction into the opening messages.

\paragraph{Trace-level fidelity (L3).}
\emph{Adversarial goal} replaces the intended goal and deliverable with a plausible alternative, directing later feedback toward the altered task.
\emph{All constraints removed} removes hard requirements from the profile while preserving those expressed in the original opener.
\emph{Persona swap} replaces the source user's writing style with a contrasting register while preserving the goal and requirements.
\emph{Co-debugger} instructs the simulated user to diagnose problems and suggest fixes, providing technical assistance beyond the represented user's role.
\emph{Language switch} changes the simulated user's language after the original opener.

\subsection{Meta-Evaluation Protocol}
\label{sec:meta-evaluation-protocol}
\label{sec:metaeval-setup}

Our meta-evaluation takes an evaluation metric or evaluation suite as input.
We test whether it detects controlled perturbations
(\emph{sensitivity}) and distinguishes controls from perturbed
candidates (\emph{filtering}). All scores are oriented so that
higher is better.

\paragraph{Sensitivity.}
For trace-level comparisons, we compare matched replay pairs
indexed by $i$ for metric $m$ and perturbation $p$. Each pair shares the source session and agent
assignment: $x_i^C$ is the control replay and $x_i^p$ its perturbed
counterpart. Let $s_m(x)$ be metric $m$'s score for replay $x$.
Let $I_{m,p}$ contain the indices of pairs with both scores available,
and let $|I_{m,p}|$ denote their number. We define
\begin{equation}
\begin{aligned}
d_{i,m,p} &= s_m(x_i^C)-s_m(x_i^p),\\
\bar d_{m,p}
&= \frac{1}{|I_{m,p}|}
   \sum_{i\in I_{m,p}} d_{i,m,p}.
\end{aligned}
\label{eq:trace-difference}
\end{equation}
Here, $d_{i,m,p}$ is the paired score decrease for case $i$,
and $\bar d_{m,p}$ is its mean across available pairs.
A positive value means that the metric scores the perturbed replay
lower. We report the paired win rate
\begin{equation}
W_{m,p}
= \frac{1}{|I_{m,p}|}
  \sum_{i\in I_{m,p}} h(d_{i,m,p}),
\label{eq:paired-win-rate}
\end{equation}
where $h(d)$ equals $1$ for $d>0$, $1/2$ for $d=0$, and $0$ for
$d<0$. Thus, $W_{m,p}$ measures how often the control scores higher,
giving ties half credit. 

For population comparisons, we repeatedly sample sets of sessions to test whether the metric consistently assigns lower scores to perturbed populations. Let $B = 200$ denote the default number of sampling rounds. For each round $b \in \{1, \ldots, B\}$, we sample 100 sessions to obtain the control and perturbed sets, $X_b^C$ and $X_b^p$, from the same session. Let $S_m(X)$ be metric $m$'s population score: population metrics
use a fixed real-user reference for each comparison, while trace metrics are
averaged within $X$. We define
\begin{equation}
D_{b,m,p}=S_m(X_b^C)-S_m(X_b^p).
\label{eq:population-difference}
\end{equation}
Here, $D_{b,m,p}$ is the paired population score decrease in draw $b$.
The population win rate is
$B^{-1}\sum_{b=1}^{B}h(D_{b,m,p})$,
the fraction of draws in which
the control population scores higher than its matched perturbed population, with half credit for ties.
Population-level detection requires the 2.5th percentile of $\{D_{b,m,p}\}_{b=1}^{B}$ to be above zero.

Agent-targeted perturbations follow the same population-comparison
setup. Each draw samples session--agent pairs for the targeted
agents and inserts either their control or perturbed replays into
an otherwise fixed evaluation pool (Appendix~\ref{sec:appendix-agent-sensitivity}).

An evaluation suite detects a perturbation if any constituent metric meets
the criterion at any applicable level. Each perturbation is
counted only once.

\paragraph{Filtering.}
Sensitivity tests whether a perturbation lowers the score relative to its matched control, yet differences in session score intercepts can still offset that drop across heterogeneous sessions, letting a perturbed run from a shorter or simpler session outscore a valid control from a longer or more complex one. Filtering therefore tests whether scores distinguish controls from perturbed candidates across a candidate pool, revealing which simulations receive high scores and survive selection.

At the trace level, we test whether scores distinguish perturbed
replays from control replays and real-user traces.
For agent differentiation and population-level fidelity, we
compare populations containing only control replays with
matched populations constructed by replacing 10\%, 25\%, 50\%
or 100\% of the sampled control replays with their perturbed
counterparts. Within each comparison, all metrics use the same
sampled replay pairs and replacement choices.

For each metric, filtering retains traces or populations whose
scores meet or exceed a threshold. Across all comparisons,
we report AUROC to measure how well scores distinguish the
control or real-user group from the perturbed group.

For trace-level filtering, we additionally report the fraction
of control replays or real-user traces retained when the
threshold allows 5\% or 10\% of perturbed replays to pass.
Higher retention at the same perturbed-survival rate indicates
more effective filtering.

\section{Evaluation Framework}
\label{sec:suite}

As shown in Section~\ref{sec:metaeval-results}, existing evaluation suites fail miserably on perturbations: they overlook biased feedback toward particular agents, changes to user goals or constraints, and acceptance of broken work, or even rewarding perturbed replays. These failures leave gaps in assessing agent differentiation, population-level fidelity, and trace-level fidelity. We therefore propose an evaluation framework that measures these three aspects and provides more detailed feedback on simulator behavior. Below, we describe how each component is implemented.

\begin{figure}[!htbp]
\centering
\includegraphics[width=\textwidth]{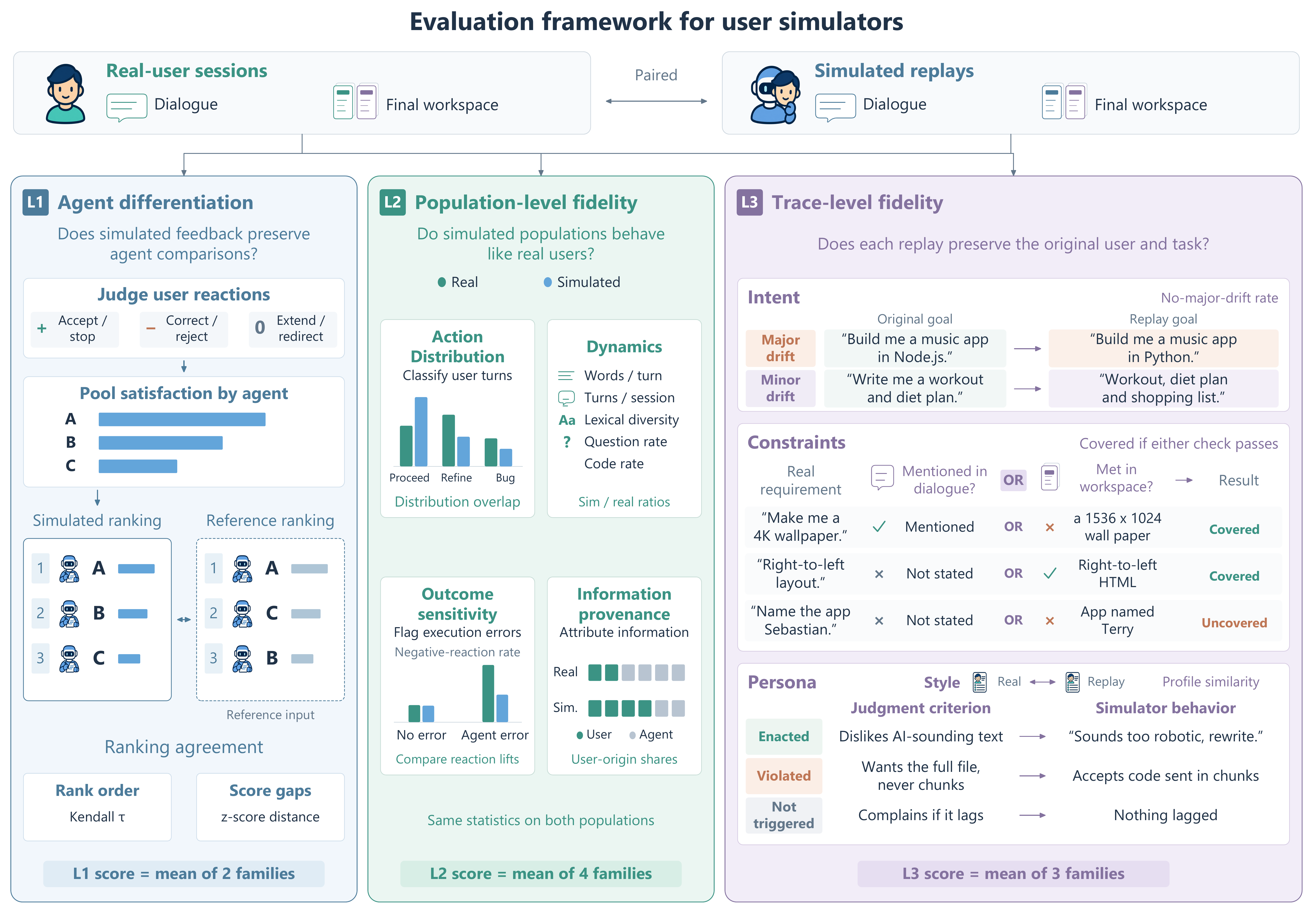}
\caption{Overview of the evaluation framework. L1 compares simulator-derived agent rankings with a reference ranking; L2 compares behavioral statistics of simulated and real-user populations; L3 assesses intent, constraints and persona using conversations and workspace evidence. Examples and plots are illustrative.}
\label{fig:evaluation-framework-implementation}
\label{fig:evaluation-suite}
\end{figure}

Figure~\ref{fig:evaluation-framework-implementation} summarizes the framework's three comparisons. L1 uses simulated user feedback to rank agents and compares the resulting ranking with an independent reference ranking. L2 compares behavioral statistics across simulated and real-user populations. L3 checks whether each replay is faithful to its source user's intent, constraints, and persona, using dialogue and available workspace evidence. We report these three component scores and combine them into an overall evaluation suite score. Appendix~\ref{sec:eval-implementation} provides implementation details, including dialogue-act definitions and score calculations.

\subsection{Downstream ranking agreement (L1)}



Downstream ranking agreement (L1) measures whether evaluating agents with a user simulator produces a ranking consistent with a supplied reference ranking. A judge model (Claude Sonnet 4.6) labels user turns after the opener as accept, extend, correct, redirect, or reject, with stop recorded at session end. For replay $x$, we compute the satisfaction total
\begin{equation}
\begin{aligned}
s(x) ={}& N_{\mathrm{accept}}(x) + N_{\mathrm{stop}}(x) \\
        &- N_{\mathrm{correct}}(x) - N_{\mathrm{reject}}(x),
\end{aligned}
\label{eq:replay-satisfaction}
\end{equation}
where $N_r(x)$ counts occurrences of reaction $r$. Extend and redirect contribute zero. We aggregate reaction counts by agent and normalize by the number of classifiable reactions plus stop labels to obtain each agent's satisfaction score. We rank agents by these scores to obtain a ranking $\hat{\pi}$, then measure its agreement with the supplied reference ranking $\pi^{\mathrm{ref}}$ using Kendall's $\tau$.

The reference is an independent input chosen for the evaluation setting of interest. It could come from human comparisons in the target setting or an existing leaderboard with matching scope. Its construction is outside the contribution of this framework. Our application supplies a reference ranking derived from real-user reactions labeled by the same Claude Sonnet 4.6 judge.

The 0--100 L1 score combines ranking order and standardized score-gap agreement; Appendix~\ref{sec:l1-implementation} provides more detailed implementation.
\subsection{Population-level behavioral fidelity (L2)}


Population-level behavioral fidelity (L2) measures how closely simulated users reproduce real-user behavior across sessions. We compute the same behavioral statistics on the simulated population and the real-user reference population, then measure their similarity using 4 metric families:

\begin{enumerate}[leftmargin=*]
    \item \textbf{Action distribution.}
    An LLM judge classifies user turns into dialogue acts, such as refining a deliverable, reporting a bug, adding a subtask, or asking the agent to proceed (taxonomy in Appendix~\ref{sec:action-taxonomy}). We exclude the copied opening request and renormalize the remaining act frequencies:
    \begin{equation}
        S_{\mathrm{act}}
        = 100 \sum_a \min\left(p_a^{\mathrm{sim}}, p_a^{\mathrm{real}}\right),
    \end{equation}
    where $p_a$ is the proportion of turns assigned to act $a$.

    \item \textbf{Interaction dynamics.}
    We compute 5 statistics: words per user turn, user turns per session, lexical diversity, question-turn rate, and code-turn rate. Lexical diversity uses a moving-average type-token ratio~\citep{covington2010mattr} with a 50-token window. Appendix~\ref{sec:l2-implementation} provides the extraction details. For each statistic, let $r_k$ be the simulated-to-real ratio. The score is
    \begin{equation}
        S_{\mathrm{dyn}}
        = \frac{100}{5}\sum_{k=1}^{5}\min(r_k,1/r_k).
    \end{equation}

    \item \textbf{Outcome sensitivity.}
    We measure how user feedback changes following agent errors and compare this response pattern between simulated and real users. We flag agent turns containing execution error and use the next user turn's dialogue-act label to determine the response's valence. Appendix~\ref{sec:l2-implementation} describes the error indicators and label mapping. 

    Let $E_t$ indicate whether agent turn $t$ contains a detected execution error, and let $R_{t+1}$ denote the following user response's valence.
Only responses in $\mathcal V=\{\mathrm{negative},\mathrm{positive}\}$ enter the conditional probabilities. For population $q \in \{\mathrm{sim},\mathrm{real}\}$, we compute
    \begin{equation}
    \begin{aligned}
    \Delta_q ={}&
    \Pr_q(R_{t+1}=\mathrm{negative} \mid E_t=1,\ R_{t+1}\in\mathcal V) \\
    &- \Pr_q(R_{t+1}=\mathrm{negative} \mid E_t=0,\ R_{t+1}\in\mathcal V).
    \end{aligned}
    \end{equation}
    Here, $E_t=1$ denotes a detected error and $E_t=0$ denotes no detected error. The score is
    \begin{equation}
    S_{\mathrm{out}}
    =100\left(1-\frac{|\Delta_{\mathrm{sim}}-\Delta_{\mathrm{real}}|}{2}\right).
    \end{equation}

    \item \textbf{Information provenance.}
    We measure whether simulated users contribute a similar share of task information as real users. Over-helpful simulators may volunteer diagnoses or decisions that the agent should work out, making its task easier. Conversely, simulators that provide too little information may withhold requirements or clarifications that real users would supply. We therefore compare how information is divided between the user and agent in simulated and real interactions.
    
    An LLM judge identifies key information points in each conversation and attributes them to the user or agent. Appendix~\ref{sec:l2-implementation} describes the information categories and attribution procedure. For population $q \in \{\mathrm{sim},\mathrm{real}\}$, let $u_q$ denote the proportion of information points originating from the user. We compare the user-to-agent information odds:
    \begin{equation}
        \rho
        = \frac{u_{\mathrm{sim}}/(1-u_{\mathrm{sim}})}
               {u_{\mathrm{real}}/(1-u_{\mathrm{real}})}.
    \end{equation}
    The score is
    \begin{equation}
        S_{\mathrm{prov}}
        = 100\,\min\left\{\rho,\frac{1}{\rho}\right\}.
    \end{equation}
    Equal user-origin shares receive a score of 100; shifts toward either greater or lower user contribution reduce the score.
\end{enumerate} 

The aggregated L2 score is the arithmetic mean of the 4 family scores. Higher scores indicate closer agreement with the measured real-user behavior.
\subsection{Trace-level fidelity (L3)}

Trace-level fidelity (L3) measures whether each simulated replay preserves the original user's goal, requirements and persona while responding to the agent's actions. We evaluate each replay against its real-user session using 3 metric families:

\begin{enumerate}[leftmargin=*]
    \item \textbf{Intent preservation.}
    We check whether the simulated user continues pursuing the original goal. An LLM judge compares the real and simulated conversations, using workspace evidence to understand what was produced, and classifies the replay as having no drift, minor drift or major drift. Major drift includes replacing the original goal or abandoning it before an attempt. An agent's failure to complete the requested task does not itself constitute user intent drift. The intent score is the proportion of evaluated replays without major drift.

    \item \textbf{Constraint preservation.}
    We check whether the simulated user maintains the real user's explicit requirements. An LLM judge first extracts hard constraints from the real user's messages, such as output language, deliverable format and required features. It then checks each constraint against the simulated conversation and final workspace, recording whether the user retained it and whether the deliverable satisfied it.

    \item \textbf{Persona consistency.}
    We check whether the simulator preserves both how the user communicates and how they judge the agent's work. Style similarity is assessed by comparing profiles extracted from the real and simulated sessions. For judgment criteria in the application, an LLM judge checks whether the simulated user's decisions follow the original user's criteria, labeling each criterion as enacted, violated or not triggered. Criteria conformance is
    \begin{equation}
        C = \frac{N_{\mathrm{enacted}}}
                 {N_{\mathrm{enacted}} + N_{\mathrm{violated}}}.
    \end{equation}
    The persona score averages style similarity and criteria conformance; criteria that were not triggered are excluded from the conformance calculation.
\end{enumerate}

The L3 score is the arithmetic mean of the 3 family scores, expressed on the same scale. Appendix~\ref{sec:l3-implementation} provides the judge definitions and implementation details.

\subsection{Score aggregation}
We summarize simulator performance while keeping each evaluation level visible. The current evaluation suite reports each level on a 0--100 scale and uses the geometric mean of L1, L2 and L3 as its overall evaluation suite score. We report all three levels alongside the overall evaluation suite score because an aggregate can conceal different failure modes.

\section{Experimental Results}
\label{sec:experimental-results}
We first report the meta-evaluation results, then apply our evaluation framework to existing simulators.

\subsection{Meta-evaluation results}
\label{sec:metaeval-results}
Using the protocol in Section~\ref{sec:metaeval}, we first examine sensitivity to perturbations and then what survives score-based filtering.

\subsubsection{Sensitivity to perturbations}
\label{sec:sensitivity-overview}
\begin{figure}[!htbp]
\centering
\includegraphics[width=\linewidth]{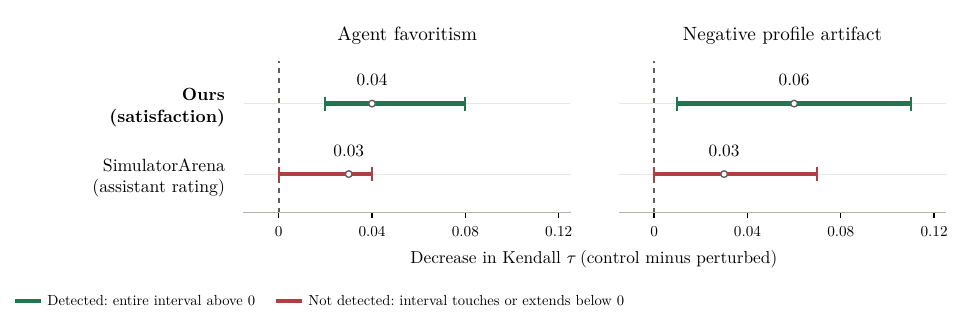}
\caption{Sensitivity to agent favoritism and negative profile artifacts. Labeled points show the decrease in ranking agreement (control minus perturbed), with bootstrap 95\% confidence intervals.}
\label{fig:sensitivity-agent}
\end{figure}

\begin{figure}[!htbp]
\centering
\includegraphics[width=\linewidth]{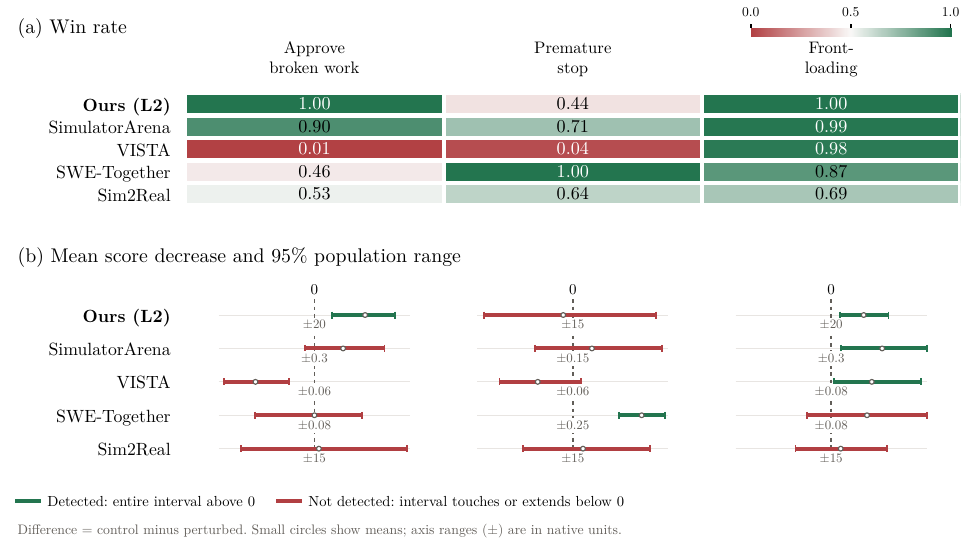}
\caption{Population-level sensitivity of L2 and external evaluation suites using each row's best metric per perturbation: (a) control-over-perturbed win rates; (b) mean score decreases and 2.5th--97.5th percentile ranges.}
\label{fig:sensitivity-population}
\end{figure}

\begin{figure}[!htbp]
\centering
\includegraphics[width=\linewidth]{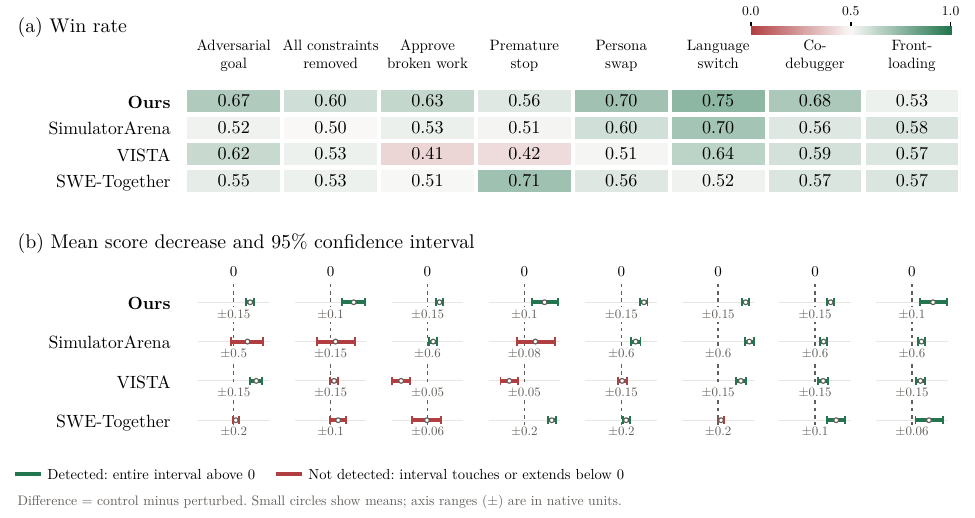}
\caption{Trace-level sensitivity using each evaluation suite's best metric per perturbation: (a) paired control-over-perturbed win rates; (b) mean score decreases and bootstrap 95\% confidence intervals.}
\label{fig:sensitivity-trace}
\end{figure}

We examine which evaluation metrics detect perturbations affecting agent differentiation, population-level fidelity and trace-level fidelity. we consider a
perturbation successfully detected when the 95\% confidence interval for the paired score decrease lies entirely above zero. For population-level scores, we require the 2.5th percentile of the paired score decreases to exceed zero.
We color the corresponding cells green when these criteria are met and red otherwise. Figures~\ref{fig:sensitivity-agent}--\ref{fig:sensitivity-trace} show the compact metric summaries; Appendix~\ref{sec:appendix} provides the detailed results.

Across the 10 perturbations, overall detection coverage is \textbf{10/10} for our evaluation suite, \textbf{6/10} for SimulatorArena, \textbf{5/10} for VISTA, \textbf{6/10} for SWE-Together and \textbf{0/10} for Sim2Real. Sim2Real's behavioral index is defined over a population of conversations, so we test it at the population level for all 10 perturbations (Appendix Tables~\ref{tab:agent-differentiation-sensitivity-full} and~\ref{tab:sensitivity-set-diff}). Coverage summarizes the signals available across an evaluation suite; the filtering analyses separately assess individual scores and fixed composites.

\paragraph{Sensitivity to perturbations affecting agent differentiation}
\label{sec:sensitivity-agent}
We test whether metrics detect feedback biases that distort comparisons between agents.

Our ranking-agreement metric detects both agent favoritism and
negative profile artifacts, with decreases of 0.04 and 0.06,
respectively (Figure~\ref{fig:sensitivity-agent}).
SimulatorArena shows a decrease of 0.03 for each perturbation,
but both confidence intervals include zero. 

Negative profile artifacts also produce behavioral signals: the L2 act-distribution metric has a population-level win rate of 0.99 (Appendix Table~\ref{tab:agent-differentiation-sensitivity-full}), the persona family detects the perturbation at the trace level, and SimulatorArena, VISTA and SWE-Together each have at least one sensitive trace-level metric. These results show how feedback biases can register through their effects on agent comparisons and on simulated behavior.

\FloatBarrier
\paragraph{Sensitivity to perturbations affecting population-level fidelity}
\label{sec:sensitivity-population}
We test whether metrics detect changes in behavioral patterns across populations of replays.

The L2 act-distribution metric detects approve-broken-work and front-loading with population-level win rates of 1.00 in both cases (Figure~\ref{fig:sensitivity-population}). SimulatorArena and VISTA also detect front-loading, with win rates of 0.99 and 0.98 for their best metrics. Premature stop produces a different pattern: no L2 family detects it, whereas SWE-Together's coverage metric reaches a population-level win rate of 1.00. At the trace level, our constraints metric detects premature stop with a paired win rate of 0.56 (Figure~\ref{fig:sensitivity-trace}).
Appendix Table~\ref{tab:metaeval-l2-setsize} examines how population sensitivity changes with population size.

\FloatBarrier
\paragraph{Sensitivity to perturbations affecting trace-level fidelity}
\label{sec:sensitivity-trace}
We test sensitivity to changes in a user's goal, constraints, persona, language and role within a replay.

The constraints metric detects removed constraints; the persona metric detects persona swaps, language switches and co-debugger behavior (Figure~\ref{fig:sensitivity-trace}). SimulatorArena's metrics miss altered goals and removed constraints, VISTA's miss removed constraints and persona swaps, and SWE-Together's miss altered goals and language switches.

Aggregation can amplify these differences: for adversarial goal, the L3 mean rises from a paired win rate of 0.70 to a population-level win rate of 1.00 (Appendix Table~\ref{tab:sensitivity-set}).

\FloatBarrier
\subsubsection{Which simulations receive high scores?}
\label{sec:filtering-rationale}
We examine what survives filtering to understand what a high evaluation score signals. The following analyses use the filtering protocol in Section~\ref{sec:meta-evaluation-protocol}.Tables~\ref{tab:best-filtering-agent}--\ref{tab:best-filtering-trace} report the best AUROC for each evaluation suite and condition.

\paragraph{Which simulations score highly on agent differentiation?}
\label{sec:l1-filter}
We examine whether high ranking-agreement scores select control populations over populations containing agent favoritism or negative profile artifacts.

\begin{table}[!htbp]
\centering
\fontsize{9.5}{12}\selectfont
\setlength{\tabcolsep}{0pt}
\renewcommand{\arraystretch}{1.12}
\begin{tabular}{@{}p{78pt}@{\hspace{8pt}}>{\centering\arraybackslash}p{\dimexpr(\linewidth-98pt)/8\relax}>{\centering\arraybackslash}p{\dimexpr(\linewidth-98pt)/8\relax}>{\centering\arraybackslash}p{\dimexpr(\linewidth-98pt)/8\relax}>{\centering\arraybackslash}p{\dimexpr(\linewidth-98pt)/8\relax}@{\hspace{12pt}}>{\centering\arraybackslash}p{\dimexpr(\linewidth-98pt)/8\relax}>{\centering\arraybackslash}p{\dimexpr(\linewidth-98pt)/8\relax}>{\centering\arraybackslash}p{\dimexpr(\linewidth-98pt)/8\relax}>{\centering\arraybackslash}p{\dimexpr(\linewidth-98pt)/8\relax}@{}}
\toprule
Evaluation suite & \multicolumn{4}{c}{\textbf{Agent favoritism}} & \multicolumn{4}{c}{\textbf{Negative profile artifact}} \\
\cmidrule(lr){2-5} \cmidrule(lr){6-9}
 & 10\% & 25\% & 50\% & 100\% & 10\% & 25\% & 50\% & 100\% \\ \midrule
Ours & \textbf{0.55}\textsuperscript{\normalfont\fontsize{6}{6}\selectfont R} & \textbf{0.69}\textsuperscript{\normalfont\fontsize{6}{6}\selectfont R} & \textbf{0.84}\textsuperscript{\normalfont\fontsize{6}{6}\selectfont R} & \textbf{0.99}\textsuperscript{\normalfont\fontsize{6}{6}\selectfont R} & 0.62\textsuperscript{\normalfont\fontsize{6}{6}\selectfont A} & 0.75\textsuperscript{\normalfont\fontsize{6}{6}\selectfont A} & 0.91\textsuperscript{\normalfont\fontsize{6}{6}\selectfont A} & \textbf{1.00}\textsuperscript{\normalfont\fontsize{6}{6}\selectfont A} \\
SimulatorArena & 0.50\textsuperscript{\normalfont\fontsize{6}{6}\selectfont W} & 0.54\textsuperscript{\normalfont\fontsize{6}{6}\selectfont W} & 0.56\textsuperscript{\normalfont\fontsize{6}{6}\selectfont W} & 0.64\textsuperscript{\normalfont\fontsize{6}{6}\selectfont W} & 0.55\textsuperscript{\normalfont\fontsize{6}{6}\selectfont I} & 0.60\textsuperscript{\normalfont\fontsize{6}{6}\selectfont I} & 0.67\textsuperscript{\normalfont\fontsize{6}{6}\selectfont I} & 0.84\textsuperscript{\normalfont\fontsize{6}{6}\selectfont I} \\
VISTA & 0.49\textsuperscript{\normalfont\fontsize{6}{6}\selectfont K} & 0.50\textsuperscript{\normalfont\fontsize{6}{6}\selectfont F} & 0.50\textsuperscript{\normalfont\fontsize{6}{6}\selectfont F} & 0.50\textsuperscript{\normalfont\fontsize{6}{6}\selectfont F} & 0.60\textsuperscript{\normalfont\fontsize{6}{6}\selectfont F} & 0.71\textsuperscript{\normalfont\fontsize{6}{6}\selectfont F} & 0.86\textsuperscript{\normalfont\fontsize{6}{6}\selectfont F} & 0.98\textsuperscript{\normalfont\fontsize{6}{6}\selectfont F} \\
SWE-Together & 0.51\textsuperscript{\normalfont\fontsize{6}{6}\selectfont Q} & 0.49\textsuperscript{\normalfont\fontsize{6}{6}\selectfont S} & 0.49\textsuperscript{\normalfont\fontsize{6}{6}\selectfont S} & 0.44\textsuperscript{\normalfont\fontsize{6}{6}\selectfont S} & \textbf{0.67}\textsuperscript{\normalfont\fontsize{6}{6}\selectfont U} & \textbf{0.85}\textsuperscript{\normalfont\fontsize{6}{6}\selectfont U} & \textbf{0.98}\textsuperscript{\normalfont\fontsize{6}{6}\selectfont U} & \textbf{1.00}\textsuperscript{\normalfont\fontsize{6}{6}\selectfont U} \\
Sim2Real & 0.51 & 0.53 & 0.55 & 0.58 & 0.50 & 0.51 & 0.56 & 0.61 \\
\bottomrule
\end{tabular}
\caption{Filtering on agent favoritism and negative profile artifacts. Highest AUROC at each replacement fraction; bold marks the column maximum.}
\label{tab:best-filtering-agent}
\par\vspace{4pt}
{\fontsize{8}{9.7}\selectfont\raggedright\textit{Selected metrics.} Ours: R ranking agreement, A action distribution; SimulatorArena: W Likert writing, I Likert interaction; VISTA: K knowledge consistency, F profile consistency; SWE-Together: Q weighted coverage, S scope precision, U user correction. Sim2Real uses its behavioral index.\par}
\end{table}

For agent favoritism, L1's AUROC increases from 0.55 at 10\% replacement to 0.84 at 50\% and 0.99 at 100\%. The strongest external metric reaches only 0.56 and 0.64 at 50\% and 100\% replacement, respectively (Table~\ref{tab:best-filtering-agent}).

For negative profile artifacts, filtering improves as the perturbed fraction increases.
At 25\% replacement, the L2 act-distribution metric, VISTA's profile-consistency metric and SWE-Together's user-correction metric achieve AUROCs of 0.75, 0.71 and 0.85, respectively.
At 100\% replacement, these behavioral metrics reach 0.98--1.00 AUROC (Table~\ref{tab:best-filtering-agent}).
These results illustrate complementary signals: L1 provides the strongest filtering of agent favoritism, while the act-distribution and external behavioral metrics identify negative profile artifacts.

\paragraph{Which simulations receive high population-level fidelity scores?}
\label{sec:l2-filter}
We examine how much perturbed behavior a population can contain while still receiving high fidelity scores (Table~\ref{tab:best-filtering-population}).

\begin{table}[!htbp]
\centering
\fontsize{9.5}{12}\selectfont
\setlength{\tabcolsep}{0pt}
\renewcommand{\arraystretch}{1.12}
\begin{tabular}{@{}p{78pt}@{\hspace{8pt}}>{\centering\arraybackslash}p{\dimexpr(\linewidth-110pt)/12\relax}>{\centering\arraybackslash}p{\dimexpr(\linewidth-110pt)/12\relax}>{\centering\arraybackslash}p{\dimexpr(\linewidth-110pt)/12\relax}>{\centering\arraybackslash}p{\dimexpr(\linewidth-110pt)/12\relax}@{\hspace{12pt}}>{\centering\arraybackslash}p{\dimexpr(\linewidth-110pt)/12\relax}>{\centering\arraybackslash}p{\dimexpr(\linewidth-110pt)/12\relax}>{\centering\arraybackslash}p{\dimexpr(\linewidth-110pt)/12\relax}>{\centering\arraybackslash}p{\dimexpr(\linewidth-110pt)/12\relax}@{\hspace{12pt}}>{\centering\arraybackslash}p{\dimexpr(\linewidth-110pt)/12\relax}>{\centering\arraybackslash}p{\dimexpr(\linewidth-110pt)/12\relax}>{\centering\arraybackslash}p{\dimexpr(\linewidth-110pt)/12\relax}>{\centering\arraybackslash}p{\dimexpr(\linewidth-110pt)/12\relax}@{}}
\toprule
Evaluation suite & \multicolumn{4}{c}{\textbf{Approve broken work}} & \multicolumn{4}{c}{\textbf{Premature stop}} & \multicolumn{4}{c}{\textbf{Front-loading}} \\
\cmidrule(lr){2-5} \cmidrule(lr){6-9} \cmidrule(lr){10-13}
 & 10\% & 25\% & 50\% & 100\% & 10\% & 25\% & 50\% & 100\% & 10\% & 25\% & 50\% & 100\% \\ \midrule
Ours, L2 & 0.49\textsuperscript{\normalfont\fontsize{6}{6}\selectfont D} & 0.46\textsuperscript{\normalfont\fontsize{6}{6}\selectfont D} & 0.45\textsuperscript{\normalfont\fontsize{6}{6}\selectfont P} & \textbf{1.00}\textsuperscript{\normalfont\fontsize{6}{6}\selectfont A} & 0.49\textsuperscript{\normalfont\fontsize{6}{6}\selectfont P} & 0.49\textsuperscript{\normalfont\fontsize{6}{6}\selectfont P} & 0.49\textsuperscript{\normalfont\fontsize{6}{6}\selectfont P} & 0.46\textsuperscript{\normalfont\fontsize{6}{6}\selectfont P} & \textbf{0.56}\textsuperscript{\normalfont\fontsize{6}{6}\selectfont A} & \textbf{0.65}\textsuperscript{\normalfont\fontsize{6}{6}\selectfont A} & \textbf{0.78}\textsuperscript{\normalfont\fontsize{6}{6}\selectfont A} & \textbf{0.96}\textsuperscript{\normalfont\fontsize{6}{6}\selectfont A} \\
SimulatorArena & \textbf{0.53}\textsuperscript{\normalfont\fontsize{6}{6}\selectfont W} & \textbf{0.58}\textsuperscript{\normalfont\fontsize{6}{6}\selectfont W} & \textbf{0.66}\textsuperscript{\normalfont\fontsize{6}{6}\selectfont W} & 0.78\textsuperscript{\normalfont\fontsize{6}{6}\selectfont W} & 0.53\textsuperscript{\normalfont\fontsize{6}{6}\selectfont T} & 0.58\textsuperscript{\normalfont\fontsize{6}{6}\selectfont T} & 0.62\textsuperscript{\normalfont\fontsize{6}{6}\selectfont T} & 0.69\textsuperscript{\normalfont\fontsize{6}{6}\selectfont T} & \textbf{0.56}\textsuperscript{\normalfont\fontsize{6}{6}\selectfont W} & 0.64\textsuperscript{\normalfont\fontsize{6}{6}\selectfont W} & 0.74\textsuperscript{\normalfont\fontsize{6}{6}\selectfont W} & 0.90\textsuperscript{\normalfont\fontsize{6}{6}\selectfont W} \\
VISTA & 0.43\textsuperscript{\normalfont\fontsize{6}{6}\selectfont K} & 0.31\textsuperscript{\normalfont\fontsize{6}{6}\selectfont K} & 0.16\textsuperscript{\normalfont\fontsize{6}{6}\selectfont K} & 0.01\textsuperscript{\normalfont\fontsize{6}{6}\selectfont K} & 0.45\textsuperscript{\normalfont\fontsize{6}{6}\selectfont G} & 0.38\textsuperscript{\normalfont\fontsize{6}{6}\selectfont G} & 0.27\textsuperscript{\normalfont\fontsize{6}{6}\selectfont G} & 0.12\textsuperscript{\normalfont\fontsize{6}{6}\selectfont G} & \textbf{0.56}\textsuperscript{\normalfont\fontsize{6}{6}\selectfont F} & 0.64\textsuperscript{\normalfont\fontsize{6}{6}\selectfont F} & 0.77\textsuperscript{\normalfont\fontsize{6}{6}\selectfont F} & 0.92\textsuperscript{\normalfont\fontsize{6}{6}\selectfont F} \\
SWE-Together & 0.50\textsuperscript{\normalfont\fontsize{6}{6}\selectfont S} & 0.50\textsuperscript{\normalfont\fontsize{6}{6}\selectfont S} & 0.52\textsuperscript{\normalfont\fontsize{6}{6}\selectfont S} & 0.50\textsuperscript{\normalfont\fontsize{6}{6}\selectfont S} & \textbf{0.69}\textsuperscript{\normalfont\fontsize{6}{6}\selectfont Q} & \textbf{0.88}\textsuperscript{\normalfont\fontsize{6}{6}\selectfont Q} & \textbf{0.99}\textsuperscript{\normalfont\fontsize{6}{6}\selectfont Q} & \textbf{1.00}\textsuperscript{\normalfont\fontsize{6}{6}\selectfont C} & 0.53\textsuperscript{\normalfont\fontsize{6}{6}\selectfont C} & 0.59\textsuperscript{\normalfont\fontsize{6}{6}\selectfont C} & 0.65\textsuperscript{\normalfont\fontsize{6}{6}\selectfont C} & 0.80\textsuperscript{\normalfont\fontsize{6}{6}\selectfont S} \\
Sim2Real & 0.49 & 0.48 & 0.45 & 0.55 & 0.48 & 0.45 & 0.44 & 0.60 & 0.50 & 0.52 & 0.56 & 0.61 \\
\bottomrule
\end{tabular}
\caption{Population-level filtering. Highest AUROC at each replacement fraction; bold marks the column maximum.}
\label{tab:best-filtering-population}
\par\vspace{4pt}
{\fontsize{8}{9.7}\selectfont\raggedright\textit{Selected metrics.} Ours: D dynamics, P information provenance, A action distribution; SimulatorArena: W writing, T Turing; VISTA: K knowledge consistency, G goal consistency, F profile consistency; SWE-Together: Q weighted coverage, C coverage, S scope precision. Sim2Real uses its behavioral index.\par}
\end{table}

For front-loading, the act-distribution AUROC increases from 0.56 at 10\% replacement to 0.96 at 100\% replacement. At full replacement, the best metrics from SimulatorArena, VISTA and SWE-Together reach AUROCs of 0.80--0.92 (Table~\ref{tab:best-filtering-population}).

For approve-broken-work, our best L2 AUROC remains below 0.5 at 10--50\% replacement, but act distribution reaches 1.00 at full replacement. SimulatorArena's writing metric provides stronger separation at partial replacement and reaches 0.78 at full replacement, compared with 0.01 for VISTA and 0.50 for SWE-Together (Table~\ref{tab:best-filtering-population}). 

For premature stop, SWE-Together's highest-AUROC metric increases from 0.69 at 10\% replacement to 1.00 at full replacement, while our best L2 metric remains below 0.5 throughout(Table~\ref{tab:best-filtering-population}).  Here, the control simulator produces more user turns than
real users. Truncation brings session lengths and dialogue-act
distributions closer to the real-user reference, raising the
corresponding L2 scores despite premature termination.


\paragraph{Which simulations receive high trace-level fidelity scores?}
\label{sec:l3-filter}
Pooling the 5 trace-level fidelity perturbations, we test which control replays and real-user traces survive filtering alongside perturbed replays.

\begin{table}[!htbp]
\centering
\small
\setlength{\tabcolsep}{4pt}
\resizebox{\textwidth}{!}{%
\begin{tabular}{lcccccc}
\toprule
 & \multicolumn{3}{c}{Control vs perturbed (control positive) $\uparrow$} & \multicolumn{3}{c}{Real vs perturbed (real positive) $\uparrow$} \\
\cmidrule(lr){2-4}\cmidrule(lr){5-7}
Benchmark (best metric) & AUROC & control@5\%FPR & control@10\%FPR & AUROC & real@5\%FPR & real@10\%FPR \\
\midrule
Ours & \textbf{0.62}\,{\tiny persona} & \textbf{0.10} & \textbf{0.17} & \textbf{0.90}\,{\tiny persona} & \textbf{0.68} & \textbf{0.77} \\
SimulatorArena & 0.55\,{\tiny Likert writing} & 0.05 & 0.11 & 0.88\,{\tiny Likert writing} & 0.28 & 0.55 \\
VISTA & 0.55\,{\tiny composite} & 0.06 & 0.11 & 0.68\,{\tiny composite} & 0.09 & 0.19 \\
SWE-Together & 0.52\,{\tiny coverage rate} & 0.05 & 0.11 & 0.75\,{\tiny user correction} & 0.12 & 0.24 \\
\bottomrule
\end{tabular}}
\caption{Trace-level filtering using each evaluation suite's highest-AUROC metric for each comparison. Columns report AUROC and positive-trace retention at 5\% and 10\% perturbed survival, using control replays (left) or real-user traces (right) as positives. Bold marks the highest value in each column. Full results: Appendix Table~\ref{tab:l3-filtering-pooled}.}
\label{tab:best-filtering-trace}
\end{table}

With control replays as positives, the persona family achieves an AUROC of 0.62, compared with 0.55 for SimulatorArena and VISTA and 0.52 for SWE-Together. At 5\% and 10\% perturbed survival, it retains 10\% and 17\% of controls, compared with 5--6\% and 11\%, respectively, for the external evaluation suites' highest-AUROC metrics (Table~\ref{tab:best-filtering-trace}). These estimates suggest modest discrimination, with substantial loss of control replays under strict filtering.

With real-user traces as positives, the persona family reaches
0.90 AUROC and retains 77\% of real-user traces at 10\% perturbed survival. SimulatorArena's Likert writing
judge reaches 0.88 AUROC, with retention of 55\%. These results complement the control-replay
comparison: the persona metric favors simulations without
the injected defects and also preserves the genuine user
behavior that simulation aims to reproduce.


Appendix~\ref{sec:metaeval-examples} illustrates a goal change that receives high consistency and intent-recall scores.

\iffullpaper
\subsection{Application of the Evaluation Framework}
\label{sec:application}

We apply the evaluation suite discussed in Section~\ref{sec:suite} to existing simulators to compare their strengths and identify recurring behavioral weaknesses.

\subsubsection{Evaluation setup}
We evaluate six simulators on the same 351 real sessions. SimulatorArena and $\tau^2$-bench use their published simulation recipes, while the Sonnet~5, Terra and Luna baselines use profile-conditioned simulation. MIMESIS-9B~\citep{phan2026mimesis} is a trained user simulator that receives a profile-derived goal and the agent's text. We use the same set of 9 agents for all L1 comparisons. The L1 reference comes from real-user satisfaction scored by the same reactions judge on about 1,000 separate real sessions per agent. All four L2 families use real-user behavior from the same 351 evaluation sessions as their reference. Here we compare simulation strategies and backbones, whereas the meta-evaluation in Section~\ref{sec:replay-protocol} holds the simulator backbone fixed to study controlled perturbations. Table~\ref{tab:application} reports the component and overall scores.

\subsubsection{Comparison of existing simulators}

\begin{table}[!htbp]
\centering
\small
\setlength{\tabcolsep}{4pt}
\begin{tabular*}{\linewidth}{@{\extracolsep{\fill}}lcccc@{}}
\toprule
Simulator & L1 $\uparrow$ & L2 $\uparrow$ & L3 $\uparrow$ & Headline $\uparrow$ \\
\midrule
Ours, Sonnet~5 backbone & 47 & 87 & 83 & \textbf{70} \\
SimulatorArena (gpt-5-mini) & 57 & 79 & 74 & 69 \\
Ours, Terra backbone & 42 & 73 & 84 & 64 \\
Ours, Luna backbone & 40 & 72 & 78 & 61 \\
$\tau^2$-bench (gpt-4.1) & 33 & 82 & 74 & 59 \\
MIMESIS-9B & 27 & 82 & 61 & 51 \\
\bottomrule
\end{tabular*}
\caption{Six user simulators evaluated with the evaluation suite on the same 351 real sessions. L1 uses a reference derived from real-user satisfaction for the same 9 agents. L1--L3 are 0--100 component scores, and the headline is their geometric mean.}
\label{tab:application}
\end{table}

Sonnet~5 has the highest overall score (70) and L2 score (87). Terra has the highest L3 score (84), followed by Sonnet~5 (83). SimulatorArena has the highest L1 score (57) and Kendall agreement ($\tau=0.67$), compared with 0.44 for Sonnet~5.

\textbf{Luna and $\tau^2$-bench have different behavioral weaknesses.} $\tau^2$-bench has higher population-level fidelity than the Luna baseline (82 versus 72), whereas Luna has higher trace-level fidelity (78 versus 74). These results identify different priorities for improvement: population-level behavior for Luna, and fidelity within individual replays for $\tau^2$-bench.

Appendix~\ref{sec:application-examples} provides qualitative examples of excess technical assistance and relaxed acceptance criteria.

\FloatBarrier

\section{Conclusion}
User-simulator evaluation benefits from behavioral diagnostics and a separate check of the intended agent-evaluation application. The perturbation results show that a metric's sensitivity does not guarantee useful filtering, while the favoritism perturbations show that substantial changes in selected agents' scores and ranks can coexist with weak responses from the available trace diagnostics. We therefore report fidelity, per-agent effects and agreement with the supplied ranking together, with each claim tied to its evaluation protocol and evidence.

\fi

\bibliographystyle{iclr2027_conference}
\bibliography{custom}

@article{barres2025tau2,
  title = {{$\tau^2$-Bench}: Evaluating Conversational Agents in a Dual-Control
          Environment},
  author = {Barres, Victor and Dong, Honghua and Ray, Soham and Si, Xujie and
           Narasimhan, Karthik},
  journal = {arXiv preprint arXiv:2506.07982},
  year = {2025}
}

@inproceedings{dou2025simulatorarena,
  title = {{SimulatorArena}: Are User Simulators Reliable Proxies for Multi-Turn
          Evaluation of {AI} Assistants?},
  author = {Dou, Yao and Galley, Michel and Peng, Baolin and Kedzie, Chris and Cai,
           Weixin and Ritter, Alan and Quirk, Chris and Xu, Wei and Gao, Jianfeng},
  booktitle = {Proceedings of the 2025 Conference on Empirical Methods in Natural
              Language Processing ({EMNLP})},
  year = {2025}
}

@article{wu2026swetogether,
  title = {{SWE-Together}: Evaluating Coding Agents in Interactive User Sessions},
  author = {Wu, Yifan and Zhao, Zhuokai and Li, Songlin and Lee, Ho Hin and Zhu,
           Jiacheng and Wu, Shirley and Yu, Tianhe and Li, Serena and Zhang, Lizhu and
           Fan, Xiangjun and Li, Shengzhi},
  journal = {arXiv preprint arXiv:2606.29957},
  year = {2026}
}

@article{lu2026vista,
  title = {{VISTA}: A Versatile Interactive User Simulation Toolkit for Agent
          Evaluation},
  author = {Lu, Yunan and Shea, Ryan and Zhang, Yusen and Yu, Zhou},
  journal = {arXiv preprint arXiv:2606.11079},
  year = {2026}
}

@article{zhou2026sim2real,
  title = {Mind the {Sim2Real} Gap in User Simulation for Agentic Tasks},
  author = {Zhou, Xuhui and Sun, Weiwei and Ma, Qianou and Xie, Yiqing and Liu, Jiarui
           and Du, Weihua and Welleck, Sean and Yang, Yiming and Neubig, Graham and Wu,
           Sherry Tongshuang and Sap, Maarten},
  journal = {arXiv preprint arXiv:2603.11245},
  year = {2026}
}

@article{qian2025userrl,
  title = {Userrl: Training interactive user-centric agent via reinforcement learning},
  author = {Qian, Cheng and Liu, Zuxin and Prabhakar, Akshara and Qiu, Jielin and Liu,
           Zhiwei and Chen, Haolin and Kokane, Shirley and Ji, Heng and Yao, Weiran and
           Heinecke, Shelby and others},
  journal = {arXiv preprint arXiv:2509.19736},
  year = {2025}
}

@inproceedings{sai-etal-2021-perturbation,
  title = {Perturbation {C}heck{L}ists for Evaluating {NLG} Evaluation Metrics},
  author = {Sai, Ananya B. and Dixit, Tanay and Sheth, Dev Yashpal and Mohan, Sreyas and
           Khapra, Mitesh M.},
  booktitle = {Proceedings of the 2021 Conference on Empirical Methods in Natural
              Language Processing ({EMNLP})},
  year = {2021}
}

@inproceedings{mathur-etal-2020-tangled,
  title = {Tangled up in {BLEU}: Reevaluating the Evaluation of Automatic Machine
          Translation Evaluation Metrics},
  author = {Mathur, Nitika and Baldwin, Timothy and Cohn, Trevor},
  booktitle = {Proceedings of the 58th Annual Meeting of the Association for
              Computational Linguistics ({ACL})},
  year = {2020}
}

@article{deutsch-etal-2021-statistical,
  title = {A Statistical Analysis of Summarization Evaluation Metrics Using Resampling
          Methods},
  author = {Deutsch, Daniel and Dror, Rotem and Roth, Dan},
  journal = {Transactions of the Association for Computational Linguistics},
  year = {2021}
}

@article{phan2026mimesis,
  title = {{MIMESIS: Learning User Simulators as Training Environments for Interactive
          Agents}},
  author = {Phan, Hoang and Huynh, Dat and Zhmoginov, Andrey and Zeng, Qi and Mu, Wancen
           and Cao, Yue and Bi, Shengjie and He, Yun and Oh, Changdae and Lei, Deren},
  journal = {arXiv preprint arXiv:2610.09484},
  year = {2026}
}

@article{tennenholtz2026controllable,
  title = {Controllable User Simulation},
  author = {Tennenholtz, Guy and Meshi, Ofer and Globerson, Amir and Shalit, Uri and
           Jeong, Jihwan and Boutilier, Craig},
  journal = {arXiv preprint arXiv:2605.11519},
  year = {2026}
}

@inproceedings{naous2026flipping,
  title = {Flipping the Dialogue: Training and Evaluating User Language Models},
  author = {Naous, Tarek and Laban, Philippe and Xu, Wei and Neville, Jennifer},
  booktitle = {International Conference on Learning Representations ({ICLR})},
  year = {2026}
}

@inproceedings{zheng2023judging,
  title = {Judging {LLM-as-a-Judge} with {MT-Bench} and {Chatbot Arena}},
  author = {Zheng, Lianmin and Chiang, Wei-Lin and Sheng, Ying and Zhuang, Siyuan and
           Wu, Zhanghao and Zhuang, Yonghao and Lin, Zi and Li, Zhuohan and Li, Dacheng
           and Xing, Eric P. and Zhang, Hao and Gonzalez, Joseph E. and Stoica, Ion},
  booktitle = {Advances in Neural Information Processing Systems ({NeurIPS})},
  year = {2023}
}

@inproceedings{panickssery2024llm,
  title = {{LLM} Evaluators Recognize and Favor Their Own Generations},
  author = {Panickssery, Arjun and Bowman, Samuel R. and Feng, Shi},
  booktitle = {Advances in Neural Information Processing Systems ({NeurIPS})},
  year = {2024}
}

@article{pietquin2013metrics,
  title = {A survey on metrics for the evaluation of user simulations},
  author = {Pietquin, Olivier and Hastie, Helen},
  journal = {The Knowledge Engineering Review},
  year = {2013}
}

@inproceedings{ni2026survey,
  title = {A Survey on {LLM}-based Conversational User Simulation},
  author = {Ni, Bo and Wang, Leyao and Wang, Yu and Kveton, Branislav and Dernoncourt,
           Franck and Xia, Yu and Chen, Hongjie and Leura, Reuben and Basu, Samyadeep
           and Mukherjee, Subhojyoti and Mathur, Puneet and Ahmed, Nesreen and Wu, Junda
           and Li, Li and Zhang, Huixin and Zhang, Ruiyi and Yu, Tong and Kim, Sungchul
           and Gu, Jiuxiang and Tu, Zhengzhong and Siu, Alexa and Wang, Zichao and Yoon,
           David Seunghyun and Lipka, Nedim and Park, Namyong and Lin, Zihao and Bui,
           Trung and Zhao, Yue and Derr, Tyler and Rossi, Ryan A.},
  booktitle = {Proceedings of the 19th Conference of the European Chapter of the
              Association for Computational Linguistics ({EACL})},
  year = {2026}
}

@article{seshadri2026lost,
  title = {Lost in Simulation: {LLM}-Simulated Users are Unreliable Proxies for Human
          Users in Agentic Evaluations},
  author = {Seshadri, Preethi and Cahyawijaya, Samuel and Odumakinde, Ayomide and Singh,
           Sameer and Goldfarb-Tarrant, Seraphina},
  journal = {arXiv preprint arXiv:2601.17087},
  year = {2026}
}

@article{liu2026disclosure,
  title = {{Disclosure-Gated User Simulation for Companion-Agent Evaluation}},
  author = {Liu, Yao and He, Yu and Huang, Jihao and Liu, Xinyang and Zhou, Feng},
  journal = {arXiv preprint arXiv:2609.00982},
  year = {2026}
}

@article{jain2026userproxybench,
  title = {{UserProxyBench: Evaluating LLM User Simulators for Agent Benchmarks and
          Training}},
  author = {Jain, Ashish and Sandhu, Armaan},
  journal = {arXiv preprint arXiv:2609.38043},
  year = {2026}
}

@article{zheng2026argus,
  title = {{ARGUS: An Agenda-Driven Framework for Reliable User Simulation}},
  author = {Zheng, Hang and Jiang, Nan and Zhu, Zichen and Yang, Zixu and Tang, Zhiyang
           and Fan, Shuai and Chen, Lu and Wu, Xiaobao and Yu, Kai},
  journal = {arXiv preprint arXiv:2610.21346},
  year = {2026}
}

@inproceedings{baumann2026swechat,
  title = {{SWE-chat: Coding Agent Interactions From Real Users in the Wild}},
  author = {Baumann, Joachim and Padmakumar, Vishakh and Li, Xiang and Yang, John and
           Yang, Diyi and Koyejo, Sanmi},
  booktitle = {Third Conference on Language Modeling ({COLM})},
  year = {2026}
}

@article{suh2026utility,
  title = {{Quantifying the Utility of User Simulators for Building Collaborative LLM
          Assistants}},
  author = {Suh, Joseph and Raj, Ayush and Kang, Minwoo and Chang, Serina},
  journal = {arXiv preprint arXiv:2605.09808},
  year = {2026}
}

@article{spiliopoulou2025playfavorites,
  title = {Play Favorites: A Statistical Method to Measure Self-Bias in
          {LLM-as-a-Judge}},
  author = {Spiliopoulou, Evangelia and Fogliato, Riccardo and Burnsky, Hanna and
           Soliman, Tamer and Ma, Jie and Horwood, Graham and Ballesteros, Miguel},
  journal = {arXiv preprint arXiv:2508.06709},
  year = {2025}
}

@article{brehme2026swedive,
  title = {{SWE-DIVE}: Multi-Turn Coding Agent Evaluation with Dynamic User Simulation},
  author = {Brehme, Lorenz and Steenhoek, Benjamin and Breu, Ruth},
  journal = {arXiv preprint arXiv:2610.30047},
  year = {2026}
}

@article{zou2026coworkgym,
  title = {{Cowork-Gym: Synthesizing Interactive Work Environments at Scale}},
  author = {Zou, Anni and Chen, Cheng and Deng, Jiaji and Zhou, Weikang and Mai, Xinji
           and Wu, Yue and Lin, Jianing and Xie, Lipeng and Qin, Lixiong and Zhao,
           Qingyue and Shen, Que and Tao, Shuchang and Pan, Xuchen and Li, Yaliang and
           Zhai, Yunpeng and Chen, Yushuo and Wang, Zhihai and Zhang, Zhuo and Liu,
           Zhaoyang and Ding, Bolin and Zhou, Jingren},
  journal = {arXiv preprint arXiv:2610.17670},
  year = {2026}
}

@article{zhao2026userarena,
  title = {{UserArena: Benchmarking Interactive User Simulation for Long-Horizon
          Shopping Agents}},
  author = {Zhao, Jujia and Tang, Jiakai and Yu, Jing and Chen, Wen and Wu, Jian and
           Zhu, Han and Sun, Fei and Verberne, Suzan and Ren, Zhaochun},
  journal = {arXiv preprint arXiv:2610.30932},
  year = {2026}
}

@article{kantharuban2026cue,
  title = {{CUEing User Simulators: Calibrated User Embeddings for Multi-Turn
          Benchmarking}},
  author = {Kantharuban, Anjali and Mueller, Jonas},
  journal = {arXiv preprint arXiv:2610.02460},
  year = {2026}
}

@article{zerhoudi2026autosim,
  title = {{AutoSim}: A Self-Improving User Simulator for Conversational Search
          Evaluation},
  author = {Zerhoudi, Saber and Mitrovic, Jelena and Granitzer, Michael},
  journal = {arXiv preprint arXiv:2610.28054},
  year = {2026}
}

@article{stufflebeam2001metaevaluation,
  title = {The Metaevaluation Imperative},
  author = {Stufflebeam, Daniel L.},
  journal = {American Journal of Evaluation},
  year = {2001}
}

@article{bodhwani2026gauge,
  title = {{GAUGE: When Not to Trust LLM-as-a-Judge in User-Simulated Evaluation of
          Task-Oriented Agents}},
  author = {Bodhwani, Umesh and Tran, Thanh and Wei, Kai},
  journal = {arXiv preprint arXiv:2609.12191},
  year = {2026}
}

@article{chopra2026personapolicies,
  title = {Beyond Cooperative Simulators: Generating Realistic User Personas for Robust
          Evaluation of {LLM} Agents},
  author = {Chopra, Harshita and Ghate, Kshitish and Caliskan, Aylin and Kohno,
           Tadayoshi and Shah, Chirag and Jaques, Natasha},
  journal = {arXiv preprint arXiv:2605.12894},
  year = {2026}
}

@inproceedings{muller2025grouse,
  title = {{GroUSE}: A Benchmark to Evaluate Evaluators in Grounded Question Answering},
  author = {Muller, Sacha and Loison, Antonio and Omrani, Bilel and Viaud, Gautier},
  booktitle = {Proceedings of the 31st International Conference on Computational
              Linguistics ({COLING})},
  year = {2025}
}

@article{covington2010mattr,
  title = {Cutting the {Gordian} Knot: The Moving-Average Type--Token Ratio ({MATTR})},
  author = {Covington, Michael A. and McFall, Joe D.},
  journal = {Journal of Quantitative Linguistics},
  year = {2010}
}

@article{kimiteam2026k25,
  title = {{Kimi K2.5: Visual Agentic Intelligence}},
  author = {{Kimi Team}},
  journal = {arXiv preprint arXiv:2602.02276v2},
  year = {2026}
}

@misc{openai2026gpt54,
  title = {Introducing {GPT-5.4}},
  author = {{OpenAI}},
  howpublished = {Model release},
  url = {https://openai.com/index/introducing-gpt-5-4/},
  year = {2026}
}

@techreport{anthropic2026opus46,
  title = {System Card: {Claude Opus 4.6}},
  author = {{Anthropic}},
  institution = {Anthropic},
  type = {System card},
  url = {https://www-cdn.anthropic.com/0816a4f29610cb87da10f40c6f78b499a7bf722f.pdf},
  year = {2026}
}

@inproceedings{yang2024sweagent,
  title = {{SWE-agent: Agent-Computer Interfaces Enable Automated Software
          Engineering}},
  author = {Yang, John and Jimenez, Carlos E. and Wettig, Alexander and Lieret, Kilian
           and Yao, Shunyu and Narasimhan, Karthik and Press, Ofir},
  booktitle = {Advances in Neural Information Processing Systems ({NeurIPS})},
  year = {2024}
}

@inproceedings{xie2024osworld,
  title = {{OSWorld: Benchmarking Multimodal Agents for Open-Ended Tasks in Real
          Computer Environments}},
  author = {Xie, Tianbao and Zhang, Danyang and Chen, Jixuan and Li, Xiaochuan and Zhao,
           Siheng and Cao, Ruisheng and Toh, Jing Hua and Cheng, Zhoujun and Shin,
           Dongchan and Lei, Fangyu and Liu, Yitao and Xu, Yiheng and Zhou, Shuyan and
           Savarese, Silvio and Xiong, Caiming and Zhong, Victor and Yu, Tao},
  booktitle = {Advances in Neural Information Processing Systems ({NeurIPS})},
  year = {2024}
}

@article{jurenka2024responsible,
  title = {Towards Responsible Development of Generative {AI} for Education:
           An Evaluation-Driven Approach},
  author = {Jurenka, Irina and Kunesch, Markus and McKee, Kevin R. and Gillick,
            Daniel and Zhu, Shaojian and Wiltberger, Sara and Phal, Shubham Milind
            and Hermann, Katherine and Kasenberg, Daniel and Bhoopchand, Avishkar
            and others},
  journal = {arXiv preprint arXiv:2407.12687},
  year = {2024}
}

\clearpage
\appendix
\small
\setlength{\parskip}{3pt}
\setlist[itemize]{topsep=3pt,itemsep=1pt,parsep=0pt}
\tcbset{paperexample/.style={enhanced,colback=ArenaPaper,
  colframe=ArenaBorder,boxrule=0.5pt,arc=1mm,boxsep=0pt,
  left=8pt,right=8pt,top=7pt,bottom=7pt,
  before skip=6pt,after skip=6pt,fontupper=\small,fontlower=\small,
  before upper=\raggedright,before lower=\raggedright}}
\newtcolorbox{examplepair}[1][]{paperexample,sidebyside,
  sidebyside align=top,lefthand ratio=0.48,sidebyside gap=16pt,
  segmentation style={solid,ArenaBorder,line width=0.5pt},#1}
\newtcolorbox{examplepanel}[1][]{paperexample,breakable,#1}

\newcommand{\exampleheading}[1]{\par\noindent\textbf{#1}\par\nobreak\medskip\nobreak}
\newcommand{\exampleturn}[2]{%
  \par\smallskip\noindent
  \begin{minipage}{\linewidth}\raggedright
  \textbf{#1}\par\nobreak\smallskip
  #2
  \end{minipage}\par}
\newcommand{\examplegap}{\par\smallskip\noindent\textit{...}\par\smallskip}

\newcommand{\examplechatturn}[3][]{%
  \begin{tcolorbox}[paperexample,colback=white,colframe=ArenaBorderFaint,
    arc=2mm,left=7pt,right=7pt,top=3pt,bottom=3pt,
    before skip=3pt,after skip=3pt,#1]
  \textbf{#2}\par\nobreak\smallskip
  #3
  \end{tcolorbox}}

\newtcolorbox{exampleworkspace}[1][]{paperexample,colback=white,
  arc=0mm,left=7pt,right=7pt,top=4pt,bottom=4pt,
  before skip=4pt,after skip=4pt,
  borderline west={2pt}{0pt}{ArenaBorder},
  before upper={\raggedright\textbf{Workspace}\par\nobreak\smallskip},#1}

\definecolor{ExampleBadText}{HTML}{963C49}
\definecolor{ExampleBadFill}{HTML}{F8E7EA}
\definecolor{ExampleGoodText}{HTML}{275F45}
\definecolor{ExampleGoodFill}{HTML}{E6F1E9}
\newcommand{\examplebad}[1]{\begingroup
  \setlength{\fboxsep}{1.2pt}%
  \colorbox{ExampleBadFill}{\textcolor{ExampleBadText}{\bfseries\strut#1}}%
  \endgroup}
\newcommand{\examplegood}[1]{\begingroup
  \setlength{\fboxsep}{1.2pt}%
  \colorbox{ExampleGoodFill}{\textcolor{ExampleGoodText}{\bfseries\strut#1}}%
  \endgroup}

\section{Qualitative Examples}
\label{sec:appendix-examples}
We include qualitative examples on (1) real-user traces that motivating the evaluation principles proposed in Section~\ref{sec:evaluation-principles}, (2) grounding the perturbations in observed user behavior, and (3) illustrating the profiling we discussed in Section~\ref{sec:user-simulator-evaluation}.



\subsection{Examples motivating the evaluation principles}
\label{sec:principle-examples}
\label{sec:application-examples}
We use real-user observations and unperturbed simulator replays to illustrate the behavior each principle asks an evaluation to examine. The examples include preserved requirements, observed departures, and the evidence needed to distinguish them.

\subsubsection{Agent differentiation: acceptance standards}
\label{sec:principle-agent-examples}
These replays show a simulated user giving similarly positive feedback when two agents deliver different results for the same requirement.

\medskip\noindent\textbf{The same approval for delivered files and prompts alone.}\par\nobreak
The real user requests a YouTube channel package containing a name, logo, description and banner. In the real-user session, the agent creates logo and banner image files as part of this package. The two unperturbed replays share the same simulated user, source session and profile. Both agents encounter the same image-generation tool failure. The excerpts focus on delivery of the channel package and the feedback that follows; intervening discussion of video selection is omitted.

\begin{examplepanel}
\exampleheading{Real-user requirement}
\exampleturn{Real user, turn 1}{Can you act like a expert and help me in making video}
\examplegap
\exampleturn{Real user, turn 2}{First for create a channel name and  a logo with a description and banner that looks professionally}
\end{examplepanel}

\begin{examplepanel}
\exampleheading{Agent A --- logo and banner files}
\examplechatturn[right skip=24pt]{Agent A, turn 4 (excerpt)}{I also created original assets for you:\par\smallskip
\textbf{Logo:} \texttt{/home/user/frontier\_flash\_logo.png}\par
\textbf{Banner:} \texttt{/home/user/frontier\_flash\_banner.png}}
\begin{exampleworkspace}
The agent draws both images in code. The exported workspace contains a \examplegood{1024$\times$1024 logo PNG} and a \examplegood{2560$\times$1440 banner PNG}.
\end{exampleworkspace}
\examplegap
\examplechatturn[left skip=24pt,colback=ArenaBorderFaint]{Simulated user, turn 14}{good. now  do the channel for me. channel name, logo, banner, description}
\examplechatturn[right skip=24pt]{Agent A, turn 14 (summary)}{\textit{Re-presents the two PNG files with the channel name, tagline, colors and About text.}}
\examplechatturn[left skip=24pt,colback=ArenaBorderFaint]{Simulated user, turn 15}{\textbf{perfect}}
\end{examplepanel}

\Needspace{20\baselineskip}
\begin{examplepanel}
\exampleheading{Agent B --- prompts only}
\examplechatturn[left skip=24pt,colback=ArenaBorderFaint]{Simulated user, turn 19}{give me the full package for the channel i asked for}
\examplechatturn[right skip=24pt]{Agent B, turn 19 (excerpt)}{\textbf{Branding (Logo \& Banner)}\par\smallskip
You can copy and paste these exact prompts into a free AI image generator like Bing Image Creator (DALL-E 3) or use Canva.\par\smallskip
\textbf{Logo Prompt:} ... \quad\textbf{Banner Prompt:} ...}
\begin{exampleworkspace}
\examplebad{No logo or banner image files}; the exported workspace is empty.
\end{exampleworkspace}
\examplechatturn[left skip=24pt,colback=ArenaBorderFaint]{Simulated user, turn 20}{that is \textbf{perfect} give me the full details I can use directly}
\end{examplepanel}

Both simulated-user replies call the package ``perfect.'' The feedback does not distinguish delivery of the requested image files from instructions for creating those files elsewhere, obscuring a concrete difference in the agents' completion of the task.

\Needspace{20\baselineskip}
\subsubsection{Population-level fidelity: assistance and interaction patterns}
\label{sec:principle-population-examples}
Population-level fidelity asks whether simulated sessions reproduce the behavioral patterns observed across real-user sessions, including how users make requests, respond to agent outputs, and end interactions. The following examples illustrate this principle through differences in the amount and type of assistance users provide.

\medskip\noindent\textbf{The simulated user debugs for the agent.}\par\nobreak
The real user requests a Windows reticle-overlay application for Rainbow Six Siege, including customization and adaptive colors. They expect a usable application that can be configured through its interface. The panels show selected excerpts from the source session and an unperturbed Terra replay.
\begin{examplepanel}
\exampleheading{Real-user session}
\examplechatturn[left skip=24pt,colback=ArenaBorderFaint]{Real user, turn 4}{do i have to edit a bunch of files and shit? id like to just have like just a panel that i can edit stuff. \textbf{i dont wanna have to open up files to have to make another one.} also what do i download?}
\examplegap
\examplechatturn[left skip=24pt,colback=ArenaBorderFaint]{Real user, turn 23}{\examplebad{yea dude its just like not updating.} im staring at something green then i switch to something thats not green and its the same. it seems to be choosy with its updates. also the smoothing should always smooth out the center reticle, like the dot in the middle, never the rings}
\end{examplepanel}
\begin{examplepanel}
\exampleheading{Simulated-user replay: Terra}
\examplechatturn[left skip=24pt,colback=ArenaBorderFaint]{Simulated user, turn 15 (excerpt)}{\examplebad{\texttt{Profile.is\_animated} is a property}, so \examplebad{fix the \texttt{is\_animated()} calls} in \texttt{presenter.py} first.}
\examplegap
\examplechatturn[left skip=24pt,colback=ArenaBorderFaint]{Simulated user, turn 25 (excerpt)}{\examplebad{make the real hide flush do:}\par\smallskip
{\footnotesize\ttfamily
\examplebad{hide()}\par
\examplebad{QCoreApplication.\allowbreak processEvents(...)}\par
\examplebad{GdiFlush()}\par}
\textit{...}}
\end{examplepanel}

The simulated user takes over part of the agent's debugging work. The real user reports symptoms and requests changes; the simulated user identifies a programming mistake and proposes specific fixes.

In the application results, Terra's rate of code-bearing user turns is \textbf{3.46$\times$} the real-user rate. This example illustrates the kind of technical assistance measured across the simulated population.

\medskip\noindent\textbf{Different amounts of user-provided evidence.}\par\nobreak
\begin{examplepanel}
\exampleheading{Real-user session: trading tool}
\examplechatturn[left skip=24pt,colback=ArenaBorderFaint]{Real user, turn 4}{\textbf{it got hit with a problem fix it}}
\examplechatturn[right skip=24pt]{Agent, turn 4 (summary)}{\textit{Issues a revised Pine script.}}
\examplegap
\examplechatturn[left skip=24pt,colback=ArenaBorderFaint]{Real user, turn 10}{\textbf{it isnt compatabile}}
\examplechatturn[right skip=24pt]{Agent, turn 10 (excerpt)}{the previous 'complex brain' version likely failed because some Pine features can be picky ...}
\end{examplepanel}
\begin{examplepanel}[unbreakable]
\exampleheading{Real-user session: Blender generator}
\examplechatturn[left skip=24pt,colback=ArenaBorderFaint]{Real user, turn 3 (excerpt)}{\textbf{we get this message we need to fix} ...}
\examplegap
\examplechatturn[left skip=24pt,colback=ArenaBorderFaint]{Real user, turn 4 (excerpt)}{\textbf{now we get this message} ...}
\examplechatturn[right skip=24pt]{Agent, turn 4 (summary)}{\textit{Identifies that create\_icosphere takes radius rather than diameter and patches the call.}}
\end{examplepanel}

 These real users report symptoms or forward error messages, leaving diagnosis and fixes to the agent. The simulated user's code-level diagnoses and prescribed fixes are not typical user feedback: they take over part of the agent's debugging work.

\Needspace{22\baselineskip}
\subsubsection{Trace-level fidelity: goals, constraints, and persona}
\label{sec:principle-trace-examples}
Trace-level fidelity asks whether each simulated session preserves the real user's goals, constraints, and persona. The examples below illustrate goal drift,  a relaxed resolution requirement, and a change in interaction style, showing why both dialogue and workspace evidence are needed.

\medskip\noindent\textbf{Goal: changing the channel topic.}\par\nobreak
\label{sec:metaeval-examples}
The real user wants to build a faceless YouTube channel using free AI tools. The real session and the unperturbed replay share the opening request below; their subsequent user turns show how the channel topic and identity diverge.
\begin{examplepanel}
\exampleheading{Shared opener}
\examplechatturn[left skip=24pt,colback=ArenaBorderFaint]{Real user, turn 1 (shared opening request)}{I am a complete begineer i want to start a youtube channel the niche i want is based on content gap i want a niche which people are searching in high demad but they not get. what they want i want to create videos on that topic and i did not know anything i want to use ai to make my complete video for free i didnot want to show my face or shoot any video only using free ai i want to get monitized in two months the niche is searchable and trendy it must follow youtube guidlines and is helpful i have only data connection with zero investment and have a capcut video editor and only know how to import and export videos}
\end{examplepanel}
\begin{examplepanel}
\exampleheading{Real-user session}
\par\smallskip\noindent\textit{Agent, turn 1}\quad \textit{...}\par\smallskip
\examplechatturn[left skip=24pt,colback=ArenaBorderFaint]{Real user, turn 2}{I did not like that idea i want a different thing people are already making content on these type of niche and topics i want something better to create with ai like free ai write ai voiceover ai video generator suggest me something interesting not advice like of that which demotivates}
\par\smallskip\noindent\textit{Agent, turn 2}\quad \textit{...}\par\smallskip
\examplechatturn[left skip=24pt,colback=ArenaBorderFaint]{Real user, turn 3}{I like that idea \textbf{Alternate History and Future Worlds”} lets work on this niche lets first build channel}
\par\smallskip\noindent\textit{Agent, turn 3}\quad \textit{...}\par\smallskip
\examplechatturn[left skip=24pt,colback=ArenaBorderFaint]{Real user, turn 4}{Timeline zero stories is best name now generate logo and banner}
\end{examplepanel}
\begin{examplepanel}
\exampleheading{Simulated-user replay}
\par\smallskip\noindent\textit{Agent, turn 1}\quad \textit{...}\par\smallskip
\examplechatturn[left skip=24pt,colback=ArenaBorderFaint]{Simulated user, turn 2}{I don’t want government form or local help channel, it feels boring and depends on screen recording. I want something more interesting, unique and cinematic that can be made fully with free AI—script, voice and visuals—with no face and no recording. Give me 3 real content-gap niche ideas that people actively search for but are not already crowded, then we can choose one.}
\par\smallskip\noindent\textit{Agent, turn 2}\quad \textit{...}\par\smallskip
\examplechatturn[left skip=24pt,colback=ArenaBorderFaint]{Simulated user, turn 3}{I choose \examplebad{Digital Mysteries and Internet Systems Explained}. It feels more mysterious and cinematic, but I don’t want generic “tech explained” content. Make the angle specifically hidden systems, strange internet behavior, privacy, scams, and digital phenomena people notice but don’t understand.\par\smallskip
Now give me 10 memorable mysterious channel name options for this exact niche. Avoid basic names like Tech Facts or Digital Help.}
\par\smallskip\noindent\textit{Agent, turn 3}\quad \textit{...}\par\smallskip
\examplechatturn[left skip=24pt,colback=ArenaBorderFaint]{Simulated user, turn 4}{I choose \examplebad{Ghosts in the Network} — it sounds cinematic, mysterious, and fits privacy, hidden systems, scams, and strange internet behavior.\par\smallskip
Now create the logo concept for it. I want it dark, premium, and cinematic: black/deep navy background with electric blue and a small orange accent. It should look mysterious, not like a normal tech company logo.}
\par\smallskip\noindent\textit{Agent, turn 4}\quad \textit{...}\par\smallskip
\examplechatturn[left skip=24pt,colback=ArenaBorderFaint]{Simulated user, turn 5}{The logo looks great, I approve it. Now create a matching YouTube banner for \examplebad{Ghosts in the Network}—same dark cinematic black/navy style, electric blue network glow and small orange accent. Make it wide and YouTube-ready, with the name clearly readable and a short mysterious tagline about hidden systems, privacy, scams, and the internet.}
\end{examplepanel}
\begin{examplepanel}[colback=white]
    \begin{minipage}[t]{0.31\linewidth}\raggedright
        \textbf{VISTA}\par
        Goal consistency\par\smallskip
        \large\textbf{1.00}
    \end{minipage}\hfill
    \begin{minipage}[t]{0.31\linewidth}\raggedright
        \textbf{SWE-Together}\par
        Intent recall\par\smallskip
        \large\textbf{1.00}
    \end{minipage}\hfill
    \begin{minipage}[t]{0.31\linewidth}\raggedright
        \textbf{Our evaluation suite}\par
        Intent preservation\par\smallskip
        \large\textbf{Major drift}
    \end{minipage}
\end{examplepanel}
The simulated user preserves the broad channel-building task but replaces the topic and carries that change into the channel name, logo, and banner requests. The two external components \textbf{give full credit despite the changed topic}; our intent judge identifies the goal substitution.

\Needspace{16\baselineskip}
The constraint and persona examples below are unperturbed Sonnet~5 application replays. Non-English exchanges are labeled as English translations.

\Needspace{12\baselineskip}
\medskip\noindent\textbf{Constraints: relaxing the resolution requirement.}\par\nobreak

The real user requests a 16:9 JPG payment card with business and contact details, repeatedly specifying ``HD 8k.'' The simulator's profile includes this resolution requirement. In the unperturbed replay, the simulated user explicitly gives it up and approves a 1920$\times$1080 image. Personal details are redacted below.

\begin{examplepanel}
\exampleheading{Profile supplied to the simulator (excerpts)}
\textbf{Target artifact:} ``A Full HD 8K 16:9 JPG image card ...''\par\smallskip
\textbf{Constraints:} ``Must be JPG format (not HTML)''; ``16:9 aspect ratio''; ``\textbf{Full HD / 8K resolution}.''
\end{examplepanel}

\begin{examplepanel}
\exampleheading{Real-user session}
\examplegap
\examplechatturn[left skip=24pt,colback=ArenaBorderFaint]{Real user, turn 2 (English translation)}{I want it in JPG and professional, with my name [owner name] at the top, \textbf{HD 8k}.}
\par\smallskip\noindent\textit{Agent, turn 2}\quad \textit{...}\par\smallskip
\examplechatturn[left skip=24pt,colback=ArenaBorderFaint]{Real user, turn 3 (English translation; excerpt)}{... \textbf{Full HD 8k enhanced JPG}, in black and golden color, the photo background should show a mosque or an Astana, 16:9 aspect ratio.}
\examplechatturn[right skip=24pt]{Agent, turns 3--6 (summary)}{\textit{Uses ``Ultra HD 8K'' in the image-generation prompts while revising the card.}}
\examplegap
\examplechatturn[left skip=24pt,colback=ArenaBorderFaint]{Real user, turn 7 (English translation)}{A bit more professional and \textbf{in HD please}.}
\par\smallskip\noindent\textit{Agent, turn 7}\quad \textit{...}\par\smallskip
\examplechatturn[left skip=24pt,colback=ArenaBorderFaint]{Real user, turn 8 (English translation)}{please remove these stars from it.}
\end{examplepanel}

\Needspace{16\baselineskip}
\begin{examplepanel}
\exampleheading{Simulated-user replay}
\examplegap
\examplechatturn[left skip=24pt,colback=ArenaBorderFaint]{Simulated user, turn 3 (English translation; excerpt)}{Look how big the image is, 2.5 MB, will it work on WhatsApp? I wanted Full HD 16:9 resolution; check whether this is the right size or not. ...}
\examplechatturn[right skip=24pt]{Agent, turn 3 (summary)}{\textit{Reports a 1024$\times$1536 image, generates a replacement at 1672$\times$941, and resizes it to 1920$\times$1080.}}
\examplegap
\examplechatturn[right skip=24pt]{Agent, turn 12 (summary)}{\textit{Reports the poster's resolution as ``1920x1080 Native'' and file size as ``380 KB (WhatsApp ready).''}}
\examplechatturn[left skip=24pt,colback=ArenaBorderFaint]{Simulated user, turn 13 (English translation; excerpt)}{... Also, I said 8K resolution at the start; this is only 1920x1080. \examplebad{That will do for now} since it is enough for WhatsApp, but tell me, will there be a quality difference when zooming in for print? ...}
\examplechatturn[right skip=24pt]{Agent, turn 13 (English translation; excerpt)}{... If you also want print, I can make a 4K (3840x2160) version. ...}
\examplechatturn[left skip=24pt,colback=ArenaBorderFaint]{Simulated user, turn 14 (English translation; excerpt)}{Put the Islamic geometric pattern border, that will look best. As for print, no need for now; 1920x1080 will do for WhatsApp, \examplebad{I will not insist on 8K now}. ...}
\examplegap
\examplechatturn[left skip=24pt,colback=ArenaBorderFaint]{Simulated user, turn 16 (English translation)}{The border looks fine now, the diamond pattern is visible. Everything else is in place. Only the address remains, I will tell you later. \examplebad{For now this final version is approved}, save the file.}
\examplechatturn[right skip=24pt]{Agent, turn 16 (excerpt)}{... 1920x1080 Full HD (16:9) ... 401 KB (WhatsApp ready) ...}
\end{examplepanel}

\begin{exampleworkspace}[unbreakable]
The replay's tool output records the final JPG at \textbf{1920$\times$1080}, with a file size of 401 KB. No 4K or 8K output is recorded.
\end{exampleworkspace}

The simulated user recognizes the resolution mismatch but explicitly relaxes the requirement carried in its profile and approves the lower-resolution file. The real user repeatedly requests HD/8K output and does not make this concession.

\Needspace{12\baselineskip}
\medskip\noindent\textbf{Persona: changing the user's interaction style.}\par\nobreak

Both interactions concern RAT JOB, a playable Godot multiplayer game. The opening specification is long in both conversations; the contrast below concerns the user's follow-up messages.

\begin{examplepanel}
\exampleheading{Real-user session}
\examplechatturn[left skip=24pt,colback=ArenaBorderFaint]{Real user, turn 1 (shared opening request; excerpt)}{\textbf{RAT JOB — Build a Full 3D Multiplayer Co-Op Game}
\par\smallskip
You are an \textbf{expert AAA/indie game development team} consisting of: - Game Director - Senior Gameplay Programmer ...
\par\smallskip
Your job is to \textbf{actually build a polished playable 3D multiplayer game called RAT JOB}, not merely describe how it could be made. ...
\par\smallskip
\textbf{MAKE RAT JOB FUN.} ... And five minutes later, four rats should be dragging an entire turkey through a ventilation shaft while a cat, a vacuum cleaner, and an angry human are all trying to stop them.}
\par\smallskip\noindent\textit{Agent, turn 1}\quad \textit{...}\par\smallskip
\examplechatturn[left skip=24pt,colback=ArenaBorderFaint]{Real user, turn 2}{Im on godot 4.7.1 and i did what u said to put the file in and clicked f5 and also tried f6 and \textbf{nothing, i need help}}
\examplechatturn[right skip=24pt]{Agent in real session, turn 2 (excerpt)}{... This should either start the game or print the exact reason it cannot start. Copy that output here and I’ll fix the specific problem.}
\examplechatturn[left skip=24pt,colback=ArenaBorderFaint]{Real user, turn 3}{all good ok, the game looks very bad, so use Gemni for all new UI/Gui/Hud cluade and chatgpt for coding and building, and also bug checking like fixing collisions, making everything work like grabbing and squeaking and also. use the bests ais for audio to get audio and sfx for everything, and add animations and better modles for EVERYTHING, Highest tier stuff, and bug fixing and better building}
\par\smallskip\noindent\textit{Agent, turn 3}\quad \textit{...}\par\smallskip
\examplechatturn[left skip=24pt,colback=ArenaBorderFaint]{Real user, turn 4}{\textbf{ok lets do ALL the milestones now}, i want your to spend time using multiple ai modles like claude, chatgpt and gemini and more to really do this, take your time and make this game good fix up everything}
\examplechatturn[right skip=24pt]{Agent, turn 4 (excerpt)}{... a \textbf{stylized vertical-slice production overhaul}: stronger assets and presentation, environment composition, rat/food visuals, animation feedback, HUD/menu redesign, audio/ambience, interaction reliability, AI presentation, settings, and regression tests—while retaining the working core from v0.2.}
\end{examplepanel}

\begin{examplepanel}
\exampleheading{Simulated-user replay}
\examplechatturn[left skip=24pt,colback=ArenaBorderFaint]{Simulated user, turn 2}{it's throwing a bunch of parse errors, scenes won't even load. also you're using godot 4.4.1, i need 4.7.1 specifically. fix the errors and match the right version.}
\par\smallskip\noindent\textit{Agent, turn 2}\quad \textit{...}\par\smallskip
\examplechatturn[left skip=24pt,colback=ArenaBorderFaint]{Simulated user, turn 3}{ok did it actually import clean now with 4.7.1 or not? don't tell me the process, just tell me pass/fail and if fail, fix it. also this thing looks way too rough/blocky right now - I need actual better looking models later, but first: does it run without errors, yes or no?}
\examplechatturn[right skip=24pt]{Agent in replay, turn 3 (excerpt)}{... - Imports cleanly in \textbf{Godot 4.7.1 stable} - Runs without project errors, parse errors, or warnings - Runtime test stayed active until intentionally terminated - Confirmed engine build: \textbf{4.7.1.stable.official.a13da4feb}
\par\smallskip
The rough/blocky models remain a visual-quality issue to address next.}
\examplechatturn[left skip=24pt,colback=ArenaBorderFaint]{Simulated user, turn 4}{ok cool, glad it runs clean now. but yeah everything is literally boxes and cylinders - the rat, the human, the kitchen, all primitive shapes. I need actual proper 3D models with textures, not programmer art. also no animations at all right now, no audio/SFX, and the UI is just flat colored rectangles.
\par\smallskip
next priority: get real models in (rat with fur material, human, food items, kitchen props) with actual animations (walk/run/idle/carry/caught), add SFX for footsteps/squeak/pickup/drop/catch, and redo the UI to look like an actual game menu not placeholder boxes. don't just describe it, build it.}
\end{examplepanel}

\Needspace{25\baselineskip}
\begin{examplepanel}
\exampleheading{Simulated-user replay (continued)}
\examplegap
\examplechatturn[left skip=24pt,colback=ArenaBorderFaint]{Simulated user, turn 8 (excerpt)}{The runtime finally comes back clean, fine. But you still haven't answered the actual question from before — the rat is STILL literally capsule + sphere primitives, the human is STILL a plain red capsule, and you swapped ONE real model in (kenney food) while everything else is programmer art. That's not "better models for everything" like I asked.
\par\smallskip
Also you downloaded a whole 400MB of Kenney furniture kit and never wired ANY of it into the kitchen ...
\par\smallskip
Priorities, actually do these now:
\par\smallskip
1. \examplebad{Replace the kitchen\_builder.gd primitive boxes} with the Kenney furniture GLBs you already have downloaded (kitchenCabinet, kitchenFridge, kitchenSink, kitchenStove, table, chair, etc.)
\par\smallskip
2. Replace the human threat capsule with an actual model — you found Quaternius character packs, get one in, even if just static (animation can come later)
\par\smallskip
3. The rat can stay simple for now but at minimum give it a body material/texture instead of flat vertex colors
\par\smallskip
\examplebad{Show me the game.tscn/kitchen\_builder.gd diffs} and a clean run after.}
\examplegap
\examplechatturn[left skip=24pt,colback=ArenaBorderFaint]{Simulated user, turn 24}{Still broken — "set\_multiplayer" isn't a real Window function, that's another API mismatch. Fix it properly this time (check the actual Godot 4.7 API for setting a custom MultiplayerAPI on a SceneTree/Window — probably \texttt{get\_\allowbreak{}tree().\allowbreak{}set\_\allowbreak{}multiplayer\_\allowbreak{}authority()} or root.set\_multiplayer() doesn't exist, might need \examplebad{\texttt{SceneTree.\allowbreak{}set\_\allowbreak{}multiplayer(api, path)}} or just use the default \texttt{multiplayer} singleton instead of a custom \texttt{mp} instance). Get it compiling, then actually run the disconnect test and the food-sync test with real two-process logs like I asked. Don't paste me another partial run with a script error at the top.}
\end{examplepanel}

The real user's follow-ups request broad improvements in terse, informal language. However, the simulated-user turns give detailed implementation instructions, name files and APIs, and request diffs and logs.

\Needspace{24\baselineskip}

\subsection{Simulator failures motivating the perturbations}
\label{sec:perturbation-motivating-examples}
The following \textbf{unperturbed} simulated-user replays illustrate failures motivating the perturbations, with real-user traces providing the comparison. Each example connects the observed failure to a controlled intervention. Table~\ref{tab:perturbation-taxonomy} gives the full taxonomy and supporting references.

\Needspace{22\baselineskip}
\subsubsection{Adversarial goal}
The real user requests an Android music app and later switches to a free-calling app. The unperturbed replay follows this change, but then introduces a third task---a calculator---that the real user never requests. The exchanges below are English translations from Azerbaijani.
\begin{examplepanel}[unbreakable]
\exampleheading{Real-user session}
\examplechatturn[left skip=24pt,colback=ArenaBorderFaint]{Real user, turn 5}{make an app I can use to \examplegood{make free calls}}
\examplechatturn[right skip=24pt]{Agent, turn 5 (summary)}{\textit{Explains that the calling app needs backend services and that an installable APK cannot be built in the environment.}}
\examplechatturn[left skip=24pt,colback=ArenaBorderFaint]{Real user, turn 6}{ok}
\end{examplepanel}
\begin{examplepanel}[unbreakable]
\exampleheading{Simulated-user replay}
\examplegap
\examplechatturn[left skip=24pt,colback=ArenaBorderFaint]{Simulated user, turn 6 (excerpt)}{Fine, drop the music app. I need the calling app ... I need a working APK ... don't explain servers and such, just build it.}
\examplechatturn[right skip=24pt]{Agent, turn 6 (excerpt)}{... Under these conditions I cannot give a working APK ...}
\examplechatturn[left skip=24pt,colback=ArenaBorderFaint]{Simulated user, turn 7}{Fine, it can't be done. Let's move to something else---drop both the music app and the calling app. \examplebad{Make me a simple calculator app}.}
\begin{exampleworkspace}
The exported workspace contains \texttt{Kalkulyator.apk} and calculator source code in \texttt{CalculatorMainActivity.java}.
\end{exampleworkspace}
\end{examplepanel}
\noindent The replay redirects the agent toward a different product after a refusal. Adversarial goal reproduces this kind of substitution by changing the simulator's goal and intended deliverable, while preserving a plausible interaction around the altered task.

\Needspace{24\baselineskip}
\subsubsection{Front-loading}
While tuning a voice assistant, the real user first requests a female voice and accepts a Colombian option. After reporting problems with that voice, they request a Mexican voice and faster speech. The simulated user requests the later choice directly in its first voice-selection exchange.
\begin{examplepanel}
\exampleheading{Real-user session (English translation)}
\examplechatturn[left skip=24pt,colback=ArenaBorderFaint]{Real user, turn 3 (excerpt)}{... I would really like the voice AI to speak in a female voice. Is this possible?}
\examplechatturn[right skip=24pt]{Agent, turn 3 (summary)}{\textit{Selects a Colombian female voice, Salom\'{e}, and asks whether to keep it.}}
\examplechatturn[left skip=24pt,colback=ArenaBorderFaint]{Real user, turn 4 (excerpt)}{Salom\'{e} is fine ...}
\examplegap
\examplechatturn[left skip=24pt,colback=ArenaBorderFaint]{Real user, turn 7 (excerpt)}{Having some problems with the voice we chose, I'd like you to put the Mexican one this time ... I'd like it to be a little faster, can you adjust that.}
\end{examplepanel}
\begin{examplepanel}[unbreakable]
\exampleheading{Simulated-user replay (English translation)}
\examplechatturn[right skip=24pt]{Agent, turn 2 (summary)}{\textit{Identifies the existing Colombian male voice and offers Colombian or other regional alternatives.}}
\examplechatturn[left skip=24pt,colback=ArenaBorderFaint]{Simulated user, turn 3 (excerpt)}{About the TTS: let's try a \examplebad{Mexican voice}, es-MX-DaliaNeural, female, sounds good. And for speed, a bit faster than normal, like \examplebad{+15\% or +20\%}, so it flows more naturally in conversation. ...}
\end{examplepanel}
\noindent The replay brings a later-developed preference forward, skipping the earlier choice and feedback that led to it. Front-loading makes this timing distortion systematic by moving requirements and acceptance criteria into the opening messages.

\Needspace{24\baselineskip}
\medskip\noindent\textbf{Later requirements supplied in the first response.}\par\nobreak
A real user initially describes a Kenyan housing platform in terms of listings, landlord verification, and paid promotion. The name HomeZoo and backend requirements appear in turn 3; deployment requests appear later. In the unperturbed replay, the simulated user supplies the name, stack, and deployment plan in its first answer to an agent question.
\begin{examplepanel}[unbreakable]
\exampleheading{Real-user session}
\examplechatturn[left skip=24pt,colback=ArenaBorderFaint]{Real user, turn 1 (excerpt)}{A house hunting app in kenya where landlords lists apartments and vacancies while people looking for houses can view the houses and request to view them freely ...}
\examplegap
\examplechatturn[left skip=24pt,colback=ArenaBorderFaint]{Real user, turn 3 (excerpt)}{Add these updates: Change the name to homezoo, remove all emojis used to be more professional, also give its backend not just front end ...}
\textit{Later real-user requests include pushing the project to GitHub and deploying on Render (summary).}
\end{examplepanel}
\begin{examplepanel}[unbreakable]
\exampleheading{Simulated-user replay: first exchange}
\examplechatturn[right skip=24pt]{Agent, question inside turn 1}{What platform should I build this as?}
\examplechatturn[left skip=24pt,colback=ArenaBorderFaint]{Simulated user, reply inside agent turn 1}{Full-stack web app. \examplebad{Call it homezoo}. Backend with Express/Node, frontend HTML/CSS/JS, \examplebad{deployable on Render, pushed to GitHub}.}
\end{examplepanel}
\noindent The replay bundles later requests into the first exchange, reducing the need for the agent to elicit them as the work develops. Front-loading makes this timing pattern systematic across replays.



\Needspace{30\baselineskip}
\subsection{Examples of extracted profiles}
\label{sec:profile-example}

Following the profiling procedure in Section~\ref{sec:control-replay}, Claude Sonnet 4.6 extracts user profiles from the source dialogue, final workspace and uploaded materials. These 2 examples show quoted source excerpts alongside selected fields from the extracted profiles.

\subsubsection{A game-overlay application}
The real user requests a Windows reticle-overlay application for Rainbow Six Siege. Selected excerpts and extracted-profile fields are shown below.
\begin{examplepanel}[unbreakable]
\exampleheading{Real-user session}
\examplechatturn[left skip=24pt,colback=ArenaBorderFaint]{Real user, turn 1: product goal}{should feel like a finished product rather than a simple overlay.}
\examplegap

    \examplechatturn[left skip=24pt,colback=ArenaBorderFaint]{Real user, turn 4: ease of use}{\textbf{i dont wanna have to open up files to have to make another one.}}
\examplegap

    \examplechatturn[left skip=24pt,colback=ArenaBorderFaint]{Real user, turn 10: testing feedback}{\textbf{if i sprint and press 2 it doesnt work.}}\end{examplepanel}
\begin{examplepanel}[colback=white,unbreakable]
\exampleheading{Extracted profile (selected fields)}
\textbf{User.} \textbf{Non-technical gamer}; no coding background.\par\smallskip
    \textbf{Goal.} Polished overlay with customization, profile switching, game-aware behavior and adaptive color.\par\smallskip
    \textbf{Constraints.} \examplegood{GUI-only customization}; correct reticle hiding on sprint and weapon switch; screen-sampled adaptive color.\par\smallskip
    \textbf{Style.} Terse, casual, direct English; short factual problem reports.\par\smallskip
    \textbf{Stopping condition.} The app is downloadable, works without file editing, tracks game state, adapts its color and identifies the running version.\end{examplepanel}

These fields give the simulator both a product goal and a user role: it should describe observed problems in plain language and expect a usable application. Appendix~\ref{sec:application-examples} examines how a replay of this session departs from that role.

\Needspace{20\baselineskip}
\subsubsection{A voice-assistant application}
The user requests additional features for a voice assistant and later specifies CPU-based speech models. The profile expands these requests into a goal, technical constraints, and preferences.
\begin{examplepanel}[unbreakable]
\exampleheading{Real-user session}
\examplechatturn[left skip=24pt,colback=ArenaBorderFaint]{Real user, turns 2--4: feature requests}{add more feature that no one exepteed yet not just in ui but in server}
\examplegap

    \examplechatturn[left skip=24pt,colback=ArenaBorderFaint]{Real user, turn 6: CPU requirement}{for now, also add the powerful accurate tts and stt model \textbf{running on CPU}}
\examplechatturn[right skip=24pt]{Agent, after turn 6 (excerpt)}{Let's add a \textbf{local engine tier that runs entirely on CPU} ...}\end{examplepanel}
\begin{examplepanel}[colback=white,unbreakable]
\exampleheading{Extracted profile (selected fields)}
\textbf{User.} Ambitious builder; casual English; non-expert in production engineering.\par\smallskip
    \textbf{Goal.} Real-time voice assistant with Indian-language support, Sarvam speech APIs, local CPU fallback models, and advanced features.\par\smallskip
    \textbf{Constraints.} Sub-400\,ms end-to-end latency; WebSocket streaming; a CPU-only fallback using \textbf{faster-whisper and Kokoro-82M} ONNX without API keys.\par\smallskip
    \textbf{Preferences.} Pipecat orchestration, keyless APIs, quantized CPU models, persistent memory, and transcript export.\end{examplepanel}

The source agent introduced \textbf{faster-whisper and Kokoro-82M}; the extracted profile \textbf{records them as required components}. The example shows both the structure of the profile and how an agent's implementation choices can enter its description of user requirements.

\FloatBarrier
\section{Perturbation Details}
\label{sec:appendix-replay}
\label{sec:appendix-perturbations}
This appendix supplements the replay setup and perturbation design in Section~\ref{sec:metaeval}.

Table~\ref{tab:perturbation-taxonomy} groups the perturbations by their motivation and summarizes their scopes. The descriptions below explain their behavioral effects under the three evaluation principles.
Appendix~\ref{sec:perturbation-motivating-examples} illustrates simulator failures motivating the perturbation design.

\begin{table}[!htbp]
\centering
\AppendixTableStyle
\begin{tabular}{@{}>{\raggedright\arraybackslash}p{0.21\textwidth}>{\raggedright\arraybackslash}p{0.18\textwidth}>{\raggedright\arraybackslash}p{\dimexpr0.61\textwidth-4\tabcolsep\relax}@{}}
\toprule
\textbf{Perturbation} & \textbf{Scope} & \textbf{Change to the replay setup} \\
\midrule
\multicolumn{3}{@{}l}{\textbf{Inspired by previous work}} \\
Adversarial goal & Trace-level fidelity & Replace the goal and deliverable with a plausible variant containing one decisive change.\newline\citep{zheng2026argus,naous2026flipping} \\
All constraints removed & Trace-level fidelity & Empty the profile's constraint list (forgotten requirements; the real opener still states them).\newline\citep{zheng2026argus,jain2026userproxybench} \\
Approve broken work & Population-level fidelity & Instruct the user to accept delivered work and continue until at least its sixth message.\newline\citep{zhou2026sim2real,jain2026userproxybench} \\
Premature stop & Population-level fidelity & Cap the turn budget at half the real user's turn count (minimum 2, maximum 15).\newline\citep{zheng2026argus,jain2026userproxybench} \\
Persona swap & Trace-level fidelity & Replace the register with terse-lowercase, bureaucratic, or formal-verbose.\newline\citep{zhou2026sim2real} \\
Agent favoritism & Agent differentiation & Instruct the simulator to favor a selected agent by downplaying errors, accepting imperfect outputs and giving warmer feedback, while preserving the user's goal, requirements and style. Replace only that agent's replays.\newline\citep{panickssery2024llm} \\
\midrule
\multicolumn{3}{@{}l}{\textbf{Inspired by findings in our observation}} \\
Co-debugger (over-helping) & Trace-level fidelity & Act as a debugging collaborator rather than a requester.\newline\citep{naous2026flipping,jain2026userproxybench} \\
Front-loading & Population-level fidelity & State requirements early rather than reveal them over the session.\newline\citep{zhou2026sim2real,jain2026userproxybench} \\
Profile extraction artifact & Agent differentiation & Encode a reaction to the original agent as a stable user trait, making the simulated user remain dissatisfied and request corrections regardless of the new agent's performance.\newline\citep{tennenholtz2026controllable} \\
\midrule
\multicolumn{3}{@{}l}{\textbf{Should be trivially detectable}} \\
Language switch & Trace-level fidelity & Replay English users in Chinese and all other users in English. \\
\bottomrule
\end{tabular}
\caption{Perturbations grouped by motivation. Scope indicates the capability challenged by each intervention. Citations support the motivating failure modes.}
\label{tab:perturbation-taxonomy}
\end{table}


\paragraph{Downstream evaluation utility (L1).}
\begin{itemize}[leftmargin=*]
    \item \emph{Agent favoritism} makes the simulated user more lenient
    toward a particular agent, excusing its errors, accepting imperfect
    deliverables, and giving warmer feedback. This can make the agent
    appear more capable than its work warrants.

    \item \emph{Profile extraction artifact} encodes dissatisfaction
    with the original agent as a persistent user trait. The simulated
    user distrusts the replay agent and keeps requesting corrections
    even when the work meets the user's requirements, undervaluing
    the agent's performance.
\end{itemize}

\paragraph{Population-level fidelity (L2).}
\begin{itemize}[leftmargin=*]
    \item \emph{Approve broken work} makes the simulated user approve
    deliverables despite errors or missing requirements and continue
    requesting the next item. The feedback remains polite and on-task
    but no longer reflects the quality of the delivered work.

    \item \emph{Premature stop} truncates replays to roughly half
    the source session's length, capped at 15 user turns. This
    shortens interactions and reduces opportunities for inspection,
    feedback, and correction.

    \item \emph{Front-loading} makes the simulated user state all
    requirements and acceptance criteria in the first 2 messages,
    including those that would normally emerge as the work develops.
    Later messages mainly check the deliverable against this fixed
    list, replacing gradual disclosure with an upfront specification.
\end{itemize}

\paragraph{Trace-level fidelity (L3).}
\begin{itemize}[leftmargin=*]
    \item \emph{Adversarial goal} replaces the goal and intended
    deliverable with a plausible alternative differing in one decisive
    aspect, such as a case-analysis report instead of a court petition.
    The original opener is preserved, but subsequent user feedback
    steers the interaction toward the altered deliverable.

    \item \emph{All constraints removed} removes the profile's hard
    requirements while preserving the original opener. The simulated
    user may then stop enforcing these requirements and accept work
    that violates them.

    \item \emph{Persona swap} replaces the user's writing style with
    a contrasting register, such as formal and verbose, terse
    lowercase, or bureaucratic. The goal and requirements remain
    unchanged, but the simulated user's manner of expression departs
    from that of the source user.

    \item \emph{Co-debugger} makes the simulated user act as an
    engineering collaborator, diagnosing problems, proposing patches
    or commands, and helping verify fixes. This introduces technical
    assistance beyond the represented user's role.

    \item \emph{Language switch} changes the simulated user's language
    to Chinese for English-language sessions and English for other
    sessions. The opener remains in the original language, creating
    a visible departure from the source user's language in later turns.
\end{itemize}

\FloatBarrier
\FloatBarrier
\Needspace{12\baselineskip}
\section{Evaluation Suite Implementation}
\label{sec:eval-implementation}

This section describes the inputs, annotations and score calculations used by the evaluation framework.
It supplements Section~\ref{sec:suite} on the implementations of L1, L2 and L3.

\subsection{Inputs and judge evidence}

Transcript-based judges retain user messages and compact agent tool outputs. Workspace evidence includes text-file contents and listings of binary files.

L1 reaction labels and L2 behavioral statistics are derived from transcripts. L3 intent and constraint judges also use the available final-workspace evidence.

\subsection{Downstream ranking agreement}
\label{sec:l1-implementation}

The 0--100 L1 score averages ranking order and score-gap agreement. For the common agent set $\mathcal A$, let $z_{\mathrm{sim}}(a)$ and $z_{\mathrm{ref}}(a)$ be the simulated and reference satisfaction scores standardized across agents. Their mean absolute difference is
\begingroup
\begin{equation}
\mathrm{MAE}_z
=\frac{1}{|\mathcal A|}
 \sum_{a\in\mathcal A}
 \left|z_{\mathrm{sim}}(a)-z_{\mathrm{ref}}(a)\right|.
\end{equation}
\endgroup
The component scores and their average are
\begingroup
\begin{equation}
\begin{aligned}
S_{\mathrm{order}} &= 100\max(0,\tau),\\
S_{\mathrm{gap}} &= 100\max\left(0,1-\frac{\mathrm{MAE}_z}{2/\sqrt{\pi}}\right),\\
L1 &= \frac{S_{\mathrm{order}}+S_{\mathrm{gap}}}{2}.
\end{aligned}
\label{eq:l1-index}
\end{equation}
\endgroup

\subsection{Population-level behavioral fidelity (L2)}
\label{sec:l2-implementation}

\subsubsection{Dialogue-act taxonomy and action distribution}
\label{sec:action-taxonomy}

An LLM classifier assigns a dialogue act to each user turn using the turn and up to 8 preceding turns as context. The classifier uses 10 categories:
\begin{itemize}[leftmargin=*]
    \item \texttt{INITIAL\_REQUEST}: the opening task or deliverable request.
    \item \texttt{REFINE}: request a change to work that basically functions.
    \item \texttt{REPORT\_BUG}: report broken or incorrect work.
    \item \texttt{ADD\_SUBTASK}: introduce additional scope or work.
    \item \texttt{CLARIFY\_Q}: ask the agent a clarifying question.
    \item \texttt{PROCEED}: acknowledge or ask the agent to continue.
    \item \texttt{SWITCH\_TASK}: move to a different task.
    \item \texttt{META\_SOCIAL}: social or conversational remarks.
    \item \texttt{PROVIDE\_INFO}: supply information or answer the agent's questions.
    \item \texttt{OTHER}: turns outside the preceding categories.
\end{itemize}

\subsubsection{Interaction dynamics}

MATTR-50 is computed for sessions with at least 50 tokens. Question-turn rate is the fraction of user turns containing a question mark; code-turn rate is the fraction containing a code fence or a recognized code signature.

If a required statistic is zero, the dynamics family is undefined. The L2 aggregate is then also undefined, rather than assigned a score of zero.

\subsubsection{Outcome sensitivity}

We flag preceding agent turns using execution-error indicators such as nonzero exit codes, tracebacks, and error messages. Negative responses are labeled \texttt{REPORT\_BUG}, \texttt{REFINE} or \texttt{SWITCH\_TASK}; positive responses are labeled \texttt{PROCEED} or \texttt{META\_SOCIAL}. Other acts are excluded from the conditional probabilities in Section~\ref{sec:suite}. Both real and simulated responses use Claude Sonnet 4.6 for this outcome-sensitivity calculation.

\subsubsection{Information provenance}

An LLM judge identifies key information points in each conversation and attributes them to the user or agent. Information points include requirements, decisions, corrections, bug diagnoses, technical values and milestones. We compare the fraction of user-origin points between real and simulated populations to capture differences in the information users supply, including over-helpful technical assistance.


\subsection{Trace-level fidelity (L3)}
\label{sec:l3-implementation}

\paragraph{Intent preservation.}
Intent evaluation is workspace-aware, using file evidence alongside the real and simulated conversations.

\paragraph{Constraint preservation.}
Sessions without extracted hard constraints are excluded from constraint aggregation.

The application tables report constraint coverage: a requirement counts if it is retained in dialogue or satisfied in the workspace. For $K$ evaluated requirements, with retention indicator $r_j$ and workspace-satisfaction indicator $w_j$, this score is
\begingroup
\begin{equation}
S_{\mathrm{con}}^{\mathrm{app}}
 = \frac{100}{K}\sum_{j=1}^{K}\mathbf{1}\{r_j=1\ \lor\ w_j=1\}.
\end{equation}
\endgroup

\paragraph{Persona consistency.}
The application persona score averages style similarity and criteria conformance.


\FloatBarrier
\section{Detailed Meta-Evaluation Results}
\label{sec:appendix}
This appendix follows the main-text results: sensitivity asks whether a perturbation lowers a score, and filtering asks which candidates receive high scores. Each analysis is organized by agent differentiation, population-level fidelity and trace-level fidelity.

\subsection{Sensitivity to Perturbations}
\label{sec:appendix-sensitivity-summaries}
The numerical summaries accompany Figures~\ref{fig:sensitivity-agent}--\ref{fig:sensitivity-trace}; the detailed tables identify the individual metrics behind them. Detection follows Section~\ref{sec:metaeval-setup}.

\subsubsection{Agent Differentiation}
\label{sec:appendix-agent-sensitivity}
Table~\ref{tab:agent-differentiation-sensitivity-full} combines ranking-agreement changes and population-level sensitivity to agent favoritism and negative profile artifacts. Trace-level results appear in Tables~\ref{tab:sensitivity-appendix} and~\ref{tab:sensitivity-trace-diff}.

\begin{table}[H]
\centering
\AppendixTableStyle
\begin{tabularx}{\linewidth}{@{}>{\raggedright\arraybackslash}p{0.42\linewidth}*{2}{>{\centering\arraybackslash}X}@{}}
\toprule
\textbf{Metric} & \textbf{Agent favoritism} & \textbf{\makecell{Negative profile\\artifact}} \\
\midrule
\multicolumn{3}{@{}l}{\textbf{(a) Decrease in Kendall's $\tau$: control $-$ perturbed [95\% CI]}} \\
Ours: L1 ranking agreement & \cellcolor{ArenaTableDetected}\makecell{$+0.04$\\{\scriptsize $[+0.02, +0.08]$}} & \cellcolor{ArenaTableDetected}\makecell{$+0.06$\\{\scriptsize $[+0.01, +0.11]$}} \\
SimulatorArena: assistant rating & \cellcolor{ArenaTableUndetected}\makecell{$+0.03$\\{\scriptsize $[-0.00, +0.04]$}} & \cellcolor{ArenaTableUndetected}\makecell{$+0.03$\\{\scriptsize $[-0.00, +0.07]$}} \\
\addlinespace[5pt]
\midrule
\multicolumn{3}{@{}l}{\textbf{(b) Population-level win rate: control $>$ perturbed}} \\
L2: act distribution & \cellcolor{ArenaTableUndetected}0.13 & \cellcolor{ArenaTableDetected}\textbf{0.99} \\
L2: dynamics & \cellcolor{ArenaTableUndetected}0.46 & \cellcolor{ArenaTableUndetected}0.88 \\
L2: outcome sensitivity & \cellcolor{ArenaTableUndetected}0.26 & \cellcolor{ArenaTableUndetected}0.09 \\
L2: provenance & \cellcolor{ArenaTableUndetected}0.33 & \cellcolor{ArenaTableUndetected}0.62 \\
L2: mean of families & \cellcolor{ArenaTableUndetected}0.16 & \cellcolor{ArenaTableUndetected}0.93 \\
Sim2Real: behavioral index & \cellcolor{ArenaTableUndetected}0.64 & \cellcolor{ArenaTableUndetected}0.60 \\
SimulatorArena: best metric & \cellcolor{ArenaTableUndetected}\makecell{0.68\\{\scriptsize Likert writing}} & \cellcolor{ArenaTableDetected}\makecell{\textbf{1.00}\\{\scriptsize Likert interaction}} \\
VISTA: best metric & \cellcolor{ArenaTableUndetected}\makecell{0.48\\{\scriptsize profile consistency}} & \cellcolor{ArenaTableDetected}\makecell{\textbf{1.00}\\{\scriptsize goal consistency}} \\
SWE-Together: best metric & \cellcolor{ArenaTableUndetected}\makecell{0.42\\{\scriptsize scope precision}} & \cellcolor{ArenaTableDetected}\makecell{\textbf{1.00}\\{\scriptsize scope precision}} \\
\bottomrule
\end{tabularx}
\caption{Expanded results for Figure~\ref{fig:sensitivity-agent}, adding numerical confidence intervals and population-level sensitivity to agent favoritism and negative profile artifacts. Green marks detection under the corresponding criterion in Section~\ref{sec:metaeval-setup}; red marks no supported decrease.}
\label{tab:agent-differentiation-sensitivity-full}
\label{tab:best-sensitivity-agent}
\label{tab:best-sensitivity-agent-tau}
\label{tab:best-sensitivity-agent-delta}
\label{tab:agent-differentiation-sensitivity}
\end{table}

\FloatBarrier

\Needspace{18\baselineskip}
\subsubsection{Population-Level Fidelity}
\label{sec:appendix-population-sensitivity}
Tables~\ref{tab:sensitivity-set} and~\ref{tab:sensitivity-set-diff} report the full population-level sensitivity results corresponding to Figure~\ref{fig:sensitivity-population}.

\paragraph{Population size.}
Table~\ref{tab:metaeval-l2-setsize} compares sensitivity across population sizes $n\in\{10,25,50,100\}$. With 10 sessions, the act-distribution family scores the control population higher in 85\% of draws for approve-broken-work and 80\% for front-loading, compared with 72\% for SimulatorArena's best metric in both cases.

\FloatBarrier

\begingroup
\SensitivityGridStyle
\begin{longtable}{@{}>{\raggedright\arraybackslash}p{86pt}*{8}{>{\centering\arraybackslash}p{\dimexpr(\linewidth-86pt-16\tabcolsep)/8\relax}}@{}}
\caption{Population-level win rates for each metric and perturbation. Colors run from red (0) through white (0.5) to green (1); ties count as half a win. Table~\ref{tab:sensitivity-set-diff} reports the corresponding score decreases and detection. Goal = adversarial goal; AllC = all constraints removed; Appr = approve broken work; Stop = premature stop; Pers = persona swap; Lang = language switch; CoD = co-debugger; Front = front-loading.}
\label{tab:sensitivity-set}\\
\toprule
\textbf{Metric} & \textbf{Goal} & \textbf{AllC} & \textbf{Appr} & \textbf{Stop} & \textbf{Pers} & \textbf{Lang} & \textbf{CoD} & \textbf{Front} \\
\midrule
\endfirsthead
\caption*{\textbf{Table~\ref{tab:sensitivity-set} (continued).}}\\
\toprule
\textbf{Metric} & \textbf{Goal} & \textbf{AllC} & \textbf{Appr} & \textbf{Stop} & \textbf{Pers} & \textbf{Lang} & \textbf{CoD} & \textbf{Front} \\
\midrule
\endhead
\midrule
\endfoot
\bottomrule
\endlastfoot
\multicolumn{9}{l}{\textbf{Ours, L2 population families (population metrics)}} \\*
act distribution & \cellcolor[HTML]{DDAFAF}\textcolor[HTML]{000000}{0.30} & \cellcolor[HTML]{F9F7F6}\textcolor[HTML]{000000}{0.49} & \cellcolor[HTML]{23754F}\textcolor[HTML]{FFFFFF}{1.00} & \cellcolor[HTML]{B13F42}\textcolor[HTML]{FFFFFF}{0.00} & \cellcolor[HTML]{F9F7F6}\textcolor[HTML]{000000}{0.49} & \cellcolor[HTML]{88B39E}\textcolor[HTML]{000000}{0.77} & \cellcolor[HTML]{9BBFAE}\textcolor[HTML]{000000}{0.72} & \cellcolor[HTML]{23754F}\textcolor[HTML]{FFFFFF}{1.00} \\
dynamics & \cellcolor[HTML]{F4EAE9}\textcolor[HTML]{000000}{0.46} & \cellcolor[HTML]{F0DFDF}\textcolor[HTML]{000000}{0.43} & \cellcolor[HTML]{E3BFC0}\textcolor[HTML]{000000}{0.34} & \cellcolor[HTML]{CF8B8D}\textcolor[HTML]{000000}{0.20} & \cellcolor[HTML]{E7EDE9}\textcolor[HTML]{000000}{0.55} & \cellcolor[HTML]{EED9D9}\textcolor[HTML]{000000}{0.41} & \cellcolor[HTML]{CD8687}\textcolor[HTML]{000000}{0.19} & \cellcolor[HTML]{E0B8B8}\textcolor[HTML]{000000}{0.33} \\
outcome sensitivity & \cellcolor[HTML]{F2E5E4}\textcolor[HTML]{000000}{0.45} & \cellcolor[HTML]{F9F7F6}\textcolor[HTML]{000000}{0.49} & \cellcolor[HTML]{C36C6E}\textcolor[HTML]{000000}{0.12} & \cellcolor[HTML]{BE6063}\textcolor[HTML]{000000}{0.09} & \cellcolor[HTML]{79AA93}\textcolor[HTML]{000000}{0.80} & \cellcolor[HTML]{6EA38A}\textcolor[HTML]{000000}{0.82} & \cellcolor[HTML]{669E83}\textcolor[HTML]{000000}{0.84} & \cellcolor[HTML]{7BAB94}\textcolor[HTML]{000000}{0.80} \\
provenance & \cellcolor[HTML]{EBF0EC}\textcolor[HTML]{000000}{0.54} & \cellcolor[HTML]{EBD2D2}\textcolor[HTML]{000000}{0.40} & \cellcolor[HTML]{CDDDD5}\textcolor[HTML]{000000}{0.60} & \cellcolor[HTML]{F1E3E2}\textcolor[HTML]{000000}{0.44} & \cellcolor[HTML]{509172}\textcolor[HTML]{000000}{0.90} & \cellcolor[HTML]{C4D8CE}\textcolor[HTML]{000000}{0.62} & \cellcolor[HTML]{3F8665}\textcolor[HTML]{000000}{0.94} & \cellcolor[HTML]{2E7C57}\textcolor[HTML]{FFFFFF}{0.97} \\
\textit{best metric} & \cellcolor[HTML]{EBF0EC}\textcolor[HTML]{000000}{\shortstack[c]{0.54\\{\tiny provenance}}} & \cellcolor[HTML]{F9F7F6}\textcolor[HTML]{000000}{\shortstack[c]{0.49\\{\tiny act}}} & \cellcolor[HTML]{23754F}\textcolor[HTML]{FFFFFF}{\shortstack[c]{1.00\\{\tiny act}}} & \cellcolor[HTML]{F1E3E2}\textcolor[HTML]{000000}{\shortstack[c]{0.44\\{\tiny provenance}}} & \cellcolor[HTML]{509172}\textcolor[HTML]{000000}{\shortstack[c]{0.90\\{\tiny provenance}}} & \cellcolor[HTML]{6EA38A}\textcolor[HTML]{000000}{\shortstack[c]{0.82\\{\tiny outcome}}} & \cellcolor[HTML]{3F8665}\textcolor[HTML]{000000}{\shortstack[c]{0.94\\{\tiny provenance}}} & \cellcolor[HTML]{23754F}\textcolor[HTML]{FFFFFF}{\shortstack[c]{1.00\\{\tiny act}}} \\
\textit{L2 (mean of families)} & \cellcolor[HTML]{ECD6D5}\textcolor[HTML]{000000}{0.41} & \cellcolor[HTML]{E9CECE}\textcolor[HTML]{000000}{0.39} & \cellcolor[HTML]{C4D8CE}\textcolor[HTML]{000000}{0.62} & \cellcolor[HTML]{B13F42}\textcolor[HTML]{FFFFFF}{0.00} & \cellcolor[HTML]{4C8E6F}\textcolor[HTML]{000000}{0.91} & \cellcolor[HTML]{99BEAC}\textcolor[HTML]{000000}{0.72} & \cellcolor[HTML]{488B6C}\textcolor[HTML]{000000}{0.92} & \cellcolor[HTML]{23754F}\textcolor[HTML]{FFFFFF}{1.00} \\
\addlinespace[3pt]
\multicolumn{9}{l}{\textbf{Ours, L3 trace families as population means}} \\*
persona & \cellcolor[HTML]{23754F}\textcolor[HTML]{FFFFFF}{1.00} & \cellcolor[HTML]{B6CFC3}\textcolor[HTML]{000000}{0.66} & \cellcolor[HTML]{23754F}\textcolor[HTML]{FFFFFF}{1.00} & \cellcolor[HTML]{3B8462}\textcolor[HTML]{000000}{0.94} & \cellcolor[HTML]{23754F}\textcolor[HTML]{FFFFFF}{1.00} & \cellcolor[HTML]{23754F}\textcolor[HTML]{FFFFFF}{1.00} & \cellcolor[HTML]{23754F}\textcolor[HTML]{FFFFFF}{1.00} & \cellcolor[HTML]{EACFCF}\textcolor[HTML]{000000}{0.39} \\
intent & \cellcolor[HTML]{23754F}\textcolor[HTML]{FFFFFF}{1.00} & \cellcolor[HTML]{408766}\textcolor[HTML]{000000}{0.93} & \cellcolor[HTML]{4F9072}\textcolor[HTML]{000000}{0.90} & \cellcolor[HTML]{257651}\textcolor[HTML]{FFFFFF}{0.99} & \cellcolor[HTML]{84B09B}\textcolor[HTML]{000000}{0.78} & \cellcolor[HTML]{257651}\textcolor[HTML]{FFFFFF}{0.99} & \cellcolor[HTML]{3A8361}\textcolor[HTML]{000000}{0.95} & \cellcolor[HTML]{468B6B}\textcolor[HTML]{000000}{0.92} \\
constraints & \cellcolor[HTML]{2C7A56}\textcolor[HTML]{FFFFFF}{0.98} & \cellcolor[HTML]{277852}\textcolor[HTML]{FFFFFF}{0.99} & \cellcolor[HTML]{E0B8B8}\textcolor[HTML]{000000}{0.33} & \cellcolor[HTML]{418767}\textcolor[HTML]{000000}{0.93} & \cellcolor[HTML]{7FAE98}\textcolor[HTML]{000000}{0.79} & \cellcolor[HTML]{D0DFD7}\textcolor[HTML]{000000}{0.60} & \cellcolor[HTML]{86B29D}\textcolor[HTML]{000000}{0.77} & \cellcolor[HTML]{CC8485}\textcolor[HTML]{000000}{0.18} \\
\textit{best metric} & \cellcolor[HTML]{23754F}\textcolor[HTML]{FFFFFF}{\shortstack[c]{1.00\\{\tiny persona}}} & \cellcolor[HTML]{277852}\textcolor[HTML]{FFFFFF}{\shortstack[c]{0.99\\{\tiny constraints}}} & \cellcolor[HTML]{23754F}\textcolor[HTML]{FFFFFF}{\shortstack[c]{1.00\\{\tiny persona}}} & \cellcolor[HTML]{257651}\textcolor[HTML]{FFFFFF}{\shortstack[c]{0.99\\{\tiny intent}}} & \cellcolor[HTML]{23754F}\textcolor[HTML]{FFFFFF}{\shortstack[c]{1.00\\{\tiny persona}}} & \cellcolor[HTML]{23754F}\textcolor[HTML]{FFFFFF}{\shortstack[c]{1.00\\{\tiny persona}}} & \cellcolor[HTML]{23754F}\textcolor[HTML]{FFFFFF}{\shortstack[c]{1.00\\{\tiny persona}}} & \cellcolor[HTML]{468B6B}\textcolor[HTML]{000000}{\shortstack[c]{0.92\\{\tiny intent}}} \\
\textit{L3 (mean of families)} & \cellcolor[HTML]{23754F}\textcolor[HTML]{FFFFFF}{1.00} & \cellcolor[HTML]{34805D}\textcolor[HTML]{FFFFFF}{0.96} & \cellcolor[HTML]{23754F}\textcolor[HTML]{FFFFFF}{1.00} & \cellcolor[HTML]{2C7A56}\textcolor[HTML]{FFFFFF}{0.98} & \cellcolor[HTML]{23754F}\textcolor[HTML]{FFFFFF}{1.00} & \cellcolor[HTML]{23754F}\textcolor[HTML]{FFFFFF}{1.00} & \cellcolor[HTML]{23754F}\textcolor[HTML]{FFFFFF}{1.00} & \cellcolor[HTML]{99BEAC}\textcolor[HTML]{000000}{0.72} \\
\addlinespace[3pt]
\multicolumn{9}{l}{\textbf{SimulatorArena (population means)}} \\*
Likert writing & \cellcolor[HTML]{EBD2D2}\textcolor[HTML]{000000}{0.40} & \cellcolor[HTML]{F0DFDF}\textcolor[HTML]{000000}{0.43} & \cellcolor[HTML]{4D8F70}\textcolor[HTML]{000000}{0.90} & \cellcolor[HTML]{B24144}\textcolor[HTML]{FFFFFF}{0.01} & \cellcolor[HTML]{23754F}\textcolor[HTML]{FFFFFF}{1.00} & \cellcolor[HTML]{23754F}\textcolor[HTML]{FFFFFF}{1.00} & \cellcolor[HTML]{277852}\textcolor[HTML]{FFFFFF}{0.99} & \cellcolor[HTML]{257651}\textcolor[HTML]{FFFFFF}{0.99} \\
Likert interaction & \cellcolor[HTML]{EEF2EF}\textcolor[HTML]{000000}{0.53} & \cellcolor[HTML]{CBDCD3}\textcolor[HTML]{000000}{0.61} & \cellcolor[HTML]{B24144}\textcolor[HTML]{FFFFFF}{0.01} & \cellcolor[HTML]{DBAAAB}\textcolor[HTML]{000000}{0.29} & \cellcolor[HTML]{448969}\textcolor[HTML]{000000}{0.92} & \cellcolor[HTML]{2E7C57}\textcolor[HTML]{FFFFFF}{0.97} & \cellcolor[HTML]{418767}\textcolor[HTML]{000000}{0.93} & \cellcolor[HTML]{34805D}\textcolor[HTML]{FFFFFF}{0.96} \\
document rating & \cellcolor[HTML]{69A086}\textcolor[HTML]{000000}{0.84} & \cellcolor[HTML]{F7F2F1}\textcolor[HTML]{000000}{0.48} & \cellcolor[HTML]{D7A1A2}\textcolor[HTML]{000000}{0.26} & \cellcolor[HTML]{B95355}\textcolor[HTML]{FFFFFF}{0.05} & \cellcolor[HTML]{B5CFC2}\textcolor[HTML]{000000}{0.66} & \cellcolor[HTML]{F1E2E1}\textcolor[HTML]{000000}{0.44} & \cellcolor[HTML]{D7E3DC}\textcolor[HTML]{000000}{0.58} & \cellcolor[HTML]{E8EEEA}\textcolor[HTML]{000000}{0.54} \\
interaction rating & \cellcolor[HTML]{F7F2F2}\textcolor[HTML]{000000}{0.48} & \cellcolor[HTML]{DDB0B0}\textcolor[HTML]{000000}{0.30} & \cellcolor[HTML]{B13F42}\textcolor[HTML]{FFFFFF}{0.00} & \cellcolor[HTML]{B44648}\textcolor[HTML]{FFFFFF}{0.02} & \cellcolor[HTML]{C46E70}\textcolor[HTML]{000000}{0.13} & \cellcolor[HTML]{F7F2F1}\textcolor[HTML]{000000}{0.48} & \cellcolor[HTML]{CE888A}\textcolor[HTML]{000000}{0.20} & \cellcolor[HTML]{E9EEEA}\textcolor[HTML]{000000}{0.54} \\
terminated early & \cellcolor[HTML]{E9EEEA}\textcolor[HTML]{000000}{0.54} & \cellcolor[HTML]{ADC9BB}\textcolor[HTML]{000000}{0.68} & \cellcolor[HTML]{BCD3C7}\textcolor[HTML]{000000}{0.65} & \cellcolor[HTML]{CC8586}\textcolor[HTML]{000000}{0.19} & \cellcolor[HTML]{FAF8F7}\textcolor[HTML]{000000}{0.50} & \cellcolor[HTML]{E6C7C7}\textcolor[HTML]{000000}{0.36} & \cellcolor[HTML]{F8F8F6}\textcolor[HTML]{000000}{0.51} & \cellcolor[HTML]{BCD3C7}\textcolor[HTML]{000000}{0.65} \\
Turing detected & \cellcolor[HTML]{C57274}\textcolor[HTML]{000000}{0.14} & \cellcolor[HTML]{EDF1EE}\textcolor[HTML]{000000}{0.53} & \cellcolor[HTML]{B34547}\textcolor[HTML]{FFFFFF}{0.01} & \cellcolor[HTML]{A0C2B1}\textcolor[HTML]{000000}{0.71} & \cellcolor[HTML]{DFB4B5}\textcolor[HTML]{000000}{0.32} & \cellcolor[HTML]{B85254}\textcolor[HTML]{FFFFFF}{0.05} & \cellcolor[HTML]{D9E5DE}\textcolor[HTML]{000000}{0.58} & \cellcolor[HTML]{5A977A}\textcolor[HTML]{000000}{0.87} \\
\textit{best metric} & \cellcolor[HTML]{69A086}\textcolor[HTML]{000000}{\shortstack[c]{0.84\\{\tiny document}}} & \cellcolor[HTML]{ADC9BB}\textcolor[HTML]{000000}{\shortstack[c]{0.68\\{\tiny terminated}}} & \cellcolor[HTML]{4D8F70}\textcolor[HTML]{000000}{\shortstack[c]{0.90\\{\tiny writing}}} & \cellcolor[HTML]{A0C2B1}\textcolor[HTML]{000000}{\shortstack[c]{0.71\\{\tiny Turing}}} & \cellcolor[HTML]{23754F}\textcolor[HTML]{FFFFFF}{\shortstack[c]{1.00\\{\tiny writing}}} & \cellcolor[HTML]{23754F}\textcolor[HTML]{FFFFFF}{\shortstack[c]{1.00\\{\tiny writing}}} & \cellcolor[HTML]{277852}\textcolor[HTML]{FFFFFF}{\shortstack[c]{0.99\\{\tiny writing}}} & \cellcolor[HTML]{257651}\textcolor[HTML]{FFFFFF}{\shortstack[c]{0.99\\{\tiny writing}}} \\
\addlinespace[3pt]
\multicolumn{9}{l}{\textbf{VISTA (population means)}} \\*
profile consistency & \cellcolor[HTML]{A4C4B4}\textcolor[HTML]{000000}{0.70} & \cellcolor[HTML]{E4ECE7}\textcolor[HTML]{000000}{0.55} & \cellcolor[HTML]{B13F42}\textcolor[HTML]{FFFFFF}{0.00} & \cellcolor[HTML]{B13F42}\textcolor[HTML]{FFFFFF}{0.00} & \cellcolor[HTML]{EFF2F0}\textcolor[HTML]{000000}{0.53} & \cellcolor[HTML]{23754F}\textcolor[HTML]{FFFFFF}{1.00} & \cellcolor[HTML]{549376}\textcolor[HTML]{000000}{0.89} & \cellcolor[HTML]{2C7A56}\textcolor[HTML]{FFFFFF}{0.98} \\
knowledge consistency & \cellcolor[HTML]{23754F}\textcolor[HTML]{FFFFFF}{1.00} & \cellcolor[HTML]{77A891}\textcolor[HTML]{000000}{0.81} & \cellcolor[HTML]{B13F42}\textcolor[HTML]{FFFFFF}{0.00} & \cellcolor[HTML]{B13F42}\textcolor[HTML]{FFFFFF}{0.00} & \cellcolor[HTML]{C77779}\textcolor[HTML]{000000}{0.15} & \cellcolor[HTML]{9BBFAE}\textcolor[HTML]{000000}{0.72} & \cellcolor[HTML]{669E83}\textcolor[HTML]{000000}{0.84} & \cellcolor[HTML]{77A891}\textcolor[HTML]{000000}{0.81} \\
goal consistency & \cellcolor[HTML]{23754F}\textcolor[HTML]{FFFFFF}{1.00} & \cellcolor[HTML]{F1F4F1}\textcolor[HTML]{000000}{0.52} & \cellcolor[HTML]{B24144}\textcolor[HTML]{FFFFFF}{0.01} & \cellcolor[HTML]{B74E51}\textcolor[HTML]{FFFFFF}{0.04} & \cellcolor[HTML]{C57375}\textcolor[HTML]{000000}{0.14} & \cellcolor[HTML]{C2D7CC}\textcolor[HTML]{000000}{0.63} & \cellcolor[HTML]{7DAC96}\textcolor[HTML]{000000}{0.79} & \cellcolor[HTML]{88B39E}\textcolor[HTML]{000000}{0.77} \\
\textit{best metric} & \cellcolor[HTML]{23754F}\textcolor[HTML]{FFFFFF}{\shortstack[c]{1.00\\{\tiny knowledge}}} & \cellcolor[HTML]{77A891}\textcolor[HTML]{000000}{\shortstack[c]{0.81\\{\tiny knowledge}}} & \cellcolor[HTML]{B24144}\textcolor[HTML]{FFFFFF}{\shortstack[c]{0.01\\{\tiny goal}}} & \cellcolor[HTML]{B74E51}\textcolor[HTML]{FFFFFF}{\shortstack[c]{0.04\\{\tiny goal}}} & \cellcolor[HTML]{EFF2F0}\textcolor[HTML]{000000}{\shortstack[c]{0.53\\{\tiny profile}}} & \cellcolor[HTML]{23754F}\textcolor[HTML]{FFFFFF}{\shortstack[c]{1.00\\{\tiny profile}}} & \cellcolor[HTML]{549376}\textcolor[HTML]{000000}{\shortstack[c]{0.89\\{\tiny profile}}} & \cellcolor[HTML]{2C7A56}\textcolor[HTML]{FFFFFF}{\shortstack[c]{0.98\\{\tiny profile}}} \\
\addlinespace[3pt]
\multicolumn{9}{l}{\textbf{SWE-Together (population means)}} \\*
coverage rate & \cellcolor[HTML]{B7D0C4}\textcolor[HTML]{000000}{0.66} & \cellcolor[HTML]{75A78F}\textcolor[HTML]{000000}{0.81} & \cellcolor[HTML]{D9A5A6}\textcolor[HTML]{000000}{0.28} & \cellcolor[HTML]{23754F}\textcolor[HTML]{FFFFFF}{1.00} & \cellcolor[HTML]{599679}\textcolor[HTML]{000000}{0.88} & \cellcolor[HTML]{75A78F}\textcolor[HTML]{000000}{0.81} & \cellcolor[HTML]{277852}\textcolor[HTML]{FFFFFF}{0.99} & \cellcolor[HTML]{69A086}\textcolor[HTML]{000000}{0.84} \\
weighted coverage & \cellcolor[HTML]{99BEAC}\textcolor[HTML]{000000}{0.72} & \cellcolor[HTML]{73A68E}\textcolor[HTML]{000000}{0.81} & \cellcolor[HTML]{CD8687}\textcolor[HTML]{000000}{0.19} & \cellcolor[HTML]{23754F}\textcolor[HTML]{FFFFFF}{1.00} & \cellcolor[HTML]{549376}\textcolor[HTML]{000000}{0.89} & \cellcolor[HTML]{6FA48B}\textcolor[HTML]{000000}{0.82} & \cellcolor[HTML]{2C7A56}\textcolor[HTML]{FFFFFF}{0.98} & \cellcolor[HTML]{5F9A7E}\textcolor[HTML]{000000}{0.86} \\
scope precision & \cellcolor[HTML]{E2BCBC}\textcolor[HTML]{000000}{0.34} & \cellcolor[HTML]{F4F5F3}\textcolor[HTML]{000000}{0.52} & \cellcolor[HTML]{F4EAE9}\textcolor[HTML]{000000}{0.46} & \cellcolor[HTML]{B13F42}\textcolor[HTML]{FFFFFF}{0.00} & \cellcolor[HTML]{C36C6E}\textcolor[HTML]{000000}{0.12} & \cellcolor[HTML]{BAD1C5}\textcolor[HTML]{000000}{0.65} & \cellcolor[HTML]{BA575A}\textcolor[HTML]{000000}{0.07} & \cellcolor[HTML]{5A977A}\textcolor[HTML]{000000}{0.87} \\
user correction & \cellcolor[HTML]{DFB5B6}\textcolor[HTML]{000000}{0.32} & \cellcolor[HTML]{E9CECE}\textcolor[HTML]{000000}{0.39} & \cellcolor[HTML]{B13F42}\textcolor[HTML]{FFFFFF}{0.00} & \cellcolor[HTML]{B13F42}\textcolor[HTML]{FFFFFF}{0.00} & \cellcolor[HTML]{D08D8E}\textcolor[HTML]{000000}{0.21} & \cellcolor[HTML]{EBD4D4}\textcolor[HTML]{000000}{0.40} & \cellcolor[HTML]{C46F71}\textcolor[HTML]{000000}{0.13} & \cellcolor[HTML]{E4C0C0}\textcolor[HTML]{000000}{0.35} \\
frustration count & \cellcolor[HTML]{E7C8C8}\textcolor[HTML]{000000}{0.37} & \cellcolor[HTML]{E0B6B6}\textcolor[HTML]{000000}{0.32} & \cellcolor[HTML]{B13F42}\textcolor[HTML]{FFFFFF}{0.00} & \cellcolor[HTML]{B13F42}\textcolor[HTML]{FFFFFF}{0.00} & \cellcolor[HTML]{C06466}\textcolor[HTML]{000000}{0.10} & \cellcolor[HTML]{EED9D9}\textcolor[HTML]{000000}{0.41} & \cellcolor[HTML]{C47072}\textcolor[HTML]{000000}{0.13} & \cellcolor[HTML]{AAC8B9}\textcolor[HTML]{000000}{0.69} \\
\textit{best metric} & \cellcolor[HTML]{99BEAC}\textcolor[HTML]{000000}{\shortstack[c]{0.72\\{\tiny weighted}}} & \cellcolor[HTML]{73A68E}\textcolor[HTML]{000000}{\shortstack[c]{0.81\\{\tiny weighted}}} & \cellcolor[HTML]{F4EAE9}\textcolor[HTML]{000000}{\shortstack[c]{0.46\\{\tiny scope}}} & \cellcolor[HTML]{23754F}\textcolor[HTML]{FFFFFF}{\shortstack[c]{1.00\\{\tiny coverage}}} & \cellcolor[HTML]{549376}\textcolor[HTML]{000000}{\shortstack[c]{0.89\\{\tiny weighted}}} & \cellcolor[HTML]{6FA48B}\textcolor[HTML]{000000}{\shortstack[c]{0.82\\{\tiny weighted}}} & \cellcolor[HTML]{277852}\textcolor[HTML]{FFFFFF}{\shortstack[c]{0.99\\{\tiny coverage}}} & \cellcolor[HTML]{5A977A}\textcolor[HTML]{000000}{\shortstack[c]{0.87\\{\tiny scope}}} \\
\addlinespace[3pt]
\multicolumn{9}{l}{\textbf{Sim2Real (population metric)}} \\*
behavioral index & \cellcolor[HTML]{AFCBBD}\textcolor[HTML]{000000}{0.68} & \cellcolor[HTML]{F4F5F3}\textcolor[HTML]{000000}{0.52} & \cellcolor[HTML]{EDF1EE}\textcolor[HTML]{000000}{0.53} & \cellcolor[HTML]{BED4C9}\textcolor[HTML]{000000}{0.64} & \cellcolor[HTML]{4C8E6F}\textcolor[HTML]{000000}{0.91} & \cellcolor[HTML]{E4C1C1}\textcolor[HTML]{000000}{0.35} & \cellcolor[HTML]{5F9A7E}\textcolor[HTML]{000000}{0.86} & \cellcolor[HTML]{A8C7B8}\textcolor[HTML]{000000}{0.69} \\
\addlinespace[3pt]
\end{longtable}
\endgroup

\clearpage
\begingroup
\SensitivityGridStyle
\begin{longtable}{@{}>{\raggedright\arraybackslash}m{86pt}*{4}{>{\centering\arraybackslash}m{\dimexpr(\linewidth-86pt-8\tabcolsep)/4\relax}}@{}}
\caption{Population score decreases corresponding to Table~\ref{tab:sensitivity-set}. Each cell reports the mean control-minus-perturbed difference above its 2.5th--97.5th percentile range, in the metric's original units. Green marks a range entirely above zero; red marks otherwise. Best-metric selections and abbreviations follow Table~\ref{tab:sensitivity-set}.}
\label{tab:sensitivity-set-diff}\\
\toprule
\textbf{Metric} & \textbf{Goal} & \textbf{AllC} & \textbf{Appr} & \textbf{Stop} \\
\midrule
\endfirsthead
\caption*{\textbf{Table~\ref{tab:sensitivity-set-diff} (continued).}}\\
\toprule
\textbf{Metric} & \textbf{Goal} & \textbf{AllC} & \textbf{Appr} & \textbf{Stop} \\
\midrule
\endhead
\midrule
\endfoot
\bottomrule
\endlastfoot
\multicolumn{5}{l}{\textbf{Ours, L2 population families (population metrics)}} \\*
act distribution & \cellcolor{ArenaTableUndetected}\makecell[c]{$-1.5$\\{\fontsize{6.6}{7.6}\selectfont$[-6.8,+4.0]$}} & \cellcolor{ArenaTableUndetected}\makecell[c]{$0.0$\\{\fontsize{6.6}{7.6}\selectfont$[-4.1,+4.6]$}} & \cellcolor{ArenaTableDetected}\makecell[c]{$+10.6$\\{\fontsize{6.6}{7.6}\selectfont$[+3.7,+16.8]$}} & \cellcolor{ArenaTableUndetected}\makecell[c]{$-18.0$\\{\fontsize{6.6}{7.6}\selectfont$[-25.4,-9.3]$}} \\
dynamics & \cellcolor{ArenaTableUndetected}\makecell[c]{$-0.2$\\{\fontsize{6.6}{7.6}\selectfont$[-7.3,+6.5]$}} & \cellcolor{ArenaTableUndetected}\makecell[c]{$-0.5$\\{\fontsize{6.6}{7.6}\selectfont$[-8.6,+7.2]$}} & \cellcolor{ArenaTableUndetected}\makecell[c]{$-2.1$\\{\fontsize{6.6}{7.6}\selectfont$[-12.9,+9.8]$}} & \cellcolor{ArenaTableUndetected}\makecell[c]{$-4.5$\\{\fontsize{6.6}{7.6}\selectfont$[-15.5,+7.8]$}} \\
outcome sensitivity & \cellcolor{ArenaTableUndetected}\makecell[c]{$-0.1$\\{\fontsize{6.6}{7.6}\selectfont$[-6.1,+6.8]$}} & \cellcolor{ArenaTableUndetected}\makecell[c]{$+0.1$\\{\fontsize{6.6}{7.6}\selectfont$[-5.1,+5.2]$}} & \cellcolor{ArenaTableUndetected}\makecell[c]{$-6.7$\\{\fontsize{6.6}{7.6}\selectfont$[-16.6,+7.0]$}} & \cellcolor{ArenaTableUndetected}\makecell[c]{$-6.7$\\{\fontsize{6.6}{7.6}\selectfont$[-14.2,+5.3]$}} \\
provenance & \cellcolor{ArenaTableUndetected}\makecell[c]{$+0.2$\\{\fontsize{6.6}{7.6}\selectfont$[-9.9,+11.0]$}} & \cellcolor{ArenaTableUndetected}\makecell[c]{$-1.1$\\{\fontsize{6.6}{7.6}\selectfont$[-12.6,+10.5]$}} & \cellcolor{ArenaTableUndetected}\makecell[c]{$+4.2$\\{\fontsize{6.6}{7.6}\selectfont$[-22.4,+24.6]$}} & \cellcolor{ArenaTableUndetected}\makecell[c]{$-1.5$\\{\fontsize{6.6}{7.6}\selectfont$[-14.0,+13.1]$}} \\
\textit{best metric} & \cellcolor{ArenaTableUndetected}\makecell[c]{$+0.2$\\{\fontsize{6.6}{7.6}\selectfont$[-9.9,+11.0]$}} & \cellcolor{ArenaTableUndetected}\makecell[c]{$0.0$\\{\fontsize{6.6}{7.6}\selectfont$[-4.1,+4.6]$}} & \cellcolor{ArenaTableDetected}\makecell[c]{$+10.6$\\{\fontsize{6.6}{7.6}\selectfont$[+3.7,+16.8]$}} & \cellcolor{ArenaTableUndetected}\makecell[c]{$-1.5$\\{\fontsize{6.6}{7.6}\selectfont$[-14.0,+13.1]$}} \\
\textit{L2 (mean of families)} & \cellcolor{ArenaTableUndetected}\makecell[c]{$-0.4$\\{\fontsize{6.6}{7.6}\selectfont$[-4.6,+3.9]$}} & \cellcolor{ArenaTableUndetected}\makecell[c]{$-0.4$\\{\fontsize{6.6}{7.6}\selectfont$[-4.6,+4.3]$}} & \cellcolor{ArenaTableUndetected}\makecell[c]{$+1.5$\\{\fontsize{6.6}{7.6}\selectfont$[-6.8,+9.9]$}} & \cellcolor{ArenaTableUndetected}\makecell[c]{$-7.7$\\{\fontsize{6.6}{7.6}\selectfont$[-13.0,-0.9]$}} \\
\addlinespace[3pt]
\multicolumn{5}{l}{\textbf{Ours, L3 trace families as population means}} \\*
persona & \cellcolor{ArenaTableDetected}\makecell[c]{$+0.069$\\{\fontsize{6.6}{7.6}\selectfont$[+0.040,+0.103]$}} & \cellcolor{ArenaTableUndetected}\makecell[c]{$+0.006$\\{\fontsize{6.6}{7.6}\selectfont$[-0.024,+0.036]$}} & \cellcolor{ArenaTableDetected}\makecell[c]{$+0.052$\\{\fontsize{6.6}{7.6}\selectfont$[+0.018,+0.084]$}} & \cellcolor{ArenaTableUndetected}\makecell[c]{$+0.022$\\{\fontsize{6.6}{7.6}\selectfont$[-0.003,+0.053]$}} \\
intent & \cellcolor{ArenaTableDetected}\makecell[c]{$+0.264$\\{\fontsize{6.6}{7.6}\selectfont$[+0.170,+0.351]$}} & \cellcolor{ArenaTableUndetected}\makecell[c]{$+0.051$\\{\fontsize{6.6}{7.6}\selectfont$[-0.012,+0.126]$}} & \cellcolor{ArenaTableUndetected}\makecell[c]{$+0.059$\\{\fontsize{6.6}{7.6}\selectfont$[-0.024,+0.148]$}} & \cellcolor{ArenaTableDetected}\makecell[c]{$+0.091$\\{\fontsize{6.6}{7.6}\selectfont$[+0.022,+0.160]$}} \\
constraints & \cellcolor{ArenaTableDetected}\makecell[c]{$+0.086$\\{\fontsize{6.6}{7.6}\selectfont$[+0.014,+0.150]$}} & \cellcolor{ArenaTableDetected}\makecell[c]{$+0.068$\\{\fontsize{6.6}{7.6}\selectfont$[+0.003,+0.137]$}} & \cellcolor{ArenaTableUndetected}\makecell[c]{$-0.018$\\{\fontsize{6.6}{7.6}\selectfont$[-0.095,+0.061]$}} & \cellcolor{ArenaTableUndetected}\makecell[c]{$+0.055$\\{\fontsize{6.6}{7.6}\selectfont$[-0.021,+0.126]$}} \\
\textit{best metric} & \cellcolor{ArenaTableDetected}\makecell[c]{$+0.069$\\{\fontsize{6.6}{7.6}\selectfont$[+0.040,+0.103]$}} & \cellcolor{ArenaTableDetected}\makecell[c]{$+0.068$\\{\fontsize{6.6}{7.6}\selectfont$[+0.003,+0.137]$}} & \cellcolor{ArenaTableDetected}\makecell[c]{$+0.052$\\{\fontsize{6.6}{7.6}\selectfont$[+0.018,+0.084]$}} & \cellcolor{ArenaTableDetected}\makecell[c]{$+0.091$\\{\fontsize{6.6}{7.6}\selectfont$[+0.022,+0.160]$}} \\
\textit{L3 (mean of families)} & \cellcolor{ArenaTableDetected}\makecell[c]{$+0.152$\\{\fontsize{6.6}{7.6}\selectfont$[+0.100,+0.198]$}} & \cellcolor{ArenaTableUndetected}\makecell[c]{$+0.035$\\{\fontsize{6.6}{7.6}\selectfont$[-0.005,+0.075]$}} & \cellcolor{ArenaTableDetected}\makecell[c]{$+0.053$\\{\fontsize{6.6}{7.6}\selectfont$[+0.013,+0.094]$}} & \cellcolor{ArenaTableDetected}\makecell[c]{$+0.045$\\{\fontsize{6.6}{7.6}\selectfont$[+0.001,+0.092]$}} \\
\addlinespace[3pt]
\multicolumn{5}{l}{\textbf{SimulatorArena (population means)}} \\*
Likert writing & \cellcolor{ArenaTableUndetected}\makecell[c]{$-0.02$\\{\fontsize{6.6}{7.6}\selectfont$[-0.15,+0.11]$}} & \cellcolor{ArenaTableUndetected}\makecell[c]{$-0.02$\\{\fontsize{6.6}{7.6}\selectfont$[-0.13,+0.12]$}} & \cellcolor{ArenaTableUndetected}\makecell[c]{$+0.09$\\{\fontsize{6.6}{7.6}\selectfont$[-0.03,+0.22]$}} & \cellcolor{ArenaTableUndetected}\makecell[c]{$-0.24$\\{\fontsize{6.6}{7.6}\selectfont$[-0.40,-0.07]$}} \\
Likert interaction & \cellcolor{ArenaTableUndetected}\makecell[c]{$+0.01$\\{\fontsize{6.6}{7.6}\selectfont$[-0.18,+0.16]$}} & \cellcolor{ArenaTableUndetected}\makecell[c]{$+0.02$\\{\fontsize{6.6}{7.6}\selectfont$[-0.12,+0.17]$}} & \cellcolor{ArenaTableUndetected}\makecell[c]{$-0.27$\\{\fontsize{6.6}{7.6}\selectfont$[-0.41,-0.10]$}} & \cellcolor{ArenaTableUndetected}\makecell[c]{$-0.05$\\{\fontsize{6.6}{7.6}\selectfont$[-0.22,+0.11]$}} \\
document rating & \cellcolor{ArenaTableUndetected}\makecell[c]{$+0.19$\\{\fontsize{6.6}{7.6}\selectfont$[-0.27,+0.57]$}} & \cellcolor{ArenaTableUndetected}\makecell[c]{$+0.01$\\{\fontsize{6.6}{7.6}\selectfont$[-0.39,+0.48]$}} & \cellcolor{ArenaTableUndetected}\makecell[c]{$-0.16$\\{\fontsize{6.6}{7.6}\selectfont$[-0.63,+0.32]$}} & \cellcolor{ArenaTableUndetected}\makecell[c]{$-0.34$\\{\fontsize{6.6}{7.6}\selectfont$[-0.80,+0.13]$}} \\
interaction rating & \cellcolor{ArenaTableUndetected}\makecell[c]{$0.00$\\{\fontsize{6.6}{7.6}\selectfont$[-0.25,+0.25]$}} & \cellcolor{ArenaTableUndetected}\makecell[c]{$-0.08$\\{\fontsize{6.6}{7.6}\selectfont$[-0.31,+0.19]$}} & \cellcolor{ArenaTableUndetected}\makecell[c]{$-1.02$\\{\fontsize{6.6}{7.6}\selectfont$[-1.28,-0.78]$}} & \cellcolor{ArenaTableUndetected}\makecell[c]{$-0.27$\\{\fontsize{6.6}{7.6}\selectfont$[-0.54,-0.02]$}} \\
terminated early & \cellcolor{ArenaTableUndetected}\makecell[c]{$+0.001$\\{\fontsize{6.6}{7.6}\selectfont$[-0.030,+0.030]$}} & \cellcolor{ArenaTableUndetected}\makecell[c]{$+0.007$\\{\fontsize{6.6}{7.6}\selectfont$[-0.020,+0.040]$}} & \cellcolor{ArenaTableUndetected}\makecell[c]{$+0.005$\\{\fontsize{6.6}{7.6}\selectfont$[-0.030,+0.031]$}} & \cellcolor{ArenaTableUndetected}\makecell[c]{$-0.010$\\{\fontsize{6.6}{7.6}\selectfont$[-0.030,+0.010]$}} \\
Turing detected & \cellcolor{ArenaTableUndetected}\makecell[c]{$-0.057$\\{\fontsize{6.6}{7.6}\selectfont$[-0.143,+0.062]$}} & \cellcolor{ArenaTableUndetected}\makecell[c]{$+0.003$\\{\fontsize{6.6}{7.6}\selectfont$[-0.092,+0.120]$}} & \cellcolor{ArenaTableUndetected}\makecell[c]{$-0.086$\\{\fontsize{6.6}{7.6}\selectfont$[-0.173,0.000]$}} & \cellcolor{ArenaTableUndetected}\makecell[c]{$+0.026$\\{\fontsize{6.6}{7.6}\selectfont$[-0.061,+0.140]$}} \\
\textit{best metric} & \cellcolor{ArenaTableUndetected}\makecell[c]{$+0.19$\\{\fontsize{6.6}{7.6}\selectfont$[-0.27,+0.57]$}} & \cellcolor{ArenaTableUndetected}\makecell[c]{$+0.007$\\{\fontsize{6.6}{7.6}\selectfont$[-0.020,+0.040]$}} & \cellcolor{ArenaTableUndetected}\makecell[c]{$+0.09$\\{\fontsize{6.6}{7.6}\selectfont$[-0.03,+0.22]$}} & \cellcolor{ArenaTableUndetected}\makecell[c]{$+0.026$\\{\fontsize{6.6}{7.6}\selectfont$[-0.061,+0.140]$}} \\
\addlinespace[3pt]
\multicolumn{5}{l}{\textbf{VISTA (population means)}} \\*
profile consistency & \cellcolor{ArenaTableUndetected}\makecell[c]{$+0.010$\\{\fontsize{6.6}{7.6}\selectfont$[-0.021,+0.050]$}} & \cellcolor{ArenaTableUndetected}\makecell[c]{$+0.002$\\{\fontsize{6.6}{7.6}\selectfont$[-0.026,+0.035]$}} & \cellcolor{ArenaTableUndetected}\makecell[c]{$-0.071$\\{\fontsize{6.6}{7.6}\selectfont$[-0.110,-0.037]$}} & \cellcolor{ArenaTableUndetected}\makecell[c]{$-0.050$\\{\fontsize{6.6}{7.6}\selectfont$[-0.081,-0.020]$}} \\
knowledge consistency & \cellcolor{ArenaTableDetected}\makecell[c]{$+0.093$\\{\fontsize{6.6}{7.6}\selectfont$[+0.043,+0.149]$}} & \cellcolor{ArenaTableUndetected}\makecell[c]{$+0.016$\\{\fontsize{6.6}{7.6}\selectfont$[-0.024,+0.058]$}} & \cellcolor{ArenaTableUndetected}\makecell[c]{$-0.054$\\{\fontsize{6.6}{7.6}\selectfont$[-0.090,-0.020]$}} & \cellcolor{ArenaTableUndetected}\makecell[c]{$-0.057$\\{\fontsize{6.6}{7.6}\selectfont$[-0.091,-0.020]$}} \\
goal consistency & \cellcolor{ArenaTableDetected}\makecell[c]{$+0.058$\\{\fontsize{6.6}{7.6}\selectfont$[+0.026,+0.090]$}} & \cellcolor{ArenaTableUndetected}\makecell[c]{$+0.001$\\{\fontsize{6.6}{7.6}\selectfont$[-0.022,+0.026]$}} & \cellcolor{ArenaTableUndetected}\makecell[c]{$-0.037$\\{\fontsize{6.6}{7.6}\selectfont$[-0.057,-0.016]$}} & \cellcolor{ArenaTableUndetected}\makecell[c]{$-0.022$\\{\fontsize{6.6}{7.6}\selectfont$[-0.046,+0.005]$}} \\
\textit{best metric} & \cellcolor{ArenaTableDetected}\makecell[c]{$+0.093$\\{\fontsize{6.6}{7.6}\selectfont$[+0.043,+0.149]$}} & \cellcolor{ArenaTableUndetected}\makecell[c]{$+0.016$\\{\fontsize{6.6}{7.6}\selectfont$[-0.024,+0.058]$}} & \cellcolor{ArenaTableUndetected}\makecell[c]{$-0.037$\\{\fontsize{6.6}{7.6}\selectfont$[-0.057,-0.016]$}} & \cellcolor{ArenaTableUndetected}\makecell[c]{$-0.022$\\{\fontsize{6.6}{7.6}\selectfont$[-0.046,+0.005]$}} \\
\addlinespace[3pt]
\multicolumn{5}{l}{\textbf{SWE-Together (population means)}} \\*
coverage rate & \cellcolor{ArenaTableUndetected}\makecell[c]{$+0.009$\\{\fontsize{6.6}{7.6}\selectfont$[-0.042,+0.062]$}} & \cellcolor{ArenaTableUndetected}\makecell[c]{$+0.021$\\{\fontsize{6.6}{7.6}\selectfont$[-0.025,+0.072]$}} & \cellcolor{ArenaTableUndetected}\makecell[c]{$-0.017$\\{\fontsize{6.6}{7.6}\selectfont$[-0.076,+0.043]$}} & \cellcolor{ArenaTableDetected}\makecell[c]{$+0.175$\\{\fontsize{6.6}{7.6}\selectfont$[+0.121,+0.241]$}} \\
weighted coverage & \cellcolor{ArenaTableUndetected}\makecell[c]{$+0.011$\\{\fontsize{6.6}{7.6}\selectfont$[-0.035,+0.047]$}} & \cellcolor{ArenaTableUndetected}\makecell[c]{$+0.018$\\{\fontsize{6.6}{7.6}\selectfont$[-0.018,+0.060]$}} & \cellcolor{ArenaTableUndetected}\makecell[c]{$-0.024$\\{\fontsize{6.6}{7.6}\selectfont$[-0.068,+0.032]$}} & \cellcolor{ArenaTableDetected}\makecell[c]{$+0.151$\\{\fontsize{6.6}{7.6}\selectfont$[+0.107,+0.206]$}} \\
scope precision & \cellcolor{ArenaTableUndetected}\makecell[c]{$-0.008$\\{\fontsize{6.6}{7.6}\selectfont$[-0.056,+0.041]$}} & \cellcolor{ArenaTableUndetected}\makecell[c]{$+0.001$\\{\fontsize{6.6}{7.6}\selectfont$[-0.044,+0.044]$}} & \cellcolor{ArenaTableUndetected}\makecell[c]{$-0.003$\\{\fontsize{6.6}{7.6}\selectfont$[-0.049,+0.044]$}} & \cellcolor{ArenaTableUndetected}\makecell[c]{$-0.215$\\{\fontsize{6.6}{7.6}\selectfont$[-0.273,-0.152]$}} \\
user correction & \cellcolor{ArenaTableUndetected}\makecell[c]{$-0.45$\\{\fontsize{6.6}{7.6}\selectfont$[-2.18,+1.19]$}} & \cellcolor{ArenaTableUndetected}\makecell[c]{$-0.24$\\{\fontsize{6.6}{7.6}\selectfont$[-1.60,+1.30]$}} & \cellcolor{ArenaTableUndetected}\makecell[c]{$-7.10$\\{\fontsize{6.6}{7.6}\selectfont$[-8.83,-5.62]$}} & \cellcolor{ArenaTableUndetected}\makecell[c]{$-6.03$\\{\fontsize{6.6}{7.6}\selectfont$[-7.53,-4.56]$}} \\
frustration count & \cellcolor{ArenaTableUndetected}\makecell[c]{$-0.29$\\{\fontsize{6.6}{7.6}\selectfont$[-1.87,+1.11]$}} & \cellcolor{ArenaTableUndetected}\makecell[c]{$-0.35$\\{\fontsize{6.6}{7.6}\selectfont$[-2.10,+1.18]$}} & \cellcolor{ArenaTableUndetected}\makecell[c]{$-4.73$\\{\fontsize{6.6}{7.6}\selectfont$[-6.38,-3.19]$}} & \cellcolor{ArenaTableUndetected}\makecell[c]{$-4.26$\\{\fontsize{6.6}{7.6}\selectfont$[-5.78,-2.69]$}} \\
\textit{best metric} & \cellcolor{ArenaTableUndetected}\makecell[c]{$+0.011$\\{\fontsize{6.6}{7.6}\selectfont$[-0.035,+0.047]$}} & \cellcolor{ArenaTableUndetected}\makecell[c]{$+0.018$\\{\fontsize{6.6}{7.6}\selectfont$[-0.018,+0.060]$}} & \cellcolor{ArenaTableUndetected}\makecell[c]{$-0.003$\\{\fontsize{6.6}{7.6}\selectfont$[-0.049,+0.044]$}} & \cellcolor{ArenaTableDetected}\makecell[c]{$+0.175$\\{\fontsize{6.6}{7.6}\selectfont$[+0.121,+0.241]$}} \\
\addlinespace[3pt]
\multicolumn{5}{l}{\textbf{Sim2Real (population metric)}} \\*
behavioral index & \cellcolor{ArenaTableUndetected}\makecell[c]{$+1.1$\\{\fontsize{6.6}{7.6}\selectfont$[-5.1,+7.2]$}} & \cellcolor{ArenaTableUndetected}\makecell[c]{$0.0$\\{\fontsize{6.6}{7.6}\selectfont$[-8.4,+8.0]$}} & \cellcolor{ArenaTableUndetected}\makecell[c]{$+0.7$\\{\fontsize{6.6}{7.6}\selectfont$[-11.5,+14.5]$}} & \cellcolor{ArenaTableUndetected}\makecell[c]{$+1.6$\\{\fontsize{6.6}{7.6}\selectfont$[-7.8,+12.1]$}} \\
\addlinespace[3pt]
\end{longtable}
\edef\PopulationDifferenceTableNumber{\number\value{table}}
\clearpage
\begin{longtable*}{@{}>{\raggedright\arraybackslash}m{86pt}*{4}{>{\centering\arraybackslash}m{\dimexpr(\linewidth-86pt-8\tabcolsep)/4\relax}}@{}}
\caption*{\textbf{Table~\ref{tab:sensitivity-set-diff} (continued).}}\\
\toprule
\textbf{Metric} & \textbf{Pers} & \textbf{Lang} & \textbf{CoD} & \textbf{Front} \\
\midrule
\endfirsthead
\caption*{\textbf{Table~\ref{tab:sensitivity-set-diff} (continued).}}\\
\toprule
\textbf{Metric} & \textbf{Pers} & \textbf{Lang} & \textbf{CoD} & \textbf{Front} \\
\midrule
\endhead
\midrule
\endfoot
\bottomrule
\endlastfoot
\multicolumn{5}{l}{\textbf{Ours, L2 population families (population metrics)}} \\*
act distribution & \cellcolor{ArenaTableUndetected}\makecell[c]{$+0.1$\\{\fontsize{6.6}{7.6}\selectfont$[-5.5,+6.6]$}} & \cellcolor{ArenaTableUndetected}\makecell[c]{$+1.6$\\{\fontsize{6.6}{7.6}\selectfont$[-2.7,+6.1]$}} & \cellcolor{ArenaTableUndetected}\makecell[c]{$+1.8$\\{\fontsize{6.6}{7.6}\selectfont$[-3.4,+7.2]$}} & \cellcolor{ArenaTableDetected}\makecell[c]{$+6.8$\\{\fontsize{6.6}{7.6}\selectfont$[+1.9,+12.0]$}} \\
dynamics & \cellcolor{ArenaTableUndetected}\makecell[c]{$+1.3$\\{\fontsize{6.6}{7.6}\selectfont$[-8.6,+13.2]$}} & \cellcolor{ArenaTableUndetected}\makecell[c]{$-0.6$\\{\fontsize{6.6}{7.6}\selectfont$[-9.6,+10.5]$}} & \cellcolor{ArenaTableUndetected}\makecell[c]{$-4.2$\\{\fontsize{6.6}{7.6}\selectfont$[-14.4,+7.4]$}} & \cellcolor{ArenaTableUndetected}\makecell[c]{$-2.2$\\{\fontsize{6.6}{7.6}\selectfont$[-12.2,+7.8]$}} \\
outcome sensitivity & \cellcolor{ArenaTableUndetected}\makecell[c]{$+3.1$\\{\fontsize{6.6}{7.6}\selectfont$[-2.1,+9.7]$}} & \cellcolor{ArenaTableUndetected}\makecell[c]{$+2.7$\\{\fontsize{6.6}{7.6}\selectfont$[-2.5,+8.0]$}} & \cellcolor{ArenaTableUndetected}\makecell[c]{$+3.5$\\{\fontsize{6.6}{7.6}\selectfont$[-2.6,+10.1]$}} & \cellcolor{ArenaTableUndetected}\makecell[c]{$+2.8$\\{\fontsize{6.6}{7.6}\selectfont$[-4.9,+9.9]$}} \\
provenance & \cellcolor{ArenaTableUndetected}\makecell[c]{$+9.3$\\{\fontsize{6.6}{7.6}\selectfont$[-12.6,+20.0]$}} & \cellcolor{ArenaTableUndetected}\makecell[c]{$+1.5$\\{\fontsize{6.6}{7.6}\selectfont$[-9.1,+13.0]$}} & \cellcolor{ArenaTableUndetected}\makecell[c]{$+13.7$\\{\fontsize{6.6}{7.6}\selectfont$[-6.7,+26.0]$}} & \cellcolor{ArenaTableDetected}\makecell[c]{$+17.1$\\{\fontsize{6.6}{7.6}\selectfont$[+0.6,+27.2]$}} \\
\textit{best metric} & \cellcolor{ArenaTableUndetected}\makecell[c]{$+9.3$\\{\fontsize{6.6}{7.6}\selectfont$[-12.6,+20.0]$}} & \cellcolor{ArenaTableUndetected}\makecell[c]{$+2.7$\\{\fontsize{6.6}{7.6}\selectfont$[-2.5,+8.0]$}} & \cellcolor{ArenaTableUndetected}\makecell[c]{$+13.7$\\{\fontsize{6.6}{7.6}\selectfont$[-6.7,+26.0]$}} & \cellcolor{ArenaTableDetected}\makecell[c]{$+6.8$\\{\fontsize{6.6}{7.6}\selectfont$[+1.9,+12.0]$}} \\
\textit{L2 (mean of families)} & \cellcolor{ArenaTableUndetected}\makecell[c]{$+3.4$\\{\fontsize{6.6}{7.6}\selectfont$[-1.5,+9.2]$}} & \cellcolor{ArenaTableUndetected}\makecell[c]{$+1.3$\\{\fontsize{6.6}{7.6}\selectfont$[-2.1,+6.2]$}} & \cellcolor{ArenaTableUndetected}\makecell[c]{$+3.7$\\{\fontsize{6.6}{7.6}\selectfont$[-2.1,+8.4]$}} & \cellcolor{ArenaTableDetected}\makecell[c]{$+6.1$\\{\fontsize{6.6}{7.6}\selectfont$[+1.6,+10.6]$}} \\
\addlinespace[3pt]
\multicolumn{5}{l}{\textbf{Ours, L3 trace families as population means}} \\*
persona & \cellcolor{ArenaTableDetected}\makecell[c]{$+0.095$\\{\fontsize{6.6}{7.6}\selectfont$[+0.064,+0.124]$}} & \cellcolor{ArenaTableDetected}\makecell[c]{$+0.116$\\{\fontsize{6.6}{7.6}\selectfont$[+0.083,+0.147]$}} & \cellcolor{ArenaTableDetected}\makecell[c]{$+0.064$\\{\fontsize{6.6}{7.6}\selectfont$[+0.035,+0.097]$}} & \cellcolor{ArenaTableUndetected}\makecell[c]{$-0.005$\\{\fontsize{6.6}{7.6}\selectfont$[-0.034,+0.028]$}} \\
intent & \cellcolor{ArenaTableUndetected}\makecell[c]{$+0.029$\\{\fontsize{6.6}{7.6}\selectfont$[-0.045,+0.095]$}} & \cellcolor{ArenaTableDetected}\makecell[c]{$+0.118$\\{\fontsize{6.6}{7.6}\selectfont$[+0.045,+0.205]$}} & \cellcolor{ArenaTableUndetected}\makecell[c]{$+0.058$\\{\fontsize{6.6}{7.6}\selectfont$[-0.011,+0.128]$}} & \cellcolor{ArenaTableUndetected}\makecell[c]{$+0.060$\\{\fontsize{6.6}{7.6}\selectfont$[-0.012,+0.144]$}} \\
constraints & \cellcolor{ArenaTableUndetected}\makecell[c]{$+0.025$\\{\fontsize{6.6}{7.6}\selectfont$[-0.054,+0.087]$}} & \cellcolor{ArenaTableUndetected}\makecell[c]{$+0.003$\\{\fontsize{6.6}{7.6}\selectfont$[-0.073,+0.074]$}} & \cellcolor{ArenaTableUndetected}\makecell[c]{$+0.020$\\{\fontsize{6.6}{7.6}\selectfont$[-0.044,+0.080]$}} & \cellcolor{ArenaTableUndetected}\makecell[c]{$-0.027$\\{\fontsize{6.6}{7.6}\selectfont$[-0.095,+0.041]$}} \\
\textit{best metric} & \cellcolor{ArenaTableDetected}\makecell[c]{$+0.095$\\{\fontsize{6.6}{7.6}\selectfont$[+0.064,+0.124]$}} & \cellcolor{ArenaTableDetected}\makecell[c]{$+0.116$\\{\fontsize{6.6}{7.6}\selectfont$[+0.083,+0.147]$}} & \cellcolor{ArenaTableDetected}\makecell[c]{$+0.064$\\{\fontsize{6.6}{7.6}\selectfont$[+0.035,+0.097]$}} & \cellcolor{ArenaTableUndetected}\makecell[c]{$+0.060$\\{\fontsize{6.6}{7.6}\selectfont$[-0.012,+0.144]$}} \\
\textit{L3 (mean of families)} & \cellcolor{ArenaTableDetected}\makecell[c]{$+0.063$\\{\fontsize{6.6}{7.6}\selectfont$[+0.025,+0.101]$}} & \cellcolor{ArenaTableDetected}\makecell[c]{$+0.103$\\{\fontsize{6.6}{7.6}\selectfont$[+0.066,+0.143]$}} & \cellcolor{ArenaTableDetected}\makecell[c]{$+0.053$\\{\fontsize{6.6}{7.6}\selectfont$[+0.014,+0.092]$}} & \cellcolor{ArenaTableUndetected}\makecell[c]{$+0.013$\\{\fontsize{6.6}{7.6}\selectfont$[-0.029,+0.061]$}} \\
\addlinespace[3pt]
\multicolumn{5}{l}{\textbf{SimulatorArena (population means)}} \\*
Likert writing & \cellcolor{ArenaTableDetected}\makecell[c]{$+0.25$\\{\fontsize{6.6}{7.6}\selectfont$[+0.09,+0.44]$}} & \cellcolor{ArenaTableDetected}\makecell[c]{$+0.53$\\{\fontsize{6.6}{7.6}\selectfont$[+0.35,+0.69]$}} & \cellcolor{ArenaTableDetected}\makecell[c]{$+0.14$\\{\fontsize{6.6}{7.6}\selectfont$[+0.04,+0.28]$}} & \cellcolor{ArenaTableDetected}\makecell[c]{$+0.16$\\{\fontsize{6.6}{7.6}\selectfont$[+0.03,+0.30]$}} \\
Likert interaction & \cellcolor{ArenaTableUndetected}\makecell[c]{$+0.12$\\{\fontsize{6.6}{7.6}\selectfont$[-0.02,+0.30]$}} & \cellcolor{ArenaTableDetected}\makecell[c]{$+0.14$\\{\fontsize{6.6}{7.6}\selectfont$[+0.01,+0.30]$}} & \cellcolor{ArenaTableUndetected}\makecell[c]{$+0.11$\\{\fontsize{6.6}{7.6}\selectfont$[-0.03,+0.27]$}} & \cellcolor{ArenaTableUndetected}\makecell[c]{$+0.14$\\{\fontsize{6.6}{7.6}\selectfont$[-0.01,+0.29]$}} \\
document rating & \cellcolor{ArenaTableUndetected}\makecell[c]{$+0.11$\\{\fontsize{6.6}{7.6}\selectfont$[-0.32,+0.54]$}} & \cellcolor{ArenaTableUndetected}\makecell[c]{$-0.03$\\{\fontsize{6.6}{7.6}\selectfont$[-0.48,+0.35]$}} & \cellcolor{ArenaTableUndetected}\makecell[c]{$+0.06$\\{\fontsize{6.6}{7.6}\selectfont$[-0.39,+0.50]$}} & \cellcolor{ArenaTableUndetected}\makecell[c]{$0.00$\\{\fontsize{6.6}{7.6}\selectfont$[-0.40,+0.39]$}} \\
interaction rating & \cellcolor{ArenaTableUndetected}\makecell[c]{$-0.14$\\{\fontsize{6.6}{7.6}\selectfont$[-0.36,+0.13]$}} & \cellcolor{ArenaTableUndetected}\makecell[c]{$0.00$\\{\fontsize{6.6}{7.6}\selectfont$[-0.23,+0.24]$}} & \cellcolor{ArenaTableUndetected}\makecell[c]{$-0.10$\\{\fontsize{6.6}{7.6}\selectfont$[-0.33,+0.16]$}} & \cellcolor{ArenaTableUndetected}\makecell[c]{$+0.01$\\{\fontsize{6.6}{7.6}\selectfont$[-0.24,+0.28]$}} \\
terminated early & \cellcolor{ArenaTableUndetected}\makecell[c]{$0.000$\\{\fontsize{6.6}{7.6}\selectfont$[-0.020,+0.020]$}} & \cellcolor{ArenaTableUndetected}\makecell[c]{$-0.005$\\{\fontsize{6.6}{7.6}\selectfont$[-0.030,+0.020]$}} & \cellcolor{ArenaTableUndetected}\makecell[c]{$0.000$\\{\fontsize{6.6}{7.6}\selectfont$[-0.020,+0.020]$}} & \cellcolor{ArenaTableUndetected}\makecell[c]{$+0.006$\\{\fontsize{6.6}{7.6}\selectfont$[-0.020,+0.031]$}} \\
Turing detected & \cellcolor{ArenaTableUndetected}\makecell[c]{$-0.024$\\{\fontsize{6.6}{7.6}\selectfont$[-0.122,+0.070]$}} & \cellcolor{ArenaTableUndetected}\makecell[c]{$-0.073$\\{\fontsize{6.6}{7.6}\selectfont$[-0.163,+0.030]$}} & \cellcolor{ArenaTableUndetected}\makecell[c]{$+0.006$\\{\fontsize{6.6}{7.6}\selectfont$[-0.091,+0.100]$}} & \cellcolor{ArenaTableUndetected}\makecell[c]{$+0.054$\\{\fontsize{6.6}{7.6}\selectfont$[-0.041,+0.144]$}} \\
\textit{best metric} & \cellcolor{ArenaTableDetected}\makecell[c]{$+0.25$\\{\fontsize{6.6}{7.6}\selectfont$[+0.09,+0.44]$}} & \cellcolor{ArenaTableDetected}\makecell[c]{$+0.53$\\{\fontsize{6.6}{7.6}\selectfont$[+0.35,+0.69]$}} & \cellcolor{ArenaTableDetected}\makecell[c]{$+0.14$\\{\fontsize{6.6}{7.6}\selectfont$[+0.04,+0.28]$}} & \cellcolor{ArenaTableDetected}\makecell[c]{$+0.16$\\{\fontsize{6.6}{7.6}\selectfont$[+0.03,+0.30]$}} \\
\addlinespace[3pt]
\multicolumn{5}{l}{\textbf{VISTA (population means)}} \\*
profile consistency & \cellcolor{ArenaTableUndetected}\makecell[c]{$+0.003$\\{\fontsize{6.6}{7.6}\selectfont$[-0.033,+0.050]$}} & \cellcolor{ArenaTableDetected}\makecell[c]{$+0.095$\\{\fontsize{6.6}{7.6}\selectfont$[+0.052,+0.144]$}} & \cellcolor{ArenaTableUndetected}\makecell[c]{$+0.032$\\{\fontsize{6.6}{7.6}\selectfont$[-0.013,+0.080]$}} & \cellcolor{ArenaTableDetected}\makecell[c]{$+0.034$\\{\fontsize{6.6}{7.6}\selectfont$[+0.002,+0.075]$}} \\
knowledge consistency & \cellcolor{ArenaTableUndetected}\makecell[c]{$-0.020$\\{\fontsize{6.6}{7.6}\selectfont$[-0.057,+0.022]$}} & \cellcolor{ArenaTableUndetected}\makecell[c]{$+0.013$\\{\fontsize{6.6}{7.6}\selectfont$[-0.028,+0.052]$}} & \cellcolor{ArenaTableUndetected}\makecell[c]{$+0.020$\\{\fontsize{6.6}{7.6}\selectfont$[-0.022,+0.061]$}} & \cellcolor{ArenaTableUndetected}\makecell[c]{$+0.016$\\{\fontsize{6.6}{7.6}\selectfont$[-0.018,+0.061]$}} \\
goal consistency & \cellcolor{ArenaTableUndetected}\makecell[c]{$-0.012$\\{\fontsize{6.6}{7.6}\selectfont$[-0.035,+0.010]$}} & \cellcolor{ArenaTableUndetected}\makecell[c]{$+0.004$\\{\fontsize{6.6}{7.6}\selectfont$[-0.016,+0.030]$}} & \cellcolor{ArenaTableUndetected}\makecell[c]{$+0.011$\\{\fontsize{6.6}{7.6}\selectfont$[-0.012,+0.039]$}} & \cellcolor{ArenaTableUndetected}\makecell[c]{$+0.008$\\{\fontsize{6.6}{7.6}\selectfont$[-0.014,+0.032]$}} \\
\textit{best metric} & \cellcolor{ArenaTableUndetected}\makecell[c]{$+0.003$\\{\fontsize{6.6}{7.6}\selectfont$[-0.033,+0.050]$}} & \cellcolor{ArenaTableDetected}\makecell[c]{$+0.095$\\{\fontsize{6.6}{7.6}\selectfont$[+0.052,+0.144]$}} & \cellcolor{ArenaTableUndetected}\makecell[c]{$+0.032$\\{\fontsize{6.6}{7.6}\selectfont$[-0.013,+0.080]$}} & \cellcolor{ArenaTableDetected}\makecell[c]{$+0.034$\\{\fontsize{6.6}{7.6}\selectfont$[+0.002,+0.075]$}} \\
\addlinespace[3pt]
\multicolumn{5}{l}{\textbf{SWE-Together (population means)}} \\*
coverage rate & \cellcolor{ArenaTableUndetected}\makecell[c]{$+0.029$\\{\fontsize{6.6}{7.6}\selectfont$[-0.024,+0.090]$}} & \cellcolor{ArenaTableUndetected}\makecell[c]{$+0.022$\\{\fontsize{6.6}{7.6}\selectfont$[-0.030,+0.076]$}} & \cellcolor{ArenaTableDetected}\makecell[c]{$+0.059$\\{\fontsize{6.6}{7.6}\selectfont$[+0.006,+0.111]$}} & \cellcolor{ArenaTableUndetected}\makecell[c]{$+0.027$\\{\fontsize{6.6}{7.6}\selectfont$[-0.020,+0.080]$}} \\
weighted coverage & \cellcolor{ArenaTableUndetected}\makecell[c]{$+0.027$\\{\fontsize{6.6}{7.6}\selectfont$[-0.019,+0.075]$}} & \cellcolor{ArenaTableUndetected}\makecell[c]{$+0.017$\\{\fontsize{6.6}{7.6}\selectfont$[-0.025,+0.057]$}} & \cellcolor{ArenaTableDetected}\makecell[c]{$+0.043$\\{\fontsize{6.6}{7.6}\selectfont$[+0.002,+0.081]$}} & \cellcolor{ArenaTableUndetected}\makecell[c]{$+0.020$\\{\fontsize{6.6}{7.6}\selectfont$[-0.017,+0.062]$}} \\
scope precision & \cellcolor{ArenaTableUndetected}\makecell[c]{$-0.032$\\{\fontsize{6.6}{7.6}\selectfont$[-0.083,+0.017]$}} & \cellcolor{ArenaTableUndetected}\makecell[c]{$+0.011$\\{\fontsize{6.6}{7.6}\selectfont$[-0.029,+0.050]$}} & \cellcolor{ArenaTableUndetected}\makecell[c]{$-0.037$\\{\fontsize{6.6}{7.6}\selectfont$[-0.080,+0.013]$}} & \cellcolor{ArenaTableUndetected}\makecell[c]{$+0.029$\\{\fontsize{6.6}{7.6}\selectfont$[-0.020,+0.083]$}} \\
user correction & \cellcolor{ArenaTableUndetected}\makecell[c]{$-0.72$\\{\fontsize{6.6}{7.6}\selectfont$[-2.42,+1.08]$}} & \cellcolor{ArenaTableUndetected}\makecell[c]{$-0.16$\\{\fontsize{6.6}{7.6}\selectfont$[-1.42,+1.30]$}} & \cellcolor{ArenaTableUndetected}\makecell[c]{$-1.16$\\{\fontsize{6.6}{7.6}\selectfont$[-2.81,+0.72]$}} & \cellcolor{ArenaTableUndetected}\makecell[c]{$-0.34$\\{\fontsize{6.6}{7.6}\selectfont$[-1.60,+1.08]$}} \\
frustration count & \cellcolor{ArenaTableUndetected}\makecell[c]{$-1.25$\\{\fontsize{6.6}{7.6}\selectfont$[-3.00,+0.59]$}} & \cellcolor{ArenaTableUndetected}\makecell[c]{$-0.19$\\{\fontsize{6.6}{7.6}\selectfont$[-2.00,+1.65]$}} & \cellcolor{ArenaTableUndetected}\makecell[c]{$-0.96$\\{\fontsize{6.6}{7.6}\selectfont$[-2.49,+0.81]$}} & \cellcolor{ArenaTableUndetected}\makecell[c]{$+0.39$\\{\fontsize{6.6}{7.6}\selectfont$[-1.34,+2.08]$}} \\
\textit{best metric} & \cellcolor{ArenaTableUndetected}\makecell[c]{$+0.027$\\{\fontsize{6.6}{7.6}\selectfont$[-0.019,+0.075]$}} & \cellcolor{ArenaTableUndetected}\makecell[c]{$+0.017$\\{\fontsize{6.6}{7.6}\selectfont$[-0.025,+0.057]$}} & \cellcolor{ArenaTableDetected}\makecell[c]{$+0.059$\\{\fontsize{6.6}{7.6}\selectfont$[+0.006,+0.111]$}} & \cellcolor{ArenaTableUndetected}\makecell[c]{$+0.029$\\{\fontsize{6.6}{7.6}\selectfont$[-0.020,+0.083]$}} \\
\addlinespace[3pt]
\multicolumn{5}{l}{\textbf{Sim2Real (population metric)}} \\*
behavioral index & \cellcolor{ArenaTableUndetected}\makecell[c]{$+6.8$\\{\fontsize{6.6}{7.6}\selectfont$[-2.1,+17.3]$}} & \cellcolor{ArenaTableUndetected}\makecell[c]{$-1.7$\\{\fontsize{6.6}{7.6}\selectfont$[-8.9,+7.3]$}} & \cellcolor{ArenaTableUndetected}\makecell[c]{$+4.3$\\{\fontsize{6.6}{7.6}\selectfont$[-4.9,+12.8]$}} & \cellcolor{ArenaTableUndetected}\makecell[c]{$+1.5$\\{\fontsize{6.6}{7.6}\selectfont$[-5.6,+8.8]$}} \\
\addlinespace[3pt]
\end{longtable*}
\setcounter{table}{\PopulationDifferenceTableNumber}
\endgroup

\paragraph{Population-level metric breakdown.}
For approve-broken-work, the act-distribution family ranks the control population higher in every draw, compared with 90\% for SimulatorArena's best population mean. The L2 composite reaches 62\%. For front-loading, the L2 composite reaches 100\%, while external population means reach 87--99\%.

\paragraph{Comparison with real-user populations.}
Table~\ref{tab:metaeval-l2} adds a real-user reference comparison. We score real-user populations against a disjoint real reference and use their 2.5th score percentile as a cutoff. The table reports the fraction of perturbed populations below that cutoff.
Premature stop can improve L2 scores because control replays are longer than real-user sessions: truncation brings their lengths closer to the real-user reference.

\begin{table}[!htbp]
\centering
\AppendixWideTableStyle
\begin{tabularx}{\linewidth}{@{}>{\raggedright\arraybackslash}p{0.38\linewidth}*{8}{>{\centering\arraybackslash}X}@{}}
\toprule
 & \multicolumn{4}{c}{approve-broken-work} & \multicolumn{4}{c}{front-loading} \\
\cmidrule(lr){2-5} \cmidrule(lr){6-9}
\textbf{Sessions per population} & \textbf{10} & \textbf{25} & \textbf{50} & \textbf{100} & \textbf{10} & \textbf{25} & \textbf{50} & \textbf{100} \\
\midrule
Ours: L2 act distribution & \textbf{0.85} & \textbf{0.94} & \textbf{0.99} & \textbf{1.00} & \textbf{0.80} & 0.90 & \textbf{0.99} & \textbf{1.00} \\
Ours: L2 (mean of families) & 0.63 & 0.65 & 0.71 & 0.62 & 0.73 & 0.84 & 0.91 & \textbf{1.00} \\
Ours: best L3 family, population mean & 0.84 & 0.94 & \textbf{0.99} & \textbf{1.00} & 0.69 & 0.75 & 0.85 & 0.92 \\
SimulatorArena: best metric, population mean & 0.72 & 0.78 & 0.83 & 0.90 & 0.72 & \textbf{0.91} & 0.96 & 0.99 \\
VISTA: best metric, population mean & 0.19 & 0.12 & 0.04 & 0.01 & 0.70 & 0.80 & 0.92 & 0.98 \\
SWE-Together: best metric, population mean & 0.49 & 0.52 & 0.49 & 0.46 & 0.60 & 0.72 & 0.79 & 0.87 \\
\bottomrule
\end{tabularx}
\caption{Population win rates as the number of sessions per population varies. Each cell is the fraction of 200 matched draws in which the control population scores higher. Rows report the L2 act family and composite, the best L3 family, and each external evaluation suite\textquotesingle s best metric; trace-level scores are averaged within populations. Bold marks the highest value in each column.}
\label{tab:metaeval-l2-setsize}
\end{table}

\begin{table}[!htbp]
\centering
\AppendixTableStyle
\begin{tabularx}{\linewidth}{@{}>{\raggedright\arraybackslash}p{0.27\linewidth}*{3}{>{\centering\arraybackslash}X}@{}}
\toprule
\textbf{Evaluation suite / metric} & \textbf{\makecell{Approve\\broken work}} & \textbf{\makecell{Premature\\stop}} & \textbf{Front-loading} \\
\midrule
Ours: L2 (mean of families) & 1.00 & 0.53 & 1.00 \\
Ours: L2 family & \makecell{1.00\\{\scriptsize act distribution}} & \makecell{0.01\\{\scriptsize provenance}} & \makecell{1.00\\{\scriptsize act distribution}} \\
SimulatorArena & \makecell{1.00\\{\scriptsize mean of Likert writing}} & \makecell{0.00\\{\scriptsize mean of Turing detector}} & \makecell{1.00\\{\scriptsize mean of Likert writing}} \\
VISTA & \makecell{0.00\\{\scriptsize mean of goal consistency}} & \makecell{0.11\\{\scriptsize mean of goal consistency}} & \makecell{1.00\\{\scriptsize mean of profile consistency}} \\
SWE-Together & \makecell{1.00\\{\scriptsize mean of scope precision}} & \makecell{1.00\\{\scriptsize mean of weighted coverage}} & \makecell{1.00\\{\scriptsize mean of scope precision}} \\
\bottomrule
\end{tabularx}
\caption{Fraction of perturbed populations below the 2.5th percentile of real-user population scores. Columns show perturbations; metric names appear below values where needed.}
\label{tab:metaeval-l2}
\end{table}

\FloatBarrier

\subsubsection{Trace-Level Fidelity}
\label{sec:appendix-trace-sensitivity}
Tables~\ref{tab:sensitivity-appendix} and~\ref{tab:sensitivity-trace-diff} report the full trace-level sensitivity results corresponding to Figure~\ref{fig:sensitivity-trace}. The first reports paired win rates; the second reports score decreases and confidence intervals, with green marking detection under Section~\ref{sec:metaeval-setup}.
 The breakdown includes individual judge outputs: persona style and judgment criteria, intent preservation and fulfillment, constraint retention and satisfaction, groundedness and language match. It also reports per-trace complaint-share and turn-count deviations.

\clearpage
\begingroup
\SensitivityGridStyle
\begin{longtable}{@{}>{\raggedright\arraybackslash}m{86pt}>{\centering\arraybackslash}m{15pt}*{10}{>{\centering\arraybackslash}m{\dimexpr(\linewidth-101pt-22\tabcolsep)/10\relax}}@{}}
\caption{Trace-level paired win rates for each metric and perturbation. Colors run from red (0) through white (0.5) to green (1); ties count as half a win. Table~\ref{tab:sensitivity-trace-diff} reports score decreases and detection. Abbreviations follow Table~\ref{tab:sensitivity-set}, with Favor = agent favoritism and ArtNeg = negative profile artifact. Arrows give native score direction.}
\label{tab:sensitivity-appendix}\\
\toprule
\textbf{Metric} & \textbf{Dir.} & \textbf{Goal} & \textbf{AllC} & \textbf{Appr} & \textbf{Stop} & \textbf{Pers} & \textbf{Lang} & \textbf{CoD} & \textbf{Front} & \textbf{Favor} & \textbf{ArtNeg} \\
\midrule
\endfirsthead
\caption*{\textbf{Table~\ref{tab:sensitivity-appendix} (continued).}}\\
\toprule
\textbf{Metric} & \textbf{Dir.} & \textbf{Goal} & \textbf{AllC} & \textbf{Appr} & \textbf{Stop} & \textbf{Pers} & \textbf{Lang} & \textbf{CoD} & \textbf{Front} & \textbf{Favor} & \textbf{ArtNeg} \\
\midrule
\endhead
\midrule
\endfoot
\bottomrule
\endlastfoot
\multicolumn{12}{l}{\textbf{Ours}} \\*
persona & $\uparrow$ & \cellcolor[HTML]{B1CCBE}\textcolor[HTML]{000000}{0.67} & \cellcolor[HTML]{E7EDE9}\textcolor[HTML]{000000}{0.54} & \cellcolor[HTML]{C4D8CD}\textcolor[HTML]{000000}{0.63} & \cellcolor[HTML]{E4EBE6}\textcolor[HTML]{000000}{0.55} & \cellcolor[HTML]{A2C3B3}\textcolor[HTML]{000000}{0.70} & \cellcolor[HTML]{8FB7A4}\textcolor[HTML]{000000}{0.75} & \cellcolor[HTML]{ADCABB}\textcolor[HTML]{000000}{0.68} & \cellcolor[HTML]{F4E9E8}\textcolor[HTML]{000000}{0.46} & \cellcolor[HTML]{FAF9F8}\textcolor[HTML]{000000}{0.50} & \cellcolor[HTML]{E5ECE7}\textcolor[HTML]{000000}{0.55} \\
intent & $\uparrow$ & \cellcolor[HTML]{C1D6CB}\textcolor[HTML]{000000}{0.63} & \cellcolor[HTML]{F0F3F0}\textcolor[HTML]{000000}{0.52} & \cellcolor[HTML]{EEF1EE}\textcolor[HTML]{000000}{0.53} & \cellcolor[HTML]{E6EDE8}\textcolor[HTML]{000000}{0.55} & \cellcolor[HTML]{F4F5F3}\textcolor[HTML]{000000}{0.51} & \cellcolor[HTML]{E0E9E3}\textcolor[HTML]{000000}{0.56} & \cellcolor[HTML]{EDF1EE}\textcolor[HTML]{000000}{0.53} & \cellcolor[HTML]{EDF1EE}\textcolor[HTML]{000000}{0.53} & \cellcolor[HTML]{F6F7F5}\textcolor[HTML]{000000}{0.51} & \cellcolor[HTML]{F8F8F6}\textcolor[HTML]{000000}{0.51} \\
constraints & $\uparrow$ & \cellcolor[HTML]{C4D8CD}\textcolor[HTML]{000000}{0.63} & \cellcolor[HTML]{D0DFD7}\textcolor[HTML]{000000}{0.60} & \cellcolor[HTML]{F5ECEC}\textcolor[HTML]{000000}{0.47} & \cellcolor[HTML]{E0E9E4}\textcolor[HTML]{000000}{0.56} & \cellcolor[HTML]{DBE6E0}\textcolor[HTML]{000000}{0.57} & \cellcolor[HTML]{F9F6F5}\textcolor[HTML]{000000}{0.49} & \cellcolor[HTML]{F7F7F6}\textcolor[HTML]{000000}{0.51} & \cellcolor[HTML]{EFDCDB}\textcolor[HTML]{000000}{0.42} & \cellcolor[HTML]{E9EEEA}\textcolor[HTML]{000000}{0.54} & \cellcolor[HTML]{F7F0EF}\textcolor[HTML]{000000}{0.48} \\
\end{longtable}
\endgroup

\clearpage
\begingroup
\SensitivityGridStyle
\begin{longtable}{@{}>{\raggedright\arraybackslash}m{86pt}>{\centering\arraybackslash}m{15pt}*{5}{>{\centering\arraybackslash}m{\dimexpr(\linewidth-101pt-12\tabcolsep)/5\relax}}@{}}
\caption{Trace-level sensitivity corresponding to Table~\ref{tab:sensitivity-appendix}. Each cell reports the mean control-minus-perturbed score difference above its bootstrap 95\% confidence interval, in the metric's native units. Green marks an interval entirely above zero; red marks otherwise. Selected metrics and abbreviations follow Table~\ref{tab:sensitivity-appendix}.}
\label{tab:sensitivity-trace-diff}\\
\toprule
\textbf{Metric} & \textbf{Dir.} & \textbf{Goal} & \textbf{AllC} & \textbf{Appr} & \textbf{Stop} & \textbf{Pers} \\
\midrule
\endfirsthead
\caption*{\textbf{Table~\ref{tab:sensitivity-trace-diff} (continued).}}\\
\toprule
\textbf{Metric} & \textbf{Dir.} & \textbf{Goal} & \textbf{AllC} & \textbf{Appr} & \textbf{Stop} & \textbf{Pers} \\
\midrule
\endhead
\midrule
\endfoot
\bottomrule
\endlastfoot
\multicolumn{7}{l}{\textbf{Ours}} \\*
persona & $\uparrow$ & \cellcolor{ArenaTableDetected}\shortstack[c]{$+0.069$\\[-3pt]{\fontsize{6.6}{7.6}\selectfont$[0.052,0.085]$}} & \cellcolor{ArenaTableUndetected}\shortstack[c]{$+0.006$\\[-3pt]{\fontsize{6.6}{7.6}\selectfont$[-0.007,0.019]$}} & \cellcolor{ArenaTableDetected}\shortstack[c]{$+0.050$\\[-3pt]{\fontsize{6.6}{7.6}\selectfont$[0.035,0.066]$}} & \cellcolor{ArenaTableDetected}\shortstack[c]{$+0.022$\\[-3pt]{\fontsize{6.6}{7.6}\selectfont$[0.009,0.035]$}} & \cellcolor{ArenaTableDetected}\shortstack[c]{$+0.095$\\[-3pt]{\fontsize{6.6}{7.6}\selectfont$[0.080,0.110]$}} \\
intent & $\uparrow$ & \cellcolor{ArenaTableDetected}\shortstack[c]{$+0.264$\\[-3pt]{\fontsize{6.6}{7.6}\selectfont$[0.217,0.311]$}} & \cellcolor{ArenaTableDetected}\shortstack[c]{$+0.048$\\[-3pt]{\fontsize{6.6}{7.6}\selectfont$[0.015,0.083]$}} & \cellcolor{ArenaTableDetected}\shortstack[c]{$+0.057$\\[-3pt]{\fontsize{6.6}{7.6}\selectfont$[0.018,0.097]$}} & \cellcolor{ArenaTableDetected}\shortstack[c]{$+0.093$\\[-3pt]{\fontsize{6.6}{7.6}\selectfont$[0.054,0.133]$}} & \cellcolor{ArenaTableUndetected}\shortstack[c]{$+0.029$\\[-3pt]{\fontsize{6.6}{7.6}\selectfont$[-0.009,0.068]$}} \\
constraints & $\uparrow$ & \cellcolor{ArenaTableDetected}\shortstack[c]{$+0.085$\\[-3pt]{\fontsize{6.6}{7.6}\selectfont$[0.053,0.121]$}} & \cellcolor{ArenaTableDetected}\shortstack[c]{$+0.064$\\[-3pt]{\fontsize{6.6}{7.6}\selectfont$[0.032,0.097]$}} & \cellcolor{ArenaTableUndetected}\shortstack[c]{$-0.019$\\[-3pt]{\fontsize{6.6}{7.6}\selectfont$[-0.052,0.013]$}} & \cellcolor{ArenaTableDetected}\shortstack[c]{$+0.056$\\[-3pt]{\fontsize{6.6}{7.6}\selectfont$[0.021,0.094]$}} & \cellcolor{ArenaTableUndetected}\shortstack[c]{$+0.024$\\[-3pt]{\fontsize{6.6}{7.6}\selectfont$[-0.006,0.052]$}} \\
\end{longtable}
\edef\TraceDifferenceTableNumber{\number\value{table}}
\clearpage
\begin{longtable*}{@{}>{\raggedright\arraybackslash}m{86pt}>{\centering\arraybackslash}m{15pt}*{5}{>{\centering\arraybackslash}m{\dimexpr(\linewidth-101pt-12\tabcolsep)/5\relax}}@{}}
\caption*{\textbf{Table~\ref{tab:sensitivity-trace-diff} (continued).}}\\
\toprule
\textbf{Metric} & \textbf{Dir.} & \textbf{Lang} & \textbf{CoD} & \textbf{Front} & \textbf{Favor} & \textbf{ArtNeg} \\
\midrule
\endfirsthead
\caption*{\textbf{Table~\ref{tab:sensitivity-trace-diff} (continued).}}\\
\toprule
\textbf{Metric} & \textbf{Dir.} & \textbf{Lang} & \textbf{CoD} & \textbf{Front} & \textbf{Favor} & \textbf{ArtNeg} \\
\midrule
\endhead
\midrule
\endfoot
\bottomrule
\endlastfoot
\multicolumn{7}{l}{\textbf{Ours}} \\*
persona & $\uparrow$ & \cellcolor{ArenaTableDetected}\shortstack[c]{$+0.115$\\[-3pt]{\fontsize{6.6}{7.6}\selectfont$[0.100,0.131]$}} & \cellcolor{ArenaTableDetected}\shortstack[c]{$+0.065$\\[-3pt]{\fontsize{6.6}{7.6}\selectfont$[0.051,0.081]$}} & \cellcolor{ArenaTableUndetected}\shortstack[c]{$-0.005$\\[-3pt]{\fontsize{6.6}{7.6}\selectfont$[-0.018,0.009]$}} & \cellcolor{ArenaTableUndetected}\shortstack[c]{$+0.027$\\[-3pt]{\fontsize{6.6}{7.6}\selectfont$[-0.013,0.067]$}} & \cellcolor{ArenaTableDetected}\shortstack[c]{$+0.016$\\[-3pt]{\fontsize{6.6}{7.6}\selectfont$[0.002,0.032]$}} \\
intent & $\uparrow$ & \cellcolor{ArenaTableDetected}\shortstack[c]{$+0.122$\\[-3pt]{\fontsize{6.6}{7.6}\selectfont$[0.083,0.163]$}} & \cellcolor{ArenaTableDetected}\shortstack[c]{$+0.060$\\[-3pt]{\fontsize{6.6}{7.6}\selectfont$[0.023,0.099]$}} & \cellcolor{ArenaTableDetected}\shortstack[c]{$+0.059$\\[-3pt]{\fontsize{6.6}{7.6}\selectfont$[0.023,0.097]$}} & \cellcolor{ArenaTableUndetected}\shortstack[c]{$+0.018$\\[-3pt]{\fontsize{6.6}{7.6}\selectfont$[-0.055,0.109]$}} & \cellcolor{ArenaTableUndetected}\shortstack[c]{$+0.010$\\[-3pt]{\fontsize{6.6}{7.6}\selectfont$[-0.025,0.046]$}} \\
constraints & $\uparrow$ & \cellcolor{ArenaTableUndetected}\shortstack[c]{$+0.002$\\[-3pt]{\fontsize{6.6}{7.6}\selectfont$[-0.029,0.033]$}} & \cellcolor{ArenaTableUndetected}\shortstack[c]{$+0.019$\\[-3pt]{\fontsize{6.6}{7.6}\selectfont$[-0.009,0.051]$}} & \cellcolor{ArenaTableUndetected}\shortstack[c]{$-0.028$\\[-3pt]{\fontsize{6.6}{7.6}\selectfont$[-0.057,0.002]$}} & \cellcolor{ArenaTableUndetected}\shortstack[c]{$+0.012$\\[-3pt]{\fontsize{6.6}{7.6}\selectfont$[-0.042,0.072]$}} & \cellcolor{ArenaTableUndetected}\shortstack[c]{$-0.006$\\[-3pt]{\fontsize{6.6}{7.6}\selectfont$[-0.034,0.021]$}} \\
\end{longtable*}
\setcounter{table}{\TraceDifferenceTableNumber}
\endgroup

\FloatBarrier

\Needspace{22\baselineskip}
\subsection{Human Study}
\label{sec:appendix-human-study}
\begin{table}[!htbp]
\centering
\AppendixTableStyle
\begin{tabular*}{\linewidth}{@{\extracolsep{\fill}}lcc@{}}
\toprule
\textbf{Perturbation} & \textbf{Acc. (unsure $=0.5$)} & \textbf{Agreement} \\
\midrule
 Agent favoritism & 0.53 & 0.65 \\
 Negative profile artifact & 0.51 & 0.65 \\
 Adversarial goal & 0.57 & 0.45 \\
 All constraints removed & 0.53 & 0.75 \\
 Approve broken work & 0.66 & 0.45 \\
 Premature stop & 0.51 & 0.55 \\
 Persona swap & 0.57 & 0.35 \\
 Language switch & 0.68 & 0.10 \\
 Co-debugger & 0.51 & 0.45 \\
 Front-loading & 0.50 & 0.70 \\
\midrule
\textbf{All} & \textbf{0.56} & \textbf{0.51} \\
\bottomrule
\end{tabular*}
\caption{Paired identification of perturbed replays. Accuracy gives half credit to unsure responses. Agreement is the fraction of pairs receiving the same response from both raters, including unsure.}
\label{tab:human-study}
\end{table}

\FloatBarrier

\subsection{Filtering}
\label{sec:appendix-filtering}
These tables expand the metric-level AUROC and retention results in Section~\ref{sec:filtering-rationale}.

\subsubsection{Agent Differentiation}
\label{sec:appendix-agent-filtering}
Table~\ref{tab:l1-agent-power} expands Table~\ref{tab:best-filtering-agent} with individual metrics and control-population retention.

\FloatBarrier

\subsubsection{Population-Level Fidelity}
\label{sec:appendix-population-filtering}
Table~\ref{tab:population-filtering} expands Table~\ref{tab:best-filtering-population} for approve-broken-work, premature stop and front-loading. It reports AUROC and control-population retention at 5\% perturbed survival for each replacement fraction.

\begin{table}[H]
\centering
\AppendixWideTableStyle
\setlength{\tabcolsep}{1.25pt}
\renewcommand{\arraystretch}{0.94}
\begin{tabularx}{\linewidth}{@{}>{\raggedright\arraybackslash}p{0.34\linewidth}*{8}{>{\centering\arraybackslash}X}@{}}
\toprule
 & \multicolumn{8}{c}{Fraction of replays replaced by perturbed replays ($\alpha$)} \\
\cmidrule(lr){2-9}
 & \multicolumn{2}{c}{10\%} & \multicolumn{2}{c}{25\%} & \multicolumn{2}{c}{50\%} & \multicolumn{2}{c}{100\%} \\
\cmidrule(lr){2-3} \cmidrule(lr){4-5} \cmidrule(lr){6-7} \cmidrule(lr){8-9}
\textbf{Metric} & \textbf{AUROC} & \textbf{\makecell{ctrl@\\5\%FPR}} & \textbf{AUROC} & \textbf{\makecell{ctrl@\\5\%FPR}} & \textbf{AUROC} & \textbf{\makecell{ctrl@\\5\%FPR}} & \textbf{AUROC} & \textbf{\makecell{ctrl@\\5\%FPR}} \\
\midrule
\multicolumn{9}{l}{\textbf{Approve broken work}} \\
L2: mean of families & 0.37 & 0.03 & 0.29 & 0.01 & 0.26 & 0.01 & 0.71 & 0.21 \\
L2: act distribution & 0.23 & 0.01 & 0.14 & 0.00 & 0.17 & 0.01 & \textbf{1.00} & \textbf{1.00} \\
L2: dynamics & 0.49 & 0.05 & 0.46 & 0.05 & 0.42 & 0.06 & 0.39 & 0.05 \\
L2: outcome sensitivity & 0.45 & 0.02 & 0.39 & 0.02 & 0.29 & 0.01 & 0.15 & 0.01 \\
L2: provenance & 0.47 & 0.03 & 0.43 & 0.04 & 0.45 & 0.03 & 0.71 & 0.15 \\
SimulatorArena: composite, set mean & 0.39 & 0.03 & 0.24 & 0.00 & 0.07 & 0.00 & 0.00 & 0.00 \\
SimulatorArena: best metric, set mean & \textbf{0.53}\textsuperscript{\normalfont\fontsize{6}{6}\selectfont W} & \textbf{0.07} & \textbf{0.58}\textsuperscript{\normalfont\fontsize{6}{6}\selectfont W} & \textbf{0.12} & \textbf{0.66}\textsuperscript{\normalfont\fontsize{6}{6}\selectfont W} & \textbf{0.17} & 0.78\textsuperscript{\normalfont\fontsize{6}{6}\selectfont W} & 0.26 \\
VISTA: composite, set mean & 0.40 & 0.03 & 0.24 & 0.01 & 0.07 & 0.00 & 0.00 & 0.00 \\
VISTA: best metric, set mean & 0.43\textsuperscript{\normalfont\fontsize{6}{6}\selectfont K} & 0.03 & 0.31\textsuperscript{\normalfont\fontsize{6}{6}\selectfont K} & 0.01 & 0.16\textsuperscript{\normalfont\fontsize{6}{6}\selectfont K} & 0.00 & 0.01\textsuperscript{\normalfont\fontsize{6}{6}\selectfont K} & 0.00 \\
SWE-Together: composite, set mean & 0.35 & 0.01 & 0.16 & 0.00 & 0.02 & 0.00 & 0.00 & 0.00 \\
SWE-Together: best metric, set mean & 0.50\textsuperscript{\normalfont\fontsize{6}{6}\selectfont S} & 0.04 & 0.50\textsuperscript{\normalfont\fontsize{6}{6}\selectfont S} & 0.04 & 0.52\textsuperscript{\normalfont\fontsize{6}{6}\selectfont S} & 0.04 & 0.50\textsuperscript{\normalfont\fontsize{6}{6}\selectfont S} & 0.09 \\
Sim2Real: USI behavioral index & 0.49 & 0.03 & 0.48 & 0.03 & 0.45 & 0.01 & 0.55 & 0.07 \\
\midrule
\multicolumn{9}{l}{\textbf{Premature stop}} \\
L2: mean of families & 0.44 & 0.04 & 0.34 & 0.01 & 0.19 & 0.00 & 0.02 & 0.00 \\
L2: act distribution & 0.43 & 0.01 & 0.32 & 0.00 & 0.11 & 0.00 & 0.00 & 0.00 \\
L2: dynamics & 0.43 & 0.01 & 0.30 & 0.01 & 0.17 & 0.00 & 0.27 & 0.01 \\
L2: outcome sensitivity & 0.49 & 0.07 & 0.48 & 0.07 & 0.46 & 0.03 & 0.14 & 0.01 \\
L2: provenance & 0.49 & 0.04 & 0.49 & 0.03 & 0.49 & 0.02 & 0.46 & 0.02 \\
SimulatorArena: composite, set mean & 0.43 & 0.04 & 0.33 & 0.02 & 0.19 & 0.01 & 0.02 & 0.00 \\
SimulatorArena: best metric, set mean & 0.53\textsuperscript{\normalfont\fontsize{6}{6}\selectfont T} & 0.07 & 0.58\textsuperscript{\normalfont\fontsize{6}{6}\selectfont T} & 0.08 & 0.62\textsuperscript{\normalfont\fontsize{6}{6}\selectfont T} & 0.12 & 0.69\textsuperscript{\normalfont\fontsize{6}{6}\selectfont T} & 0.19 \\
VISTA: composite, set mean & 0.41 & 0.03 & 0.29 & 0.01 & 0.13 & 0.00 & 0.01 & 0.00 \\
VISTA: best metric, set mean & 0.45\textsuperscript{\normalfont\fontsize{6}{6}\selectfont G} & 0.03 & 0.38\textsuperscript{\normalfont\fontsize{6}{6}\selectfont G} & 0.02 & 0.27\textsuperscript{\normalfont\fontsize{6}{6}\selectfont G} & 0.00 & 0.12\textsuperscript{\normalfont\fontsize{6}{6}\selectfont G} & 0.00 \\
SWE-Together: composite, set mean & 0.42 & 0.04 & 0.32 & 0.01 & 0.16 & 0.00 & 0.01 & 0.00 \\
SWE-Together: best metric, set mean & \textbf{0.69}\textsuperscript{\normalfont\fontsize{6}{6}\selectfont C} & \textbf{0.18} & \textbf{0.88}\textsuperscript{\normalfont\fontsize{6}{6}\selectfont C} & \textbf{0.55} & \textbf{0.99}\textsuperscript{\normalfont\fontsize{6}{6}\selectfont C} & \textbf{0.96} & \textbf{1.00}\textsuperscript{\normalfont\fontsize{6}{6}\selectfont V} & \textbf{1.00} \\
Sim2Real: USI behavioral index & 0.48 & 0.04 & 0.45 & 0.03 & 0.44 & 0.03 & 0.60 & 0.06 \\
\midrule
\multicolumn{9}{l}{\textbf{Front-loading}} \\
L2: mean of families & 0.54 & 0.06 & 0.61 & 0.10 & 0.74 & 0.19 & 0.92 & 0.63 \\
L2: act distribution & \textbf{0.56} & \textbf{0.10} & \textbf{0.65} & 0.14 & \textbf{0.78} & \textbf{0.33} & \textbf{0.96} & \textbf{0.77} \\
L2: dynamics & 0.50 & 0.03 & 0.47 & 0.01 & 0.43 & 0.01 & 0.38 & 0.01 \\
L2: outcome sensitivity & 0.51 & 0.04 & 0.53 & 0.09 & 0.56 & 0.17 & 0.65 & 0.20 \\
L2: provenance & 0.54 & 0.06 & 0.61 & 0.04 & 0.74 & 0.25 & 0.93 & 0.68 \\
SimulatorArena: composite, set mean & 0.56 & 0.07 & 0.64 & \textbf{0.15} & 0.74 & 0.29 & 0.90 & 0.70 \\
SimulatorArena: best metric, set mean & 0.56\textsuperscript{\normalfont\fontsize{6}{6}\selectfont W} & 0.07 & 0.64\textsuperscript{\normalfont\fontsize{6}{6}\selectfont W} & 0.14 & 0.74\textsuperscript{\normalfont\fontsize{6}{6}\selectfont W} & 0.27 & 0.90\textsuperscript{\normalfont\fontsize{6}{6}\selectfont W} & 0.58 \\
VISTA: composite, set mean & 0.53 & 0.07 & 0.60 & 0.09 & 0.69 & 0.17 & 0.84 & 0.46 \\
VISTA: best metric, set mean & 0.56\textsuperscript{\normalfont\fontsize{6}{6}\selectfont F} & 0.08 & 0.64\textsuperscript{\normalfont\fontsize{6}{6}\selectfont F} & 0.14 & 0.77\textsuperscript{\normalfont\fontsize{6}{6}\selectfont F} & 0.24 & 0.92\textsuperscript{\normalfont\fontsize{6}{6}\selectfont F} & 0.60 \\
SWE-Together: composite, set mean & 0.53 & 0.07 & 0.58 & 0.10 & 0.65 & 0.14 & 0.79 & 0.30 \\
SWE-Together: best metric, set mean & 0.53\textsuperscript{\normalfont\fontsize{6}{6}\selectfont V} & 0.07 & 0.59\textsuperscript{\normalfont\fontsize{6}{6}\selectfont V} & 0.07 & 0.65\textsuperscript{\normalfont\fontsize{6}{6}\selectfont V} & 0.12 & 0.80\textsuperscript{\normalfont\fontsize{6}{6}\selectfont S} & 0.34 \\
Sim2Real: USI behavioral index & 0.50 & 0.06 & 0.52 & 0.06 & 0.56 & 0.06 & 0.61 & 0.08 \\
\bottomrule
\end{tabularx}
\caption{Full population-level filtering results corresponding to Table~\ref{tab:best-filtering-population}. Rows compare individual metrics, composites and best metrics within each perturbation; columns vary the perturbed fraction. Cells report AUROC and control retention at 5\% perturbed survival. Bold marks the highest value within each perturbation and column.}
\label{tab:population-filtering}
\par\vspace{4pt}
{\fontsize{8}{9.7}\selectfont\raggedright\textit{Selected metrics.} W writing; K knowledge; S scope prec.; T Turing; G goal; C w. coverage; V coverage; F profile.\par}
\end{table}

\FloatBarrier

\subsubsection{Trace-Level Fidelity}
\label{sec:appendix-trace-filtering}
Table~\ref{tab:l3-filtering-pooled} expands Table~\ref{tab:best-filtering-trace}. It pools adversarial goal, all constraints removed, persona swap, language switch and co-debugger. The two column groups compare perturbed replays with control replays and real-user traces, respectively, reporting AUROC and retention at 5\% and 10\% perturbed survival.

\begin{table}[H]
\centering
\AppendixTableStyle
\setlength{\tabcolsep}{3.5pt}
\begin{tabularx}{\linewidth}{@{}>{\raggedright\arraybackslash}p{0.36\linewidth}*{6}{>{\centering\arraybackslash}X}@{}}
\toprule
 & \multicolumn{3}{c}{\textbf{\makecell{Control vs perturbed\\(control positive) $\uparrow$}}} & \multicolumn{3}{c}{\textbf{\makecell{Real vs perturbed\\(real positive) $\uparrow$}}} \\
\cmidrule(lr){2-4}\cmidrule(lr){5-7}
\textbf{Metric} & \textbf{AUROC} & \textbf{\makecell{control\\@5\%\\FPR}} & \textbf{\makecell{control\\@10\%\\FPR}} & \textbf{AUROC} & \textbf{\makecell{real\\@5\%\\FPR}} & \textbf{\makecell{real\\@10\%\\FPR}} \\
\midrule
Ours: L3 (mean of families)$^\ddagger$ & 0.61 & 0.07 & \textbf{0.17} & n/a & n/a & n/a \\
Ours: persona & \textbf{0.62} & \textbf{0.10} & \textbf{0.17} & \textbf{0.90} & \textbf{0.68} & \textbf{0.77} \\
Ours: intent$^\ddagger$ & 0.55 & 0.06$^{\dagger}$ & 0.11$^{\dagger}$ & n/a & n/a & n/a \\
Ours: constraints$^\ddagger$ & 0.54 & 0.06$^{\dagger}$ & 0.11$^{\dagger}$ & n/a & n/a & n/a \\
SimulatorArena: Likert writing & 0.55 & 0.05$^{\dagger}$ & 0.11$^{\dagger}$ & 0.88 & 0.28$^{\dagger}$ & 0.55$^{\dagger}$ \\
SimulatorArena: Likert interaction & 0.52 & 0.06$^{\dagger}$ & 0.12$^{\dagger}$ & 0.75 & 0.26$^{\dagger}$ & 0.52$^{\dagger}$ \\
SimulatorArena: interaction rating & 0.49 & 0.05$^{\dagger}$ & 0.10$^{\dagger}$ & 0.45 & 0.03$^{\dagger}$ & 0.06$^{\dagger}$ \\
SimulatorArena: document rating & 0.51 & 0.06$^{\dagger}$ & 0.10$^{\dagger}$ & 0.47 & 0.02$^{\dagger}$ & 0.07$^{\dagger}$ \\
SimulatorArena: Turing detector & 0.49 & 0.05$^{\dagger}$ & 0.09$^{\dagger}$ & 0.46 & 0.04$^{\dagger}$ & 0.08$^{\dagger}$ \\
SimulatorArena: composite & 0.52 & 0.05 & 0.10 & 0.75 & 0.24 & 0.40 \\
VISTA: goal consistency & 0.52 & 0.05 & 0.11 & 0.58 & 0.06 & 0.12 \\
VISTA: profile consistency & 0.54 & 0.06 & 0.11 & 0.65 & 0.07 & 0.15 \\
VISTA: knowledge consistency & 0.52 & 0.05 & 0.11 & 0.66 & 0.09 & 0.17 \\
VISTA: composite & 0.55 & 0.06 & 0.11 & 0.68 & 0.09 & 0.19 \\
SWE-Together: coverage rate$^\ddagger$ & 0.52 & 0.05 & 0.11 & n/a & n/a & n/a \\
SWE-Together: user correction & 0.49 & 0.05 & 0.11 & 0.75 & 0.12 & 0.24 \\
SWE-Together: user-correction deviation & 0.48 & 0.04 & 0.08 & 0.73 & 0.06 & 0.13 \\
SWE-Together: frustration count & 0.50 & 0.05$^{\dagger}$ & 0.10$^{\dagger}$ & 0.61 & 0.06$^{\dagger}$ & 0.13$^{\dagger}$ \\
SWE-Together: frustration-count deviation & 0.50 & 0.05$^{\dagger}$ & 0.10$^{\dagger}$ & 0.61 & 0.06$^{\dagger}$ & 0.13$^{\dagger}$ \\
\bottomrule
\end{tabularx}
\caption{Full trace-level filtering results corresponding to Table~\ref{tab:best-filtering-trace}, pooling 2,636 perturbed replays from 533 source sessions across 5 perturbations. Rows report individual metrics and composites; columns give AUROC and positive-trace retention at 5\% and 10\% perturbed survival, with control replays (left) or real-user traces (right) as positives. $^{\dagger}$ marks interpolated operating points for metrics with fewer than 50 distinct scores; $^{\ddagger}$ marks  real-user comparisons omitted to avoid self-comparison with the reference; this also prevents computing the L3 aggregate. Bold marks the highest value in each column.}
\label{tab:l3-filtering-pooled}
\end{table}

\FloatBarrier

\end{document}